\documentclass[11pt]{article}

\usepackage[final]{acl}

\usepackage{times}
\usepackage{latexsym}
\usepackage{amssymb}

\usepackage[T1]{fontenc}
\usepackage[utf8]{inputenc}
\usepackage{microtype}
\usepackage{inconsolata}
\usepackage{graphicx}

\usepackage{multirow}
\usepackage{booktabs}
\usepackage{CJKutf8}
\usepackage{enumitem}

\title{CogToM: A Comprehensive Theory of Mind Benchmark\\inspired by Human Cognition for Large Language Models}

\author{
    \textbf{Haibo Tong}$^{1,4}$\thanks{ Equal contribution.},
    \textbf{Zeyang Yue}$^{1}$\footnotemark[1],
    \textbf{Feifei Zhao}$^{1}$\footnotemark[1]\footnotemark[2],
    \textbf{Erliang Lin}$^{1}$,
    \textbf{Lu Jia}$^{1}$,
    \\
    \textbf{Ruolin Chen}$^{1,4}$,
    \textbf{Yinqian Sun}$^{1}$,
    \textbf{Qian Zhang}$^{1}$,
    \textbf{Yi Zeng}$^{1,2,3,4,5}$\thanks{ Corresponding author.},
    \\
    $^{1}$BrainCog Lab, Institute of Automation, Chinese Academy of Sciences 
    \\
    $^{2}$Beijing Institute of AI Safety and Governance (Beijing-AISI)
    \\
    $^{3}$Beijing Key Laboratory of Safe AI and Superalignment
    \\
    $^{4}$School of Artificial Intelligence, UCAS \quad
    $^{5}$Long-term AI
    \\
    \texttt{\{tonghaibo2023, zhaofeifei2014, yi.zeng\}@ia.ac.cn}
}

\begin{document}
\maketitle

\begin{abstract}
Whether Large Language Models (LLMs) truly possess human-like Theory of Mind (ToM) capabilities has garnered increasing attention. However, existing benchmarks remain largely restricted to narrow paradigms like false belief tasks, failing to capture the full spectrum of human cognitive mechanisms. We introduce \textbf{CogToM}, a comprehensive, theoretically grounded benchmark comprising over 8000 bilingual instances across 46 paradigms, validated by 49 human annotator.A systematic evaluation of 22 representative models, including frontier models like GPT-5.1 and Qwen3-Max, reveals significant performance heterogeneities and highlights persistent bottlenecks in specific dimensions. Further analysis based on human cognitive patterns suggests potential divergences between LLM and human cognitive structures. CogToM offers a robust instrument and perspective for investigating the evolving cognitive boundaries of LLMs.
\end{abstract}

\section{Introduction}

Theory of Mind (ToM) is fundamental to human social interaction, enabling us to represent and infer others' mental states: for example, recognizing a comment on room temperature as a veiled request to close a window. Since its first introduction~\cite{premack1978does}, psychologists have developed diverse paradigms to study this multifaceted cognitive capacity, including False Belief tasks~\cite{Task-FBC, Task-FBL}, Faux Pas recognition~\cite{Task-FRT}, and non-literal comprehension tasks like Strange Stories~\cite{Task-SS}.

As LLMs increasingly engage in social scenarios, evaluating their human-like ToM capabilities has become a central challenge. Early findings based on false-belief tasks\cite{tomi} suggested near-human proficiency~\cite{kosinski2024evaluating}. However, these claims have been scrutinized due to models' vulnerability to context perturbations, indicating a reliance on shallow pattern matching rather than genuine reasoning~\cite{ullman2023large, shapira2024clever}. Despite subsequent efforts to extend reasoning depth~\cite{hitom}, introduce dialogue scenarios~\cite{fantom} or incorporate diverse dimensions~\cite{tombench}, existing benchmarks still lack the paradigm diversity compared with psychological research. Consequently, they fail to encompass the full cognitive spectrum or adequately characterize the potential discrepancies between machine and human intelligence.

To bridge this gap, we introduce \textbf{CogToM}, a comprehensive ToM evaluation benchmark for LLMs inspired by human cognitive psychology. CogToM encompasses 46 task paradigms and comprises over 8,000 bilingual (Chinese-English) instances, all of which have been meticulously annotated and verified by multiple human experts. Distinguished by its substantial scale and excellent text and annotation quality, CogToM offers unprecedented task coverage by synergizing established evaluation tasks with newly introduced paradigms.

We use CogToM to conduct a large-scale evaluation of 22 representative LLMs, spanning diverse release timelines, parameter scales, and model families, including frontier models such as GPT-5.1 and Qwen3-Max. The observed variance in performance across these models underscores the robust discriminative power of our benchmark. Furthermore, by integrating human-centric cognitive analyses, specifically the joint correlation between model accuracy and human inter-annotator agreement rate (IAR), alongside an assessment of alignment with human developmental milestones, our findings suggest the potential presence of Moravec’s Paradox~\cite{moravec1988mind} within the cognitive architectures of modern LLMs.
This work makes the following key contributions:


    
    


\begin{itemize}[leftmargin=*, nosep]

    \item \textbf{A Theoretically Grounded, Large-scale ToM Benchmark:} We introduce \textbf{CogToM}, the most comprehensive ToM benchmark to date, featuring 46 task paradigms and 8,000+ expert-verified bilingual instances.
    
    \item \textbf{Large-scale Systematic Evaluation:} Through a extensive evaluation of 22 representative LLMs, we provide a detailed landscape of current LLMs' ToM ability boundaries, demonstrating our benchmark's superior discriminative power.
    
    \item \textbf{Insights into Cognitive Heterogeneity:} By evaluating models against human-centric metrics and milestones, this study reveals fundamental distinctions between LLMs and human ToM, offering a critical perspective for understanding this cognitive divide.

\end{itemize}

\section{Related Works}

In the context of ToM evaluation of LLMs, story-based textual datasets play a pivotal role. As a early effort, ToMi established a systematic evaluation framework centered on false belief and second-order belief reasoning~\cite{tomi}, laying a pivotal foundation for subsequent ToM datasets and LLM evaluation paradigms. Following this, numerous benchmarks have sought to expand the false belief paradigm through diverse approaches, including the introduction of rich narrative stimuli~\cite{tomchallenges}, dialogue-based interaction formats~\cite{fantom}, higher-order reasoning depth~\cite{hitom}, and complex knowledge-belief scenario constructions~\cite{mindgames, simpletom, bigtom}. Despite these effortss, such benchmarks remain essentially tethered to the false belief framework. Based on results from 40 crafted false belief tasks, Kosinski suggested that ToM may have spontaneously ``emerged'' in LLMs~\cite{kosinski2024evaluating}. However, this conclusion was subsequently challenged~\cite{ullman2023large, shapira2024clever}, underscoring the inherent limitations of evaluating ToM capabilities through evaluation methods strictly confined to the false belief task paradigm.

Subsequent research has sought to introduce more diverse cognitive dimensions to overcome the limitations of single-task evaluations. Strachan et al. employed five distinct ToM task types to assess LLMs~\cite{strachan2024testing}. EmoBench specializes in the comprehensive evaluation of emotional reasoning~\cite{emobench}. Although OpenToM~\cite{opentom} and NegotiationToM~\cite{negotiationtom} formally resemble traditional false belief tasks, their underlying designs inherently incorporate assessments across cognitive dimensions including emotiona, desire and intention. Drawing inspiration from the ATOMS framework~\cite{atoms}, ToMBench~\cite{tombench} further extends this coverage by adopting eight different task paradigms. Nevertheless, these benchmarks still remain largely constrained by narrow task paradigms and limited theoretical grounding. Consequently, they struggle to encompass the broad spectrum of human Theory of Mind or adequately characterize the potential cognitive discrepancies between LLMs and human intelligence.

\begin{figure}[htbp]
    \centering
    \includegraphics[width=0.95\linewidth]{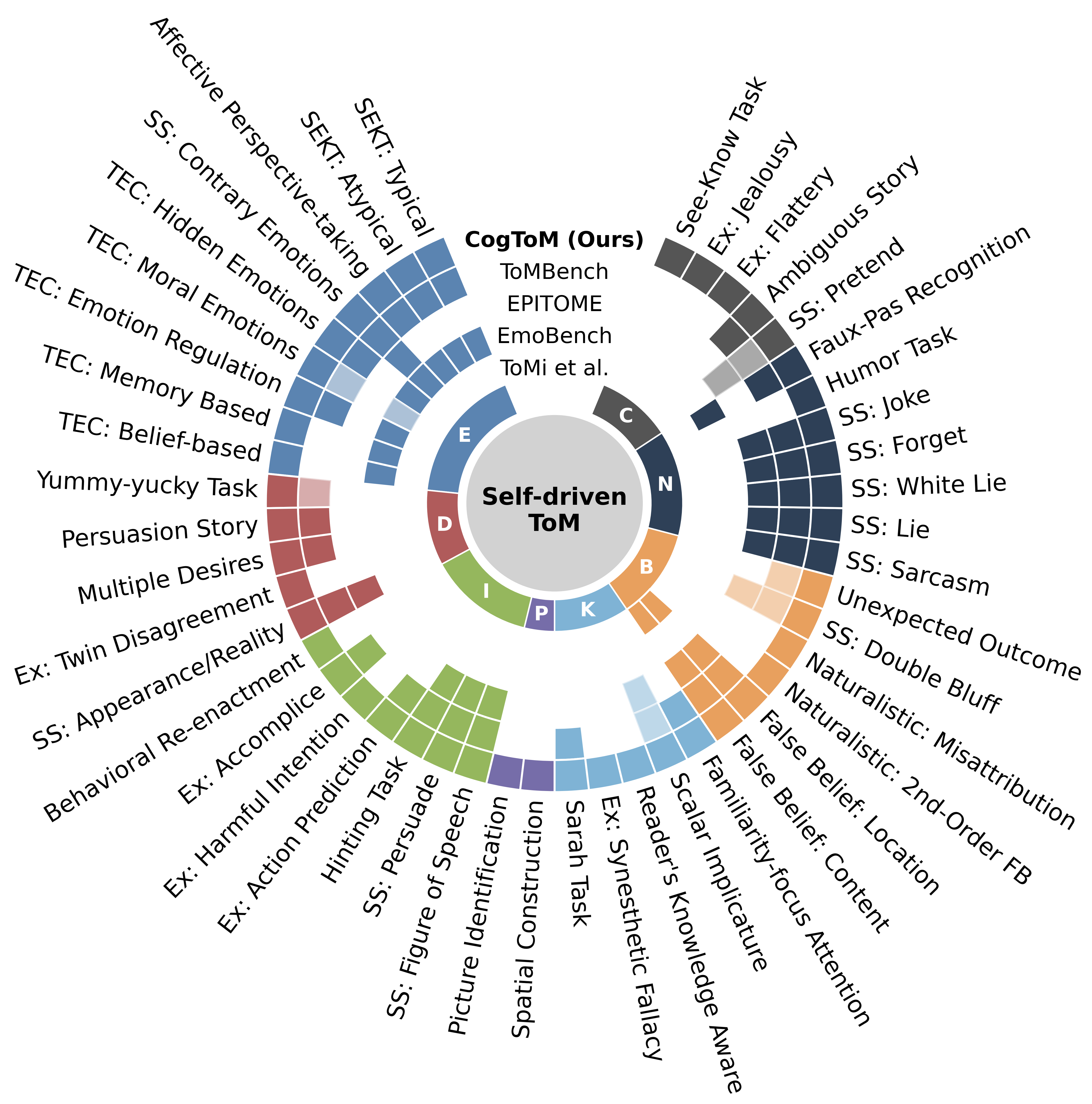}
    \caption{Comparison of task coverage across different ToM benchmarks.}
    \label{fig:comparision}
\end{figure}

On the other hand, established and systematic psychological research provides a wealth of rigorously validated task paradigms encompassing diverse cognitive dimensions. These established measures range from the Tests of Emotion Comprehension (TEC)~\cite{Task-TEC} for affective understanding and the Yummy-yucky Task~\cite{Task-YT} for probing subjective preferences, to the See-Know Task~\cite{Task-SkT} for analyzing perception-knowledge links, and Faux Pas Recognition~\cite{Task-FRT} for interpreting non-literal communication within complex social contexts.

\begin{figure*}[t]
    \centering
    \includegraphics[width=1.0\linewidth]{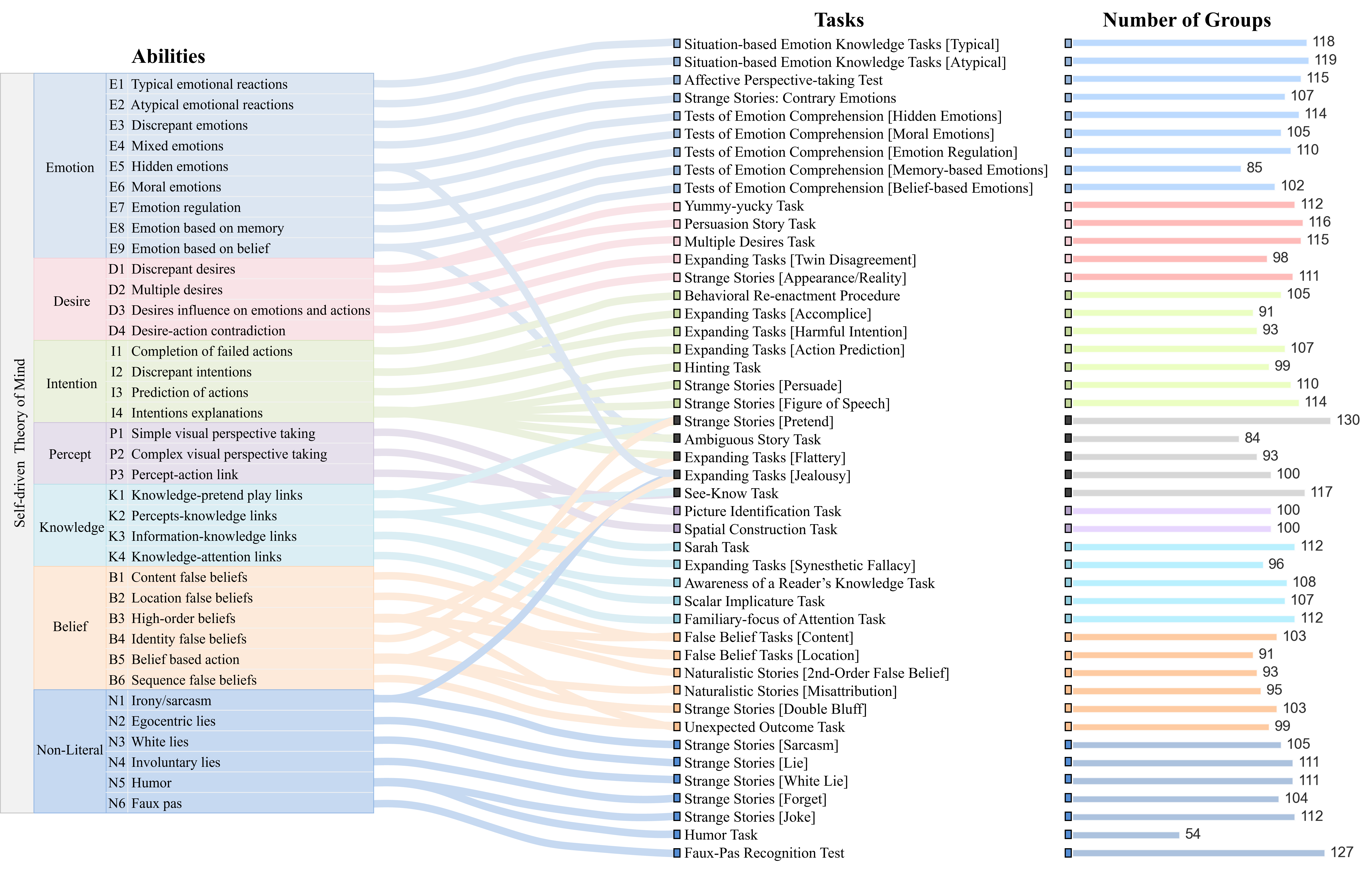}
    \caption{Overview of the CogToM framework.}
    \label{fig:sankey}
\end{figure*}

Consequently, to overcome the structural limitations of existing ToM benchmarks for LLMs, we draw upon these mature and extensively validated psychological paradigms to construct an entirely new evaluation dataset. Figure~\ref{fig:comparision} visualizes the breadth of ToM task coverage, delineating the expanded scope of our dataset relative to previous evaluation frameworks.

\section{CogToM Framework}
We established CogToM, a human-cognition-inspired ToM evaluation framework for LLMs. This process involved summarizing psychological ToM paradigms and rewriting them into a standardized ``scene-based multiple-choice'' format suitable for LLMs. Combined with original assessment tasks and subjected to rigorous supervision and annotation by 49 individuals, the final benchmark encompasses 46 tasks and 8,513 data entries.

\subsection{Task Design}

Strictly adhering to the core definition of ToM (the ability to model the mental states of others and to distinguish them from one's own), we established standards for task design, adaptation, and original creation. This process yielded 46 ToM tasks in a multiple-choice format (54-130 data groups per task, as shown in the right panel of Fig.\ref{fig:sankey}), such as \textit{False Belief Tasks [Location]}, \textit{Scalar Implicature Task}, \textit{Strange Stories [Lie]}, and \textit{Test of Emotion Comprehension [Hidden Emotions]}(square brackets denote categories subdivided based on internal variations within the original paradigms). Furthermore, we mapped these tasks to ToM cognitive capabilities (referencing the ATOMS classification~\cite{atoms}), summarizing them into 7 capability categories and 36 sub-capabilities (see Fig.\ref{fig:sankey} and Appendix~\ref{app:36subability}). 
We summarize the principal contributions of our assessment framework as follows.

\begin{figure*}[htbp]
    \centering
    \includegraphics[width=1.0\linewidth]{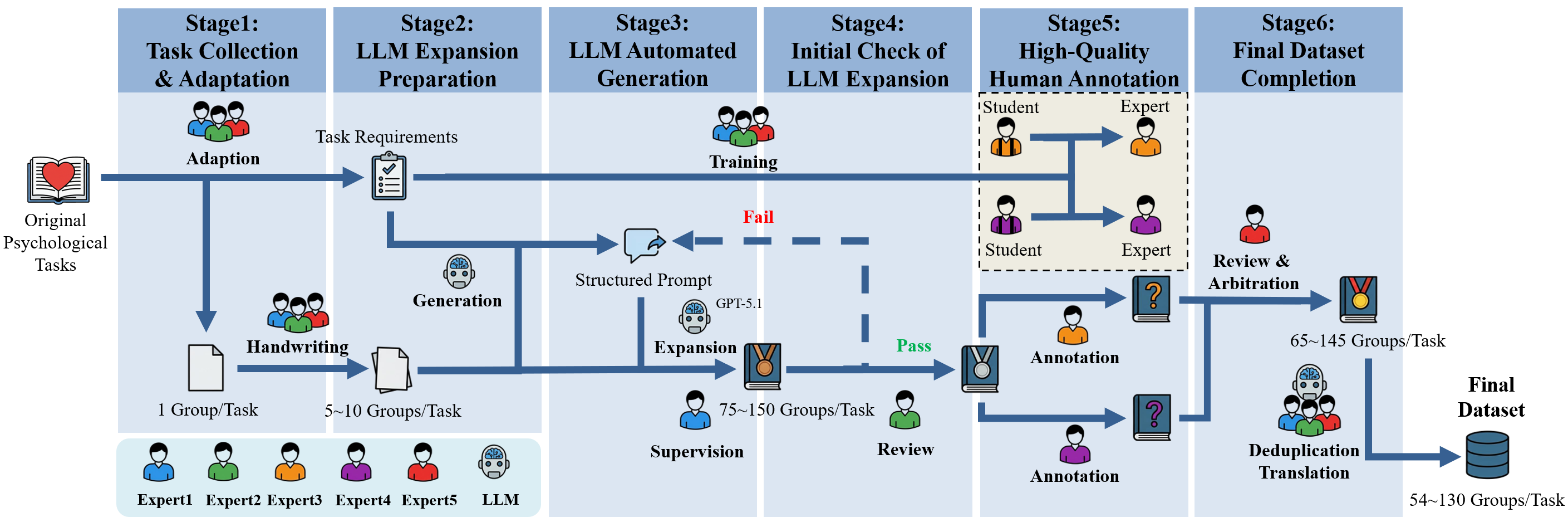}
    \caption{The data construction pipeline of CogToM.}
    \label{fig:dataset_construct_pipeline}
\end{figure*}

\noindent \textbf{Refinement of Classic Assessment Tasks.} For example, we specifically revised the numerical standards for options in the \textit{Scalar Implicature Task} and the story logic of \textit{Strange Stories [Double Bluff]}. Furthermore, for the \textit{Unexpected Outcome Task}, we shifted the assessment focus from atypical emotions to sequence false beliefs (see Table.\ref{tab:k4_example},\ref{tab:b5_example},\ref{tab:b6_example}). 
    
\noindent \textbf{Enhancement of Assessment Depth and Breadth.} For instance, the newly introduced \textit{Naturalistic Stories [2nd-Order False Belief]} delves into higher-order beliefs and the revision of 2nd-Order false beliefs (see Table.\ref{tab:b3_example}). Additionally, we adapted or created 5 comprehensive tasks that span diverse capabilities. Notably, the \textit{Strange Stories [Pretend]} incorporates a contrast between pretend play and identity false beliefs (see Table.\ref{tab:c1_example}).
    
\noindent \textbf{Four Original Assessment Paradigms.} For instance, \textit{Expanding Tasks [Synesthetic Fallacy]} (see Table.\ref{tab:k2_example}), inspired by ``The Blind Men and the Elephant,'' evaluates whether the model understands that relying on a single sense to perceive multi-sensory objects can lead to cognitive misconceptions.
    
\noindent \textbf{Examination of Spatial Perspective Reconstruction.} The \textit{Spatial Construction Task} focuses on evaluating the model's perspective taking capabilities, specifically its ability to imagine perceptual information from another individual's viewpoint (see Table.\ref{tab:p2_example}).

\subsection{Construction of the CogToM Dataset}

The construction of our dataset proceeds through 6 stages, as illustrated in Fig.~\ref{fig:dataset_construct_pipeline}. Throughout this process, each data entry underwent at least 5 rounds of human supervision, simultaneously yielding structured, high-quality task requirement specifications.

\subsubsection{Data Collection}

\noindent {\bf Task Collection and Adaptation.} Based on psychological ToM assessment paradigms, we adapted one group of LLM evaluation data, wherein each data entry consists of one scene and 1-5 multiple-choice questions:

\begin{itemize}[leftmargin=*, nosep]

    \item \textbf{Scene} situated within diverse social settings, the content was reconstructed based on original psychological assessment materials and is presented in a narrative format.
    
    \item \textbf{Question} examines specific aspects of the scene. Within a single task, the number of questions per group is fixed, and questions with the same index maintain consistent inquiry formats. This design enables the assessment of the identical ToM capability across diverse social contexts. Each question corresponds exclusively to one ToM sub-capabilities. 
    
    \item \textbf{Options} serves as an extension of the question, each item consists of four options, only one of which is correct. Among incorrect choices, distractors that closely resemble the correct option are included.
    
\end{itemize}

\subsubsection{LLM-based Question Expansion}

\noindent {\bf LLM Expansion Preparation.} Building upon Stage 1, our expert team expanded each task to 5–10 data groups and formulated structured task specifications for the subsequent stages.

\noindent {\bf LLM Automated Generation.} We conducted preliminary interactions with LLMs to transform the outputs from Stage 2 into structured prompts (examples provided in Appendix~\ref{app:data_expansion} and Table.\ref{tab:exprompt}). These prompts were then input into GPT-5.1 to generate expanded data, ultimately resulting in 75–150 Chinese data groups per task. Expert 1 supervised this process, primarily verifying the formatting and ensuring that the generated scenes and questions met the task assessment requirements.

\noindent {\bf Initial Check of LLM Expansion.} Expert 2 reviewed the supervision results and additionally assessed the richness of the scenes. If the data quality was deemed substandard, the process reverted to Stage 3 for prompt regeneration. The roles of Expert 1 and Expert 2 were assumed by 6 core members of the research team on a rotating basis.

\subsubsection{Human Annotation and Data Validation}

\noindent {\bf High-quality Human Annotation.} 42 graduate students (each paid with \$30) specializing in Philosophy and Artificial Intelligence received training as detailed in Table.\ref{tab:exprompt} and Table.\ref{tab:e1_example}-\ref{tab:c5_example}. Subsequently, they rotated in the roles of Expert 3 and Expert 4 to perform a double-blind annotation of the outputs from Stage 4, focusing on answers and quality of the questions. Statistical data regarding the annotation process are presented in Appendix~\ref{app:human_anno} and Table.\ref{tab:agrate}.

\noindent {\bf Final Dataset Completion.} Expert 5 collected the annotation results. For data instances exhibiting discrepancies between human and answers by GPT-5.1 in Stage 3, or those flagged as ``quality issues present'' by annotators, Expert 5 consulted with the responsible Expert 3 and Expert 4 to revise or discard the items. Following further de-duplication, 54–130 data groups remained for each task. Finally, the dataset was translated into English using the Baidu API, supplemented by manual verification.

\section{Experimental Results and Analysis}

\subsection{Experimental Setups}

\subsubsection{Evaluated Models}

We comprehensively evaluated a total of 22 representative models, spanning from early versions released in July 2023 to the most recent state-of-the-art systems, covering multiple well-known open source and closed source series, including: 
GPT-3.5-Turbo~\cite{gpt-3.5-turbo}, GPT-4o-2024-11-20~\cite{gpt-4o}, GPT-4o-mini~\cite{gpt-4o-mini}, GPT-5.1~\cite{gpt-5.1},
Llama-2-7B-Chat~\cite{llama-2}, Llama-2-13B-Chat~\cite{llama-2}, Llama-3-8B-Instruct~\cite{llama-3}, Llama-3.1-8B-Instruct~\cite{llama-3.1},
Mistral-7B-Instruct-v0.1~\cite{mistral7b}, Mixtral-8x7B-Instruct-v0.1~\cite{mixtral},
Qwen-7B-Chat~\cite{qwen}, Qwen1.5-7B-Chat~\cite{qwen1.5}, Qwen2-7B-Instruct~\cite{qwen2}, Qwen2.5-7B-Instruct~\cite{qwen2.5}, Qwen2.5-72B-Instruct~\cite{qwen2.5}, Qwen3-235B-A22B-Instruct~\cite{qwen3}, Qwen3-Max~\cite{qwen3max}, Qwen3-Next-80B-A3B-Instruct~\cite{qwen3next},
DeepSeek-v3~\cite{ds-v3}, DeepSeek-v3.2~\cite{ds-v3.2}, Grok-4-Fast~\cite{grok4fast}, and Kimi-k2-0905~\cite{kimi-k2}.

\subsubsection{Evaluation Methods and Metrics}

We evaluate all the models using a zero-shot vanilla prompt, requiring them to output answers directly in the format of ``[[Option Letter]]'' (see Appendix~\ref{app:eval_prompt} for the full bilingual prompts). To ensure deterministic outputs and strict adherence to formatting requirements, we set the generation temperature to 0 for all models. To mitigate the impact of positional bias, each question is tested 5 times, comprising four cyclic rotations of the options and one additional random shuffle distinct from the rotations. The average accuracy across these five trials serves as the primary evaluation metric. Given that the ground-truth labels in our dataset were established through multiple rounds of manual annotation and expert review, the accuracy also reflects the consistency rate between model selections and human expert judgments.

\subsection{Main Results}

Here we present the key experimental findings and provide a comprehensive analysis of our results. Given that the performance disparity between the Chinese and English datasets is marginal, the results reported herein represent the aggregated mean across both languages unless otherwise specified. Detailed comparisons of cross-lingual performance variations are available in Appendix~\ref{app:zh_en_diff}.

\subsubsection{Temporal Evolution and Scaling Dynamics}

\begin{figure}[htbp]
    \centering
    \includegraphics[width=1.0\linewidth]{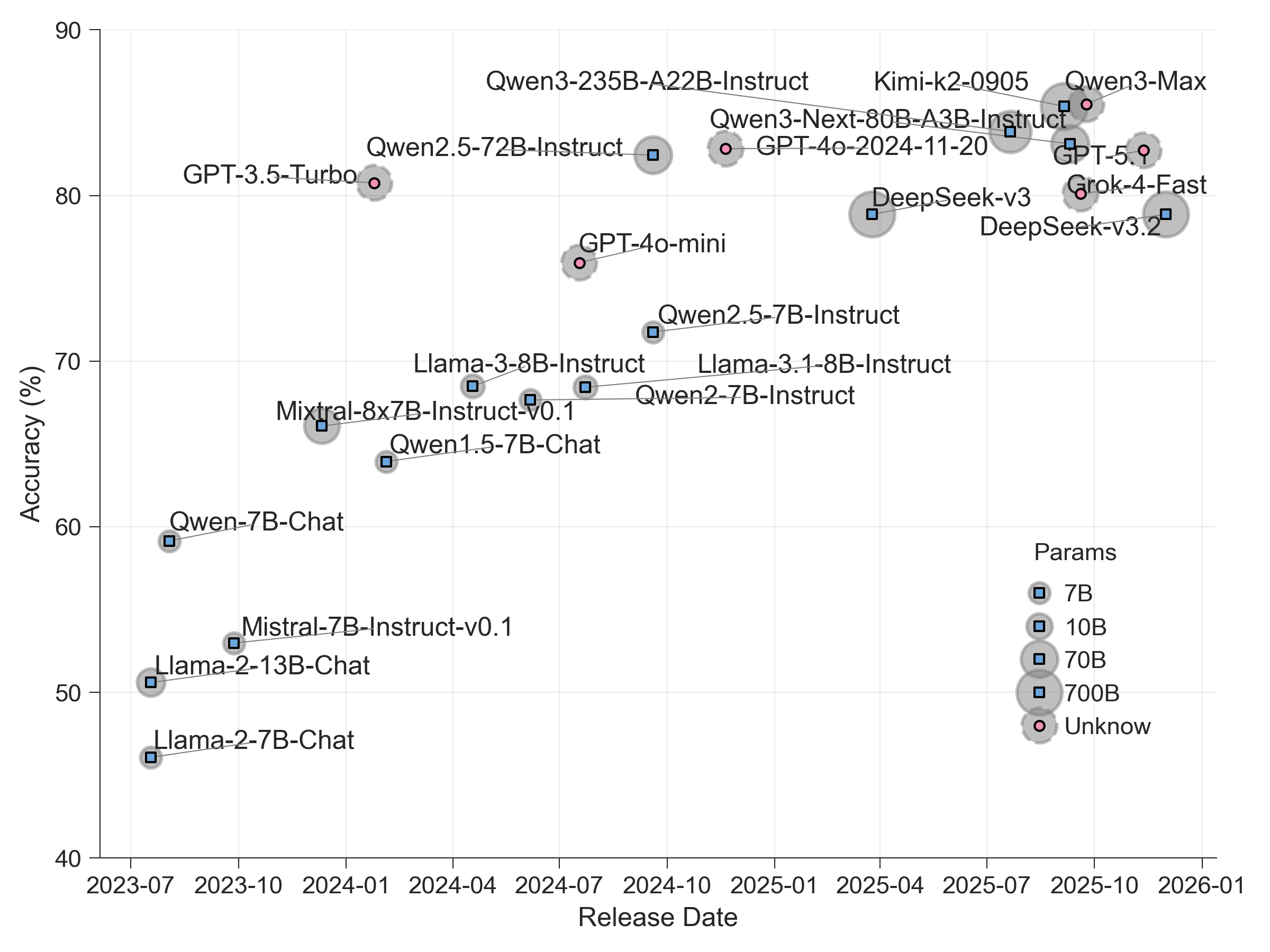}
    \caption{Overall average performance of open-source and closed-source models across various release dates, scales and families.}
    \label{fig:time_acc}
\end{figure}

We begin by analyzing the average performance of various models across bilingual tests for all different tasks. As illustrated in Figure~\ref{fig:time_acc}, there is a pronounced upward trajectory in model capabilities over time. While early mainstream models, such as the Llama-2 series, exhibited accuracies within the 45\%–55\% range, frontier models by late 2025 (e.g., Qwen3-Max and GPT-5.1) have successfully surpassed 80\%.

Consistent with the intuition provided by scaling laws~\cite{kaplan2020scaling}, we observe that within the same model family and release window, larger parameter counts directly correlate with higher accuracy (e.g., Llama-2-13B vs. 7B, and Qwen2.5-72B vs. 7B). Concurrently, a significant leap in parameter efficiency is evident. For instance, Qwen2.5-7B outperforms earlier, substantially larger architectures such as Llama-2-13B and Mixtral-8x7B. Furthermore, the performance gap between open-source and proprietary models is narrowing. Frontier open-source models exemplified by Qwen3-235B, now demonstrate accuracy levels that are comparable to, or even exceed, those of GPT-5.1.

\subsubsection{Differences Across ToM Cognitive Dimensions}

\begin{figure}[htbp]
    \centering
    \includegraphics[width=1.0\linewidth]{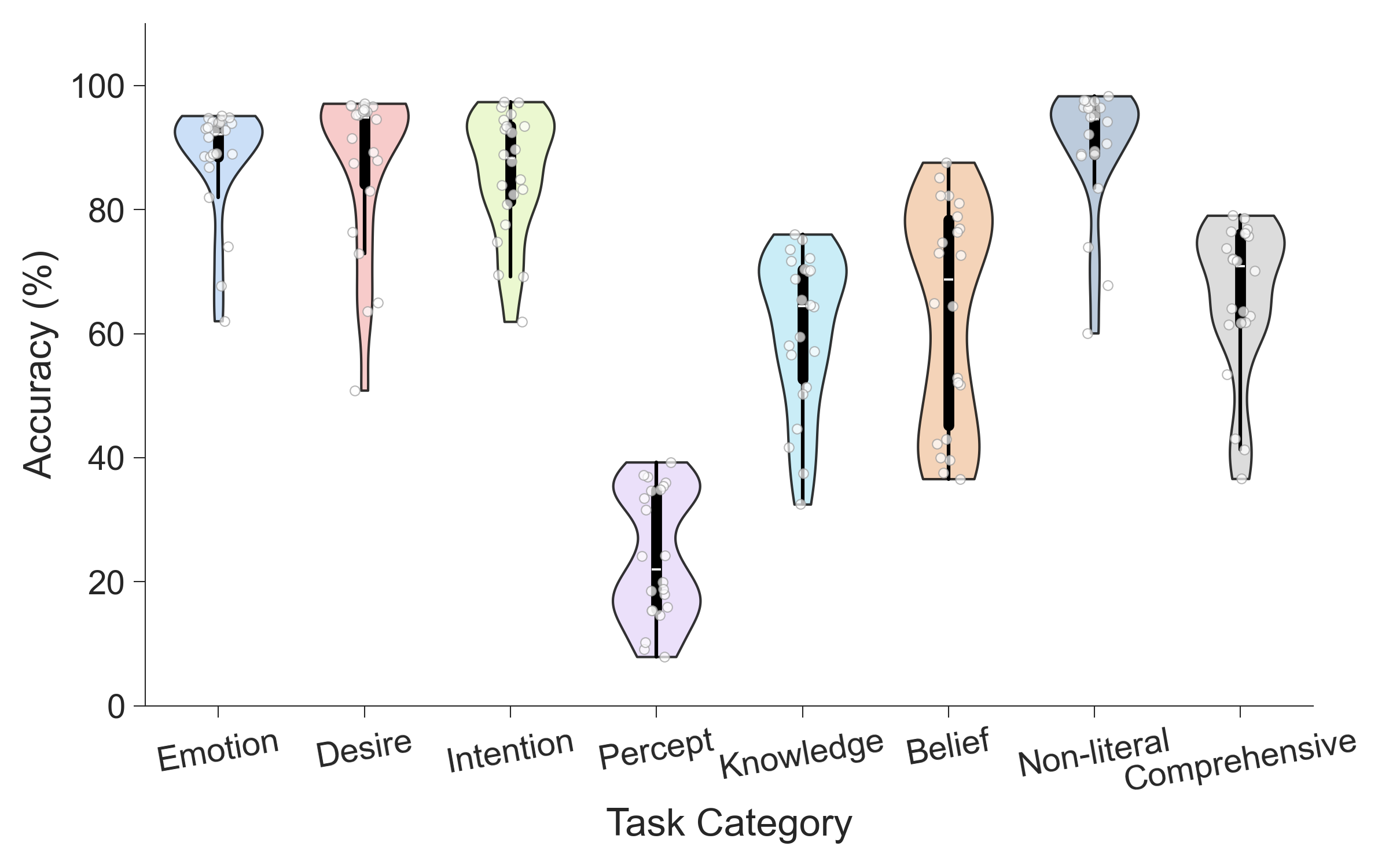}
    \caption{Accuracy distribution of models across different ToM task categories.}
    \label{fig:violin}
\end{figure}

We delve deeper into the specific capabilities of models across 8 primary categories. As illustrated in Figure~\ref{fig:violin}, LLMs exhibit significant heterogeneity in performance across different ToM task categories. Models demonstrate near-ceiling performance in \textbf{Emotion}, \textbf{Desire}, and \textbf{Non-literal} reasoning tasks, with the majority of data points clustered between 80\% and 95\%. This suggests that modern LLMs have attained a high degree of proficiency in interpreting social-emotional context and linguistic nuances.

In contrast, the \textbf{Percept} category emerges as a critical performance bottleneck for all tested models, with a median accuracy of only approximately 20\%. This deficiency highlights the models' inherent struggle in resolving perspective disparity between the self and others, as well as their inability to perform robust perspective-taking to infer other's observation of a shared environment. 
Furthermore, the high variance can be observed in \textbf{Belief} and \textbf{Knowledge} tasks, characterized by elongated violin bodies. This distribution implies that models have varying abilities in inferring the beliefs held by others or whether they can acquire certain knowledge, and a substantial gap remains between top-tier and mid-tier architectures. 
In conclusion, while LLMs demonstrate near-human proficiency in socio-affective semantic understanding, they continue to face fundamental hurdles in foundational cognitive reasoning linked to physical perception.

\begin{figure*}[htbp]
    \centering
    \includegraphics[width=1.0\linewidth]{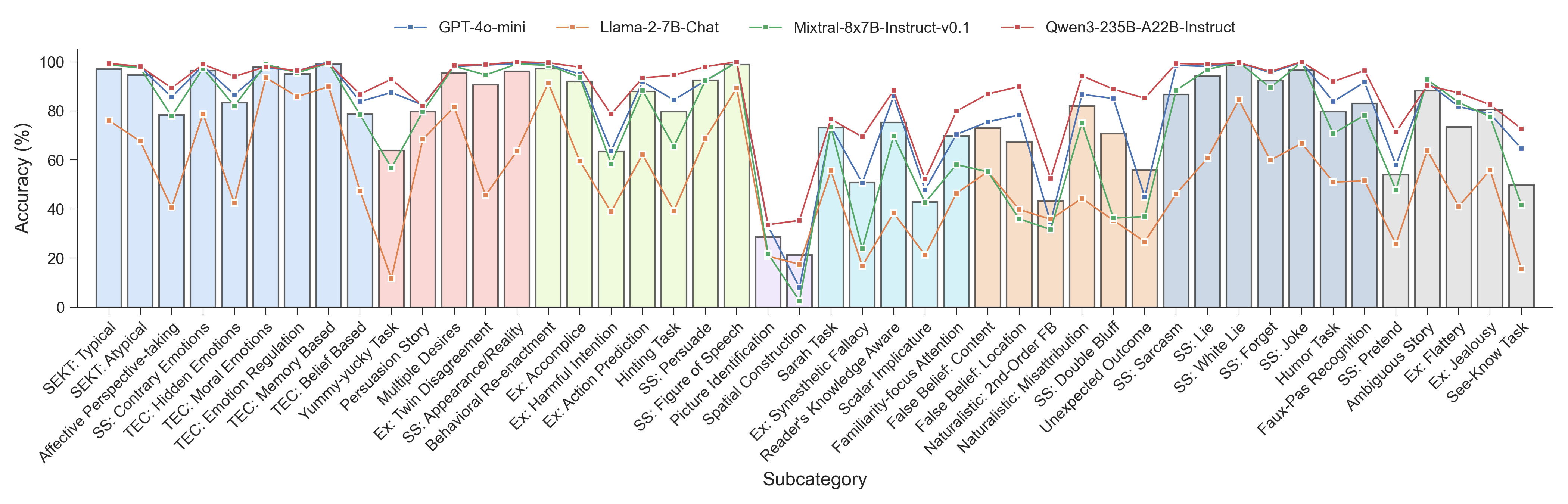}
    \caption{LLMs' performance across 46 tasks. Color of Bars indicate their respective primary categories.}
    \label{fig:subcat_bar_line}
\end{figure*}

\subsubsection{Detailed Comparison across 46 Tasks}
\label{sec:result_46_tasks}

We further examined their performance across 46 tasks. As illustrated in Figure~\ref{fig:subcat_bar_line}, the bars represent the aggregate average accuracy of all tested models. The overlaying line plots depict the performance trajectories of four representative models selected to span a wide spectrum of capabilities: Llama-2-7B, Mixtral-8x7B, GPT-4o-mini, and Qwen3-235B. The comprehensive performance trajectories for all evaluated models are provided in the Appendix~\ref{app:full_results_46task}. These results reveal a significant intra-category heterogeneity in model performance, while the distinct trajectories of individual models further highlight the performance gaps and unique capability profiles of the evaluated models.

While the \textbf{Desire} category (red bars) demonstrates high aggregate performance, granular analysis uncovers a notable imbalance. Most models achieve near-ceiling accuracy in tasks like \textit{Multiple Desires} (average accuracy approximately 95\%), yet exhibit a marked degradation in the \textit{Yummy-yucky Task} (average accuracy around 60\%). Specifically, Mixtral-8x7B's accuracy drops from near 100\% in Multiple Desires to approximately 65\% in the \textit{Yummy-yucky Task}, while Llama-2-7B's performance plummets to below 20\%.

In the \textbf{Belief} domain (orange bars), accuracy for second-order false belief and \textit{Unexpected Outcome Tasks} is substantially lower than that for first-order tasks (about 15\%), suggesting that increased cognitive complexity poses a significant hurdle. Within the \textbf{Knowledge} category (cyan bars), the performance on \textit{Reader's Knowledge Aware} is significantly higher, with an average accuracy of over 70\%, compared to the \textit{Ex: Synesthetic Fallacy} task, where the average accuracy is around 50\%. This disparity suggests that the models' proficiency in tracking a reader's knowledge may stem from stylistic pattern matching or linguistic heuristics learned from explanatory corpora~\cite{ullman2023large, shapira2024clever}, rather than genuine ToM capabilities. Conversely, the failure to resolve \textit{Ex: Synesthetic Fallacy}, which involve foundational sensory common sense, underscores the models' inherent difficulty in representing and reasoning about others' perceptual states.

\subsubsection{Discriminative Power of New Tasks}
\label{sec:result_new_tasks}

Following our analysis of task performance variance, we observed significant performance degradations in specific tasks within the same primary ability categories. Notable examples include \textit{Yummy-yucky Task} in \textbf{Desire} , \textit{Ex: Harmful Intention} in \textbf{Intention}, and \textit{Ex: Synesthetic Fallacy} in \textbf{Knowledge}, etc. As illustrated in Figure~\ref{fig:comparision}, these challenging tasks are the new task paradigms introduced in this study as opposed to existing benchmarks. Consequently, we partition the dataset into ``Existing'' and ``New'' categories to analyze the distribution of average item difficulty, defined as the mean accuracy across all evaluated models for each individual question. A comprehensive list of ``new tasks'' is provided in Appendix~\ref{app:def_new_tasks}.

\begin{figure}[htbp]
    \centering
    \includegraphics[width=1.0\linewidth]{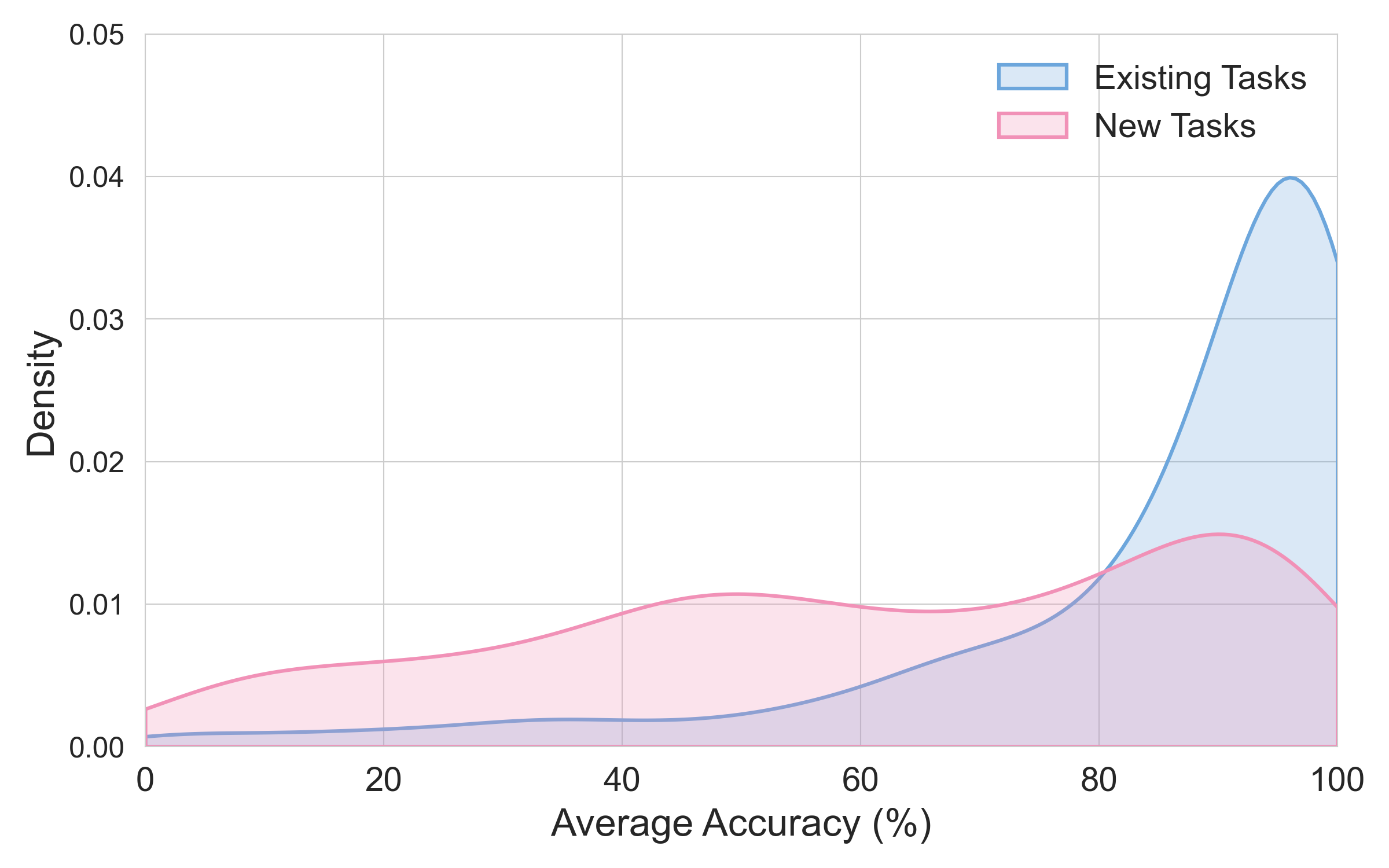}
    \caption{Density distribution of average model accuracy for each question across existing and new tasks.}
    \label{fig:acc_density}
\end{figure}

As illustrated in Figure~\ref{fig:acc_density}, the existing tasks exhibit a pronounced ceiling effect, with density peaks heavily clustered around the 90\%–100\% accuracy range. This suggests that traditional ToM benchmarks are approaching saturation for current frontier models. In contrast, our newly proposed tasks display a significantly broader and more balanced distribution, with a substantial portion of questions falling within the 0\%–60\% low-to-mid accuracy bracket. This structural shift demonstrates that CogToM offers enhanced discriminative power and provides a more granular assessment of multi-faceted, sophisticated Theory of Mind capabilities.

\subsection{Indepth Analysis}


\subsubsection{Joint Analysis of Inter-annotator Agreement Rate and Model Accuracy}

\begin{figure}[htbp]
    \centering
    \includegraphics[width=1.0\linewidth]{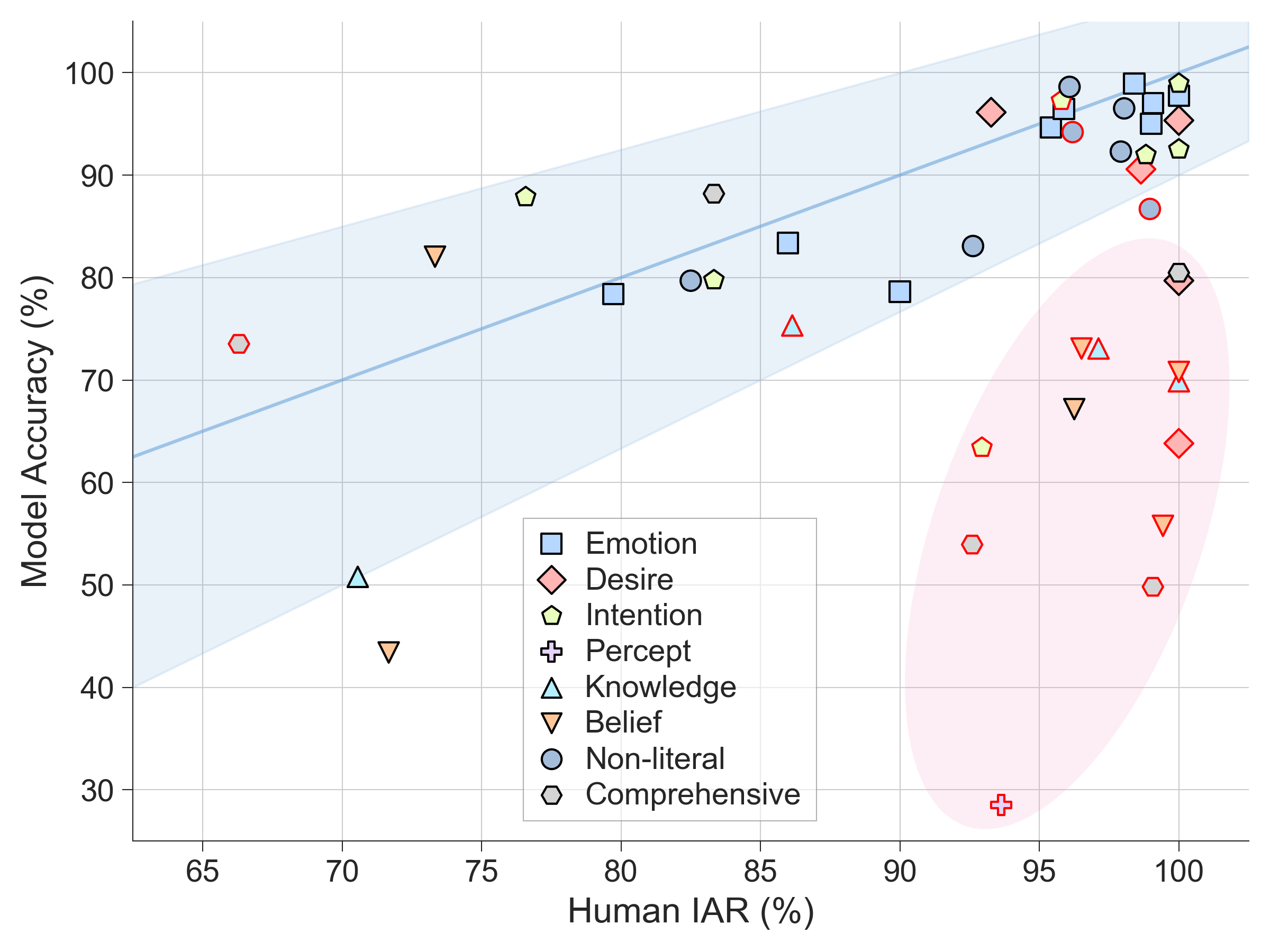}
    \caption{Correlation between human Inter-annotator Agreement Rate (IAR) and average model accuracy across different tasks.}
    \label{fig:iar_acc}
\end{figure}

To investigate the correlation between LLM performance and human cognitive patterns, we conducted a joint analysis of average model accuracy and human inter-annotator agreement rate (IAR) across tasks. IAR, defined as the percentage of questions where two experts reach consensus when annotating answers, serves as a metric for task difficulty for humans or its semantic ambiguity. As illustrated in Figure~\ref{fig:iar_acc}, each data point represents a specific task, with its marker shape and color denoting its corresponding ability category.

We observe that most data points cluster within the $y=x$ alignment zone, indicating a high degree of performance alignment between LLMs and human experts across these dimensions. Conversely, the red elliptical region in the bottom-right corner reveals a profound cognitive asymmetry. While these tasks exhibit near-perfect human consensus (IAR > 90\%) and are considered unambiguous ``common sense'' by experts, average model accuracy significantly degrades to the 30\%–80\% range. Especially, Percept tasks (indicated by purple crosses) reside at the very bottom. Despite 100\% human agreement, model performance is abysmal, below 30\%. This disparity underscores the presence of Moravec’s Paradox~\cite{moravec1988mind} within LLM cognitive architectures. Notably, the vast majority of points within this red region correspond to our newly introduced task paradigms (highlighted with red outlines), demonstrating that CogToM effectively exposes critical vulnerabilities and delineates the true cognitive boundaries of LLM Theory of Mind.

\subsubsection{Assessment of Alignment with Human Developmental Milestones}

Psychological research indicates that the development of ToM in human children follows distinct, staged milestones. Specifically, children typically acquire basic subjective preference differentiation first, followed by the ability to judge others' knowledge states based on perceptual cues, and finally, sophisticated reasoning regarding beliefs and complex emotions~\cite{wellman2004scaling}. We aim to investigate whether LLMs exhibit developmental trajectories analogous to these human cognitive patterns. To this end, we have carefully selected a sequence of tasks that closely align with the established chronological developmental milestones of human ToM.

\begin{figure}[h]
    \centering
    \includegraphics[width=1.0\linewidth]{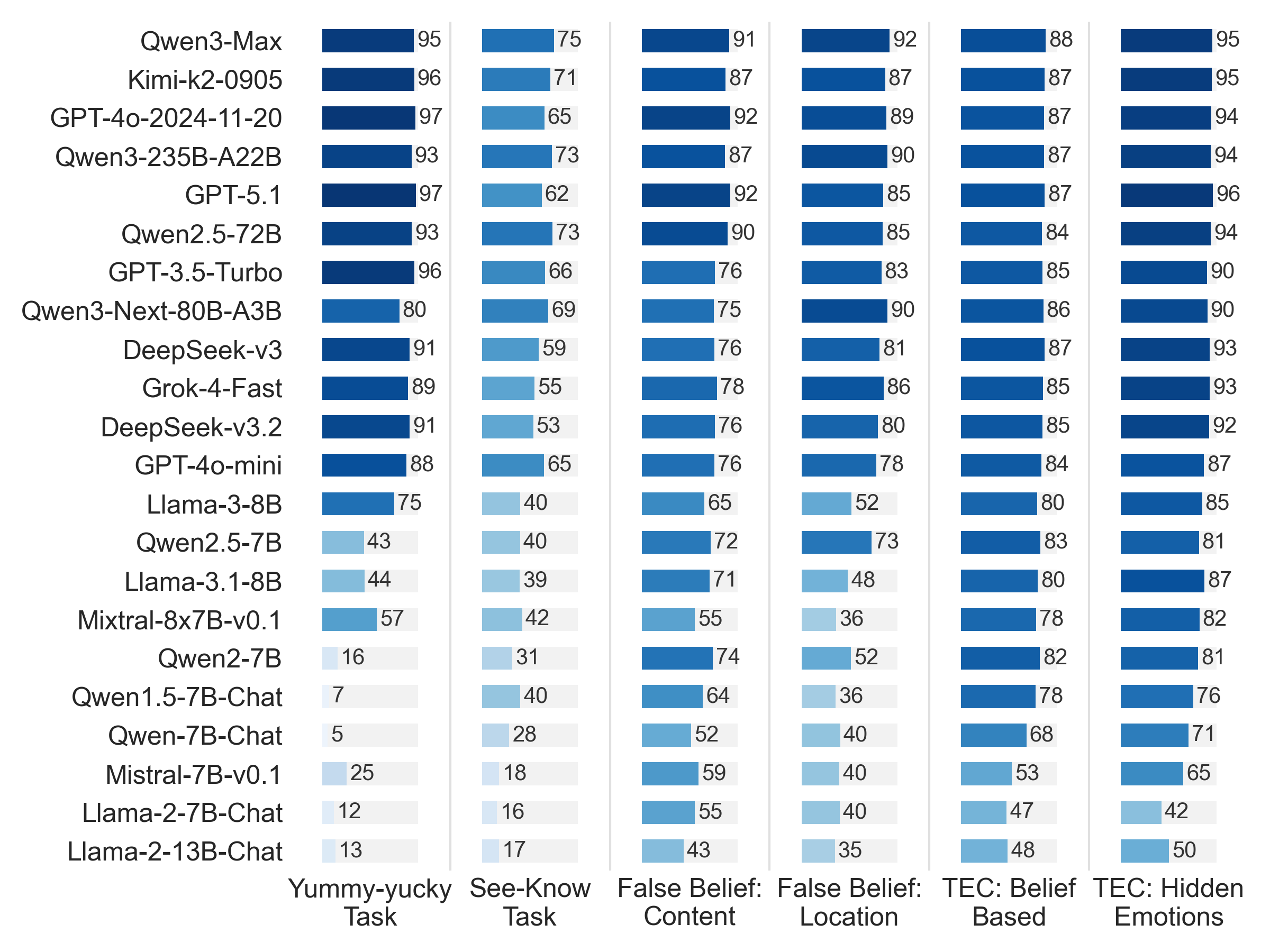}
    \caption{Accuracy of LLMs across the developmental sequence of ToM in human children.}
    \label{fig:develop_seq}
\end{figure}

As illustrated in Figure~\ref{fig:develop_seq}, selected task are arranged chronologically from left to right (approx. 0–6 years), ranging from early-stage tasks like the \textit{Yummy-yucky Task} to late-stage challenges such as \textit{TEC: Hidden Emotions}. The bars indicate the accuracy of each model for the respective tasks. 

Our experimental results reveal a striking ``developmental inversion'' in a considerable portion models. These models demonstrate near-human proficiency in late-acquired emotional reasoning, yet paradoxically fail at the most elementary sensory preference tests. For instance, Qwen-7B-Chat exhibits accuracies of 5\%, 28\%, 52\%, 40\%, 68\%, and 71\% across the sequenced tasks, demonstrating a pronounced upward trajectory. Besides, even frontier models with superior overall performance exhibit an anomalous performance dip in the \textit{See-Know Task}, a milestone typically mastered in early childhood. For example, GPT-5.1 achieves only 62\% accuracy on this task, a significant degradation compared to its near-perfect 96\% on the more complex \textit{ TEC: Hidden Emotions task}.

This suggests that LLMs likely achieve a form of ``simulated ToM'' through linguistic pattern matching and probabilistic prediction derived from massive corpora, rather than through a cognitive developmental process grounded in perception and embodiment as seen in humans. Ultimately, our analysis highlights the presence of Moravec’s Paradox in LLM cognitive architectures: high-level cognitive capabilities are relatively accessible through scaling, whereas low-level perceptual reasoning remains challenging to replicate.



\section{Conclusion}


In this paper, we present \textbf{CogToM}, a theoretically grounded benchmark that integrates 46 task paradigms. It offers comprehensive task coverage and demonstrates robust discriminative power. Experimental results suggest that model performance is heterogeneous across different cognitive dimensions. Further analysis point toward potential divergences between machine intelligence and human cognitive patterns. In summary, CogToM offers a practical instrument and a new perspective for further investigating the cognitive boundaries of Theory of Mind in LLMs.

\section*{Limitations}

Despite its comprehensive scope, several limitations of our work should be acknowledged.

\noindent \textbf{Linguistic and Cultural Scope:} Currently, the benchmark is limited to Chinese and English. Considering that ToM reasoning is deeply intertwined with linguistic structures and cultural norms, future work should expand to a broader range of languages and cultural contexts.

\noindent \textbf{Modality Constraints:} Many classic ToM evaluation paradigms, such as yoni task~\cite{yonitask} and animated triangle task~\cite{triangletask}, inherently rely on visual or dynamic cues. While we adapted these into textual descriptions, this transition inevitably results in a loss of ecological validity compared to original psychological assessments.

\noindent \textbf{Evaluation Paradigm:} CogToM primarily utilizes a multiple-choice format. While this ensures objective scoring, it may not fully capture the generative and nuanced nature of ToM in open-ended social interactions.

\noindent \textbf{Static vs. Interactive Inference:} Our tasks consist of static scenes. In real-world settings, ToM involves the recursive and dynamic updating of mental states during live interaction, a dimension that current single-turn evaluations cannot fully replicate.

\section*{Acknowledgments}

The authors acknowledge the use of large language models (LLMs) as writing assistants to refine grammar and improve phrasing. These models were used solely for linguistic editing and did not contribute to the research idea, experimental design, or data analysis. The authors take full responsibility for the correctness and integrity of the content.

\bibliography{custom}

\appendix

\definecolor{Emotion}{RGB}{197, 224, 245}
\definecolor{NonLiteral}{RGB}{208, 216, 240}
\definecolor{Intention}{RGB}{226, 240, 217}
\definecolor{Belief}{RGB}{252, 228, 214}
\definecolor{Comprehensive}{RGB}{217, 217, 217}
\definecolor{Knowledge}{RGB}{205, 235, 245}
\definecolor{Desire}{RGB}{244, 204, 204}
\definecolor{Percept}{RGB}{230, 220, 235}

\section{Construction and Experiment Details}

\subsection{Prompts for Data Expansion}
\label{app:data_expansion}

\begin{CJK*}{UTF8}{gbsn}
\begin{table*}[t!] 
\footnotesize
    \centering      
    
    \begin{tabular}{|p{0.95\linewidth}|}
        \hline
        你是一个AI助手，需要生成基于心理理论（Theory of Mind）的情绪问题。请严格参照以下说明和示例，生成至少10组问题。每组问题包括一个情境和两个问题（问题编号1和2），询问两个不同人物的情绪。情绪选项是四个两字词语，对于每个问题，答案指定正确选项（A、B、C或D）。输出必须是一个表格，包含以下列：情境、问题编号、问题、A、B、C、D、答案。表格应以Json格式呈现，但不包括第一行说明。 \\

        \\

        \#\#\# 说明：\\
        下面这些问题都有这样的共同点：（丙做了）一件事，对甲和乙造成了不同影响。因此，（丙做的）这件事情让甲和乙产生了相反的情绪。你需要设计两个问题分别询问甲和乙的情绪，两个问题选项相同，除去各自的正确答案外还有两个无关答案。每个答案是一个描述情绪的两字词语。生成问题的时候不能直接用“甲”“乙”“丙”，而要取好名字，不同问题之间用的名字有所区分。约一半的问题涉及到第三方“丙”的出现，另一半问题仅涉及自然发生的事件。\\

        \\

        \#\#\# 示例问题：\\
        | 情境 | 问题编号 | 问题 | A | B | C | D | 答案 |\\
        |------|----------|------|---|---|---|---|------|\\
        | 小红本来应该去帮助她的俱乐部为这次活动做准备，但她却去看望了一个朋友。 | 1 | 小红的朋友会有怎样的心情？ | 生气 | 自豪 | 感激 | 后悔 | C |\\
        | 小红本来应该去帮助她的俱乐部为这次活动做准备，但她却去看望了一个朋友。 | 2 | 俱乐部成员会有怎样的心情？ | 生气 | 自豪 | 感激 | 后悔 | A |\\
        | 小丽的男朋友邀请小丽的闺蜜小芳去看一个浪漫电影。 | 1 | 小丽会有怎样的心情？ | 生气 | 开心 | 悲伤 | 尴尬 | A |\\
        | 小丽的男朋友邀请小丽的闺蜜小芳去看一个浪漫电影。 | 2 | 小芳会有怎样的心情？ | 生气 | 开心 | 悲伤 | 尴尬 | D |\\
        | 小飞最近在遭遇悲惨事故后住院，所以小刚接小飞的女朋友出去看望他。 | 1 | 小飞会有怎样的心情？ | 兴奋 | 感动 | 担忧 | 后悔 | B |\\
        | 小飞最近在遭遇悲惨事故后住院，所以小刚接小飞的女朋友出去看望他。 | 2 | 小飞的女朋友会有怎样的心情？ | 兴奋 | 感动 | 担忧 | 后悔 | C |\\
        | 公司经理宣布小明获得了晋升，而小李对这次晋升期待已久。 | 1 | 小明会有怎样的心情？ | 高兴 | 生气 | 嫉妒 | 期待 | A |\\
        | 公司经理宣布小明获得了晋升，而小李对这次晋升期待已久。 | 2 | 小李会有怎样的心情？ | 高兴 | 生气 | 嫉妒 | 期待 | D |\\
        | 喜欢极限运动的小明拉着小红坐过山车，过程中小明一直大声尖叫，而小红则紧闭双眼。 | 1 | 小明会有怎样的心情？ | 兴奋 | 惊讶 | 害怕 | 后悔 | A |\\
        | 喜欢极限运动的小明拉着小红坐过山车，过程中小明一直大声尖叫，而小红则紧闭双眼。 | 2 | 小红会有怎样的心情？ | 兴奋 | 惊讶 | 害怕 | 后悔 | C |\\

        \\
        
       \#\#\# 要求：\\
        - 生成至少10组新问题（即至少10个情境，每个情境对应两个问题行，总共至少20行）。\\
        - 情境描述应类似示例的句式和篇幅（1-2句话）。\\
        - 问题句式统一为：“[人物名字]会有怎样的心情？”\\
        - 选项为四个两字情绪词（如：生气、高兴、悲伤等），对于每个情境，两个问题共享四个选项，但是答案不同。\\
        - 人物名字不能重复使用示例中的名字，且不同情境间名字要区分。\\
        - 超过70\(\%\)的情境涉及第三方“丙”（如第一组示例问题中的小红，第二组问题中的小丽的男朋友，第三组问题中的小刚）。\\
        - “甲”和“乙”必须同时出现在情境中。\\
        - 输出表格必须包含列：情境、问题编号、问题、A、B、C、D、答案。\\

        \\

        现在，请直接输出生成的表格（Json格式）。\\
        \hline
    \end{tabular}
    \caption{An example of the prompt used for the expansion of \colorbox{Emotion}{Affective Perspective-taking Test}.}
    \label{tab:exprompt}
\end{table*} 
\end{CJK*}

In Stage 3 of dataset construction, we utilized GPT-5.1 to expand the total number of scenes for 44 tasks. Taking \textit{Affective Perspective-taking Test} as an example, we present the prompt used in the generation process, as shown in Table \ref{tab:exprompt}. 

It is worth noting that during the expansion of the \textit{Scalar Implicature Task} and the \textit{Spatial Construction Task} using LLMs, three consecutive batches of generated results failed to pass the joint review by Expert 1 and Expert 2 due to fundamental and pervasive quality defects. Consequently, we ultimately did not employ Stages 3–6 described in the main text to expand the data for these two tasks.

In the expansion process of the \textit{Scalar Implicature Task}, while the story backgrounds generated by the LLM met the requirements, the model failed to produce satisfactory numerical values for the options. Consequently, we retained over 100 scenes generated by the LLM. Expert 1 then individually reviewed each scene to revise the specific numerical values within the stories and further adjusted the option settings and correct answers according to our established standards. These results were finalized after verification by Expert 2.

In the expansion process of the \textit{Spatial Construction Task}, the LLM was entirely unable to extend data with identical structures based on the provided examples. Therefore, using the 5 groups of example data as templates, we prompted the LLM to randomly generate object nouns and employed a batch processing script to replace the original nouns in the examples with these keywords, thereby generating new data.

\begin{table*}[htbp]
\footnotesize
    \centering
    \setlength{\tabcolsep}{4pt}
    \begin{tabular}{lcccccc}
        \toprule
        \textbf{Task Name} & \textbf{\#Q} & \textbf{IAR} & \textbf{WAR} & \textbf{SAR} & \textbf{QAR} & \textbf{CDR} \\
        \midrule
        \multicolumn{7}{l}{\textbf{Emotion}} \\
        \midrule
        Situation-based Emotion Knowledge Tasks [Typical] & 118 & 99.07\% & 100.00\% & 99.07\% & 1.85\% & 0.00\% \\
        Situation-based Emotion Knowledge Tasks [Atypical] & 119 & 95.41\% & 100.00\% & 95.41\% & 0.92\% & 0.00\% \\
        Affective Perspective-taking Test & 230 & 79.72\% & 95.28\% & 78.77\% & 6.13\% & 8.49\% \\
        Strange Stories [Contrary Emotions] & 107 & 95.88\% & 100.00\% & 95.88\% & 2.06\% & 2.06\% \\
        Tests of Emotion Comprehension [Hidden Emotions] & 228 & 85.98\% & 97.20\% & 84.58\% & 3.27\% & 8.41\% \\
        Tests of Emotion Comprehension [Moral Emotions] & 105 & 100.00\% & 100.00\% & 100.00\% & 0.00\% & 0.00\% \\
        Tests of Emotion Comprehension [Emotion Regulation] & 110 & 99.01\% & 100.00\% & 99.01\% & 0.00\% & 0.99\% \\
        Tests of Emotion Comprehension [Memory-based Emotions] & 85 & 98.40\% & 100.00\% & 98.40\% & 0.00\% & 0.00\% \\
        Tests of Emotion Comprehension [Belief-based Emotions] & 204 & 90.00\% & 91.58\% & 82.63\% & 5.79\% & 12.11\% \\
        \midrule
        \multicolumn{7}{l}{\textbf{Desire}} \\
        \midrule
        Yummy-yucky Task & 112 & 100.00\% & 100.00\% & 100.00\% & 0.00\% & 0.00\% \\
        Persuasion Story Task & 116 & 100.00\% & 83.02\% & 83.02\% & 0.00\% & 16.98\% \\
        Multiple Desires Task & 115 & 100.00\% & 100.00\% & 100.00\% & 0.00\% & 0.00\% \\
        Expanding Tasks [Twin Disagreement] & 490 & 98.64\% & 99.09\% & 97.73\% & 0.00\% & 0.00\% \\
        Strange Stories [Appearance/Reality] & 111 & 93.27\% & 100.00\% & 93.27\% & 2.88\% & 2.88\% \\
        \midrule
        \multicolumn{7}{l}{\textbf{Intention}} \\
        \midrule
        Behavioral Re-enactment Procedure & 105 & 95.79\% & 100.00\% & 95.79\% & 4.21\% & 5.26\% \\
        Expanding Tasks [Accomplice] & 182 & 98.82\% & 92.35\% & 91.18\% & 0.00\% & 4.12\% \\
        Expanding Tasks [Harmful Intention] & 186 & 92.94\% & 100.00\% & 92.94\% & 0.00\% & 1.18\% \\
        Expanding Tasks [Action Prediction] & 107 & 76.58\% & 96.40\% & 74.77\% & 1.80\% & 14.41\% \\
        Hinting Task & 99 & 83.33\% & 89.58\% & 79.17\% & 8.33\% & 11.46\% \\
        Strange Stories [Persuade] & 110 & 100.00\% & 99.01\% & 99.01\% & 0.00\% & 0.99\% \\
        Strange Stories [Figure of Speech] & 114 & 100.00\% & 97.12\% & 97.12\% & 0.00\% & 0.00\% \\
        \midrule
        \multicolumn{7}{l}{\textbf{Percept}} \\
        \midrule
        Picture Identification Task & 200 & 93.63\% & 93.63\% & 89.71\% & 8.33\% & 9.31\% \\
        Spatial Construction Task & 300 & - & - & - & - & - \\
        \midrule
        \multicolumn{7}{l}{\textbf{Knowledge}} \\
        \midrule
        Sarah Task & 112 & 97.12\% & 99.04\% & 96.15\% & 0.96\% & 1.92\% \\
        Expanding Tasks [Synesthetic Fallacy] & 192 & 70.56\% & 90.65\% & 64.49\% & 14.49\% & 27.57\% \\
        Awareness of a Reader’s Knowledge Task & 216 & 86.14\% & 99.50\% & 85.64\% & 0.50\% & 4.95\% \\
        Scalar Implicature Task & 214 & - & - & - & - & - \\
        Familiary-focus of Attention Task & 112 & 100.00\% & 100.00\% & 100.00\% & 0.88\% & 0.88\% \\
        \midrule
        \multicolumn{7}{l}{\textbf{Belief}} \\
        \midrule
        False Belief Tasks [Content] & 412 & 96.51\% & 100.00\% & 96.51\% & 2.69\% & 0.54\% \\
        False Belief Tasks [Location] & 364 & 96.25\% & 98.34\% & 94.59\% & 11.16\% & 7.21\% \\
        Naturalistic Stories [2nd-Order False Belief] & 279 & 71.67\% & 97.50\% & 70.00\% & 15.83\% & 10.83\% \\
        Naturalistic Stories [Misattribution] & 95 & 73.33\% & 95.24\% & 71.43\% & 24.76\% & 23.81\% \\
        Strange Stories [Double Bluff] & 206 & 100.00\% & 100.00\% & 100.00\% & 0.00\% & 0.00\% \\
        Unexpected Outcome Task & 392 & 99.43\% & 100.00\% & 99.43\% & 0.00\% & 0.00\% \\
        \midrule
        \multicolumn{7}{l}{\textbf{Non-literal}} \\
        \midrule
        Strange Stories [Sarcasm] & 105 & 98.96\% & 100.00\% & 98.96\% & 2.08\% & 1.04\% \\
        Strange Stories [Lie] & 111 & 96.19\% & 100.00\% & 96.19\% & 7.62\% & 5.71\% \\
        Strange Stories [White Lie] & 111 & 96.08\% & 100.00\% & 96.08\% & 0.98\% & 0.98\% \\
        Strange Stories [Forget] & 104 & 97.92\% & 98.96\% & 97.92\% & 3.13\% & 2.08\% \\
        Strange Stories [Joke] & 112 & 98.04\% & 100.00\% & 98.04\% & 0.00\% & 0.00\% \\
        Humor Task & 108 & 82.50\% & 95.00\% & 82.50\% & 20.00\% & 20.00\% \\
        Faux-pas Recognition Test & 127 & 92.62\% & 98.36\% & 90.98\% & 0.82\% & 4.10\% \\
        \midrule
        \multicolumn{7}{l}{\textbf{Comprehensive}} \\
        \midrule
        Strange Stories [Pretend] & 390 & 92.59\% & 100.00\% & 92.59\% & 13.33\% & 13.58\% \\
        Ambiguous Story Task & 168 & 83.33\% & 96.15\% & 80.13\% & 3.21\% & 8.33\% \\
        Expanding Tasks [Flattery] & 279 & 66.30\% & 100.00\% & 66.30\% & 9.42\% & 10.14\% \\
        Expanding Tasks [Jealousy] & 300 & 100.00\% & 100.00\% & 100.00\% & 0.00\% & 0.00\% \\
        See-know Task & 351 & 99.07\% & 99.69\% & 98.75\% & 0.00\% & 0.31\% \\
        \bottomrule
    \end{tabular}%
    \caption{Statistics of Human Annotation. \#Q: Number of Questions. IAR: Inter-Annotator Agreement Rate. WAR: Weak Human-Model Agreement Rate. SAR: Strong Human-Model Agreement Rate. QAR: Quality Annotation Rate. CDR: Comprehensive Defect Rate.}
    \label{tab:agrate}
\end{table*} 

\subsection{Detials For Human Annotation}
\label{app:human_anno}
In Stage 5 of dataset construction, we trained annotators Expert 3 and Expert 4 to label the quality and answers of the data. Quality was annotated as either ``quality issues present'' or ``no quality issues'', based on the example questions and task requirement descriptions for each task. Subsequently, in Stage 6, Expert 5 reviewed the annotation results and performed arbitration. Throughout this process, we calculated statistics for all tasks regarding inter-annotator agreement, weak human-model agreement, strong human-model agreement, quality annotation rate, and comprehensive defect rate, as is shown in Table \ref{tab:agrate}.

\begin{itemize}[leftmargin=*, nosep]

    \item \textbf{Inter-Annotator Agreement} refers to questions where both annotators provided the same answer (regardless of whether their quality annotations differed).
    
    \item \textbf{Weak Human-LLM Agreement} Refers to questions where the answer provided by at least one annotator matched the answer automatically generated by GPT-5.1. 
    
    \item \textbf{Strong Human-LLM Agreement} Refers to questions where the answers provided by both annotators matched the answer automatically generated by GPT-5.1. 
    
    \item \textbf{Quality Annotation} Refers to questions where at least one annotator labeled the item as having ``quality issues present''.  
    
    \item \textbf{Comprehensive Defect} Refers to questions that Expert 5, following arbitration, determined to have actual defects in either the question quality or the answer automatically generated by GPT-5.1.
    
\end{itemize}

\subsection{Prompts for ToM Evaluation}
\label{app:eval_prompt}
To ensure transparency and facilitate experimental reproducibility, the specific prompts utilized in our ToM evaluation are detailed in Table \ref{tab:zhprompt} and \ref{tab:enprompt}.

\begin{CJK*}{UTF8}{gbsn}
\begin{table*}[t!] 
\footnotesize
    \centering      
    \begin{tabular}{|p{0.95\linewidth}|}
        \hline
        你是一个乐于助人的AI助手。请仔细阅读下面的情境，然后回答问题。\\

        \\

          【情境】\\
          \{scene\}\\

          \\
        
          【问题】\\
          \{question\}\\

          \\
        
          【选项】\\
          \{options\_str\}\\

          \\
        
          【要求】\\
          1. 请根据情境内容，选出最正确的选项。\\
          2. 不需要输出思考过程或解释原因。\\
          3. 请严格按照以下格式直接输出答案：[[选项字母]]。\\

          \\
          
          示例：[[A]]\\

        \hline
    \end{tabular}
    \caption{Chinese prompt for evaluation.}
    \label{tab:zhprompt}
\end{table*} 
\end{CJK*}

\begin{table*}[t!] 
\footnotesize
    \centering      
    \begin{tabular}{|p{0.95\linewidth}|}
        \hline
        You are a helpful assistant. Please read the following scenario carefully and answer the question.\\

        \\

           [Scenario]\\
          \{scene\}\\

          \\
        
           [Question]\\
          \{question\}\\

          \\
        
          [Options]\\
          \{options\_str\}\\

          \\
        
          [Requirements]\\
          1. Select the most correct option based on the scenario.\\
          2. Do not provide any explanation or reasoning.\\
          3. Output the answer strictly in the following format: [[Option Letter]].\\

          \\
          
        Example: [[A]]\\

        \hline
    \end{tabular}
    \caption{English prompt for evaluation.}
    \label{tab:enprompt}
\end{table*} 

\subsection{Definition of New Tasks}
\label{app:def_new_tasks}
Here, we provide a clear list of the ``new tasks'' mentioned in the Section~\ref{sec:result_new_tasks}. These new tasks include:
\textit{Yummy-yucky Task},
\textit{Expanding Tasks [Twin Disagreement]},
\textit{Behavioral Re-enactment Procedure},
\textit{Expanding Tasks [Harmful Intention]},
\textit{Picture Identification Task},
\textit{Spatial Construction Task},
\textit{Expanding Tasks [Synesthetic Fallacy]},
\textit{Awareness of a Reader’s Knowledge Task},
\textit{Scalar Implicature Task},
\textit{Naturalistic Stories [2nd-Order False Belief]},
\textit{Naturalistic Stories [Misattribution]},
\textit{Strange Stories [Double Bluff]},
\textit{Unexpected Outcome Task},
\textit{Humor Task},
\textit{Strange Stories [Pretend]},
\textit{Expanding Tasks [Flattery]},
\textit{Expanding Tasks [Jealousy]}, and
\textit{See-know Task}.

\section{Additional Results}

\subsection{Differences in Bilingual Test}
\label{app:zh_en_diff}

\begin{figure}[htbp]
    \centering
    \includegraphics[width=1.0\linewidth]{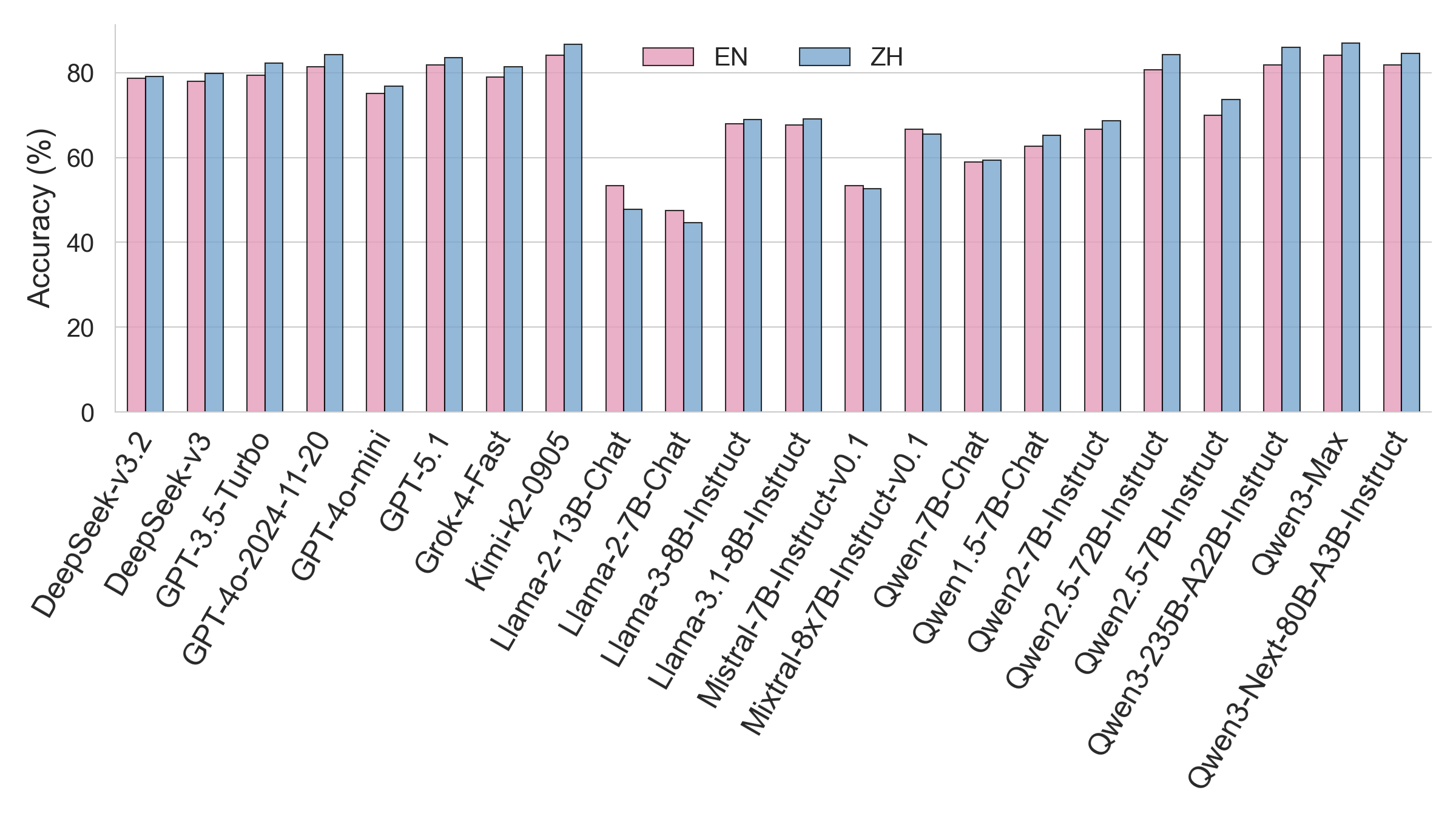}
    \caption{Differences of models in bilingual test}
    \label{fig:lang_diff}
\end{figure}

We further investigated the impact of language on model performance through a cross-lingual evaluation. As illustrated in Figure~\ref{fig:lang_diff}, the majority of evaluated models demonstrate nearly identical average performance across the bilingual test sets, indicating robust cross-lingual consistency. Notably, for most contemporary models, the results in Chinese exhibit a marginal accuracy advantage. Conversely, for earlier architectures such as the Llama-2 and Mistral series, performance in English remains slightly superior. 

\subsection{Full Restults of Models' Performance Trajectories across 46 Tasks}
\label{app:full_results_46task}

\begin{figure*}[htbp]
    \centering
    \includegraphics[width=0.9\linewidth]{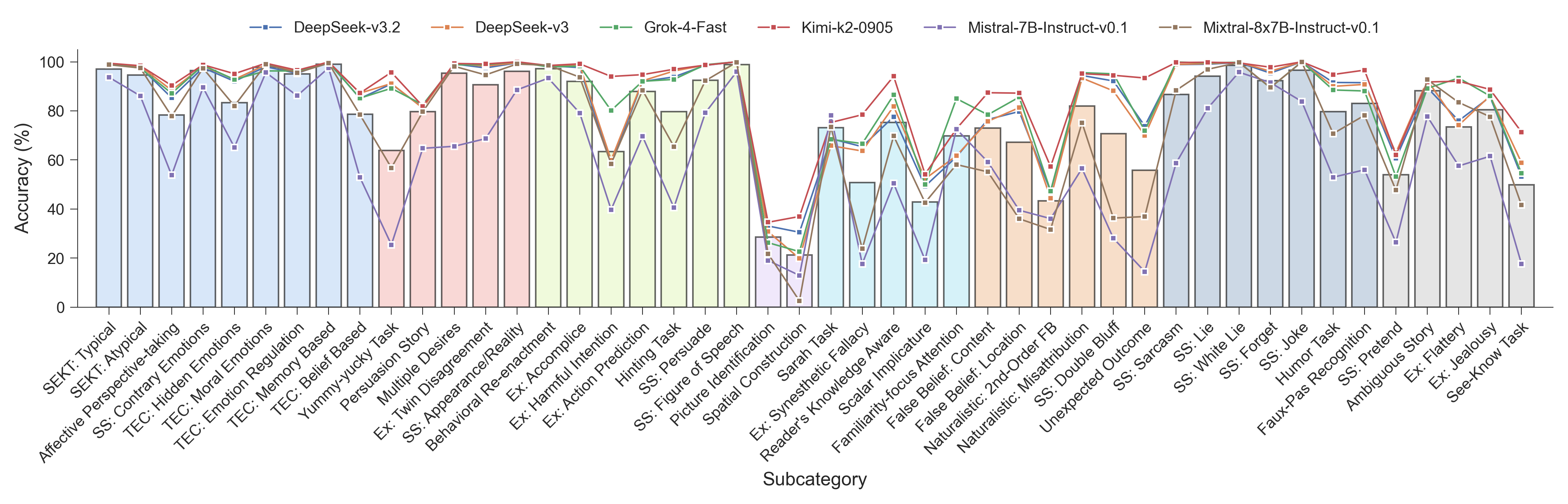}
    \includegraphics[width=0.9\linewidth]{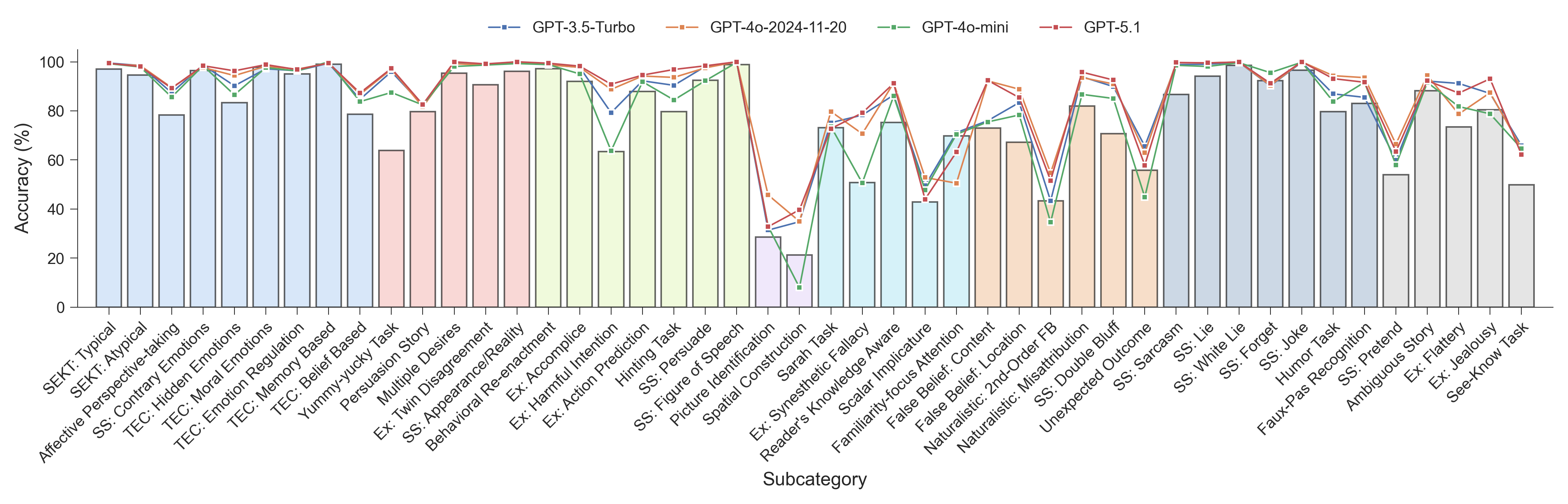}
    \includegraphics[width=0.9\linewidth]{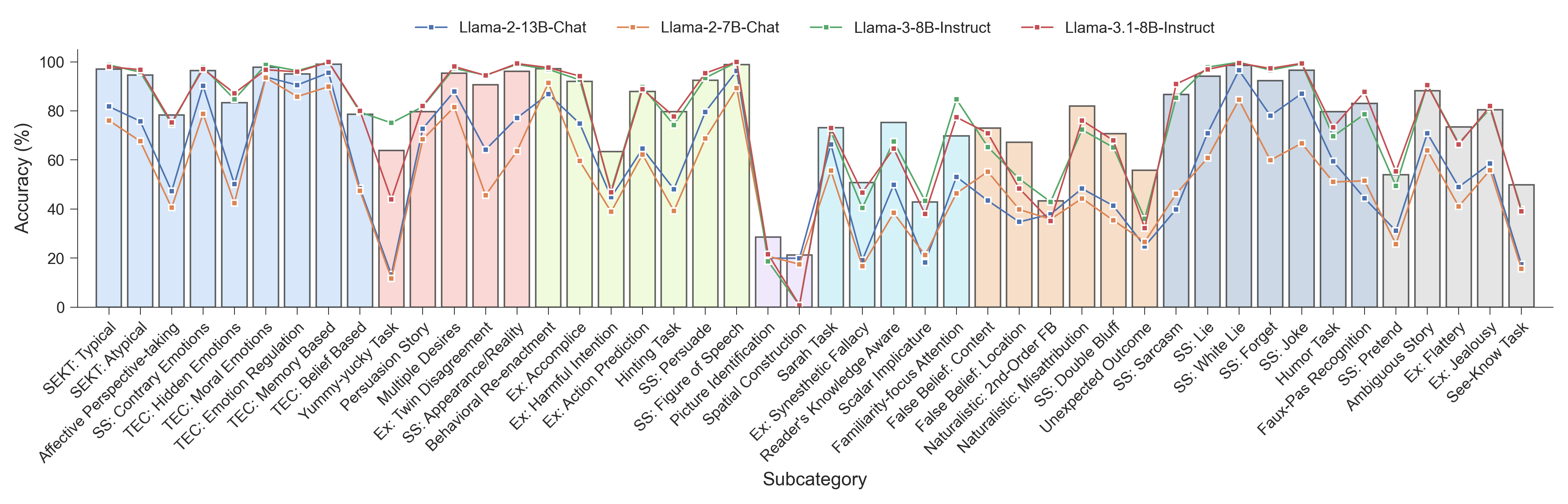}
    \includegraphics[width=0.9\linewidth]{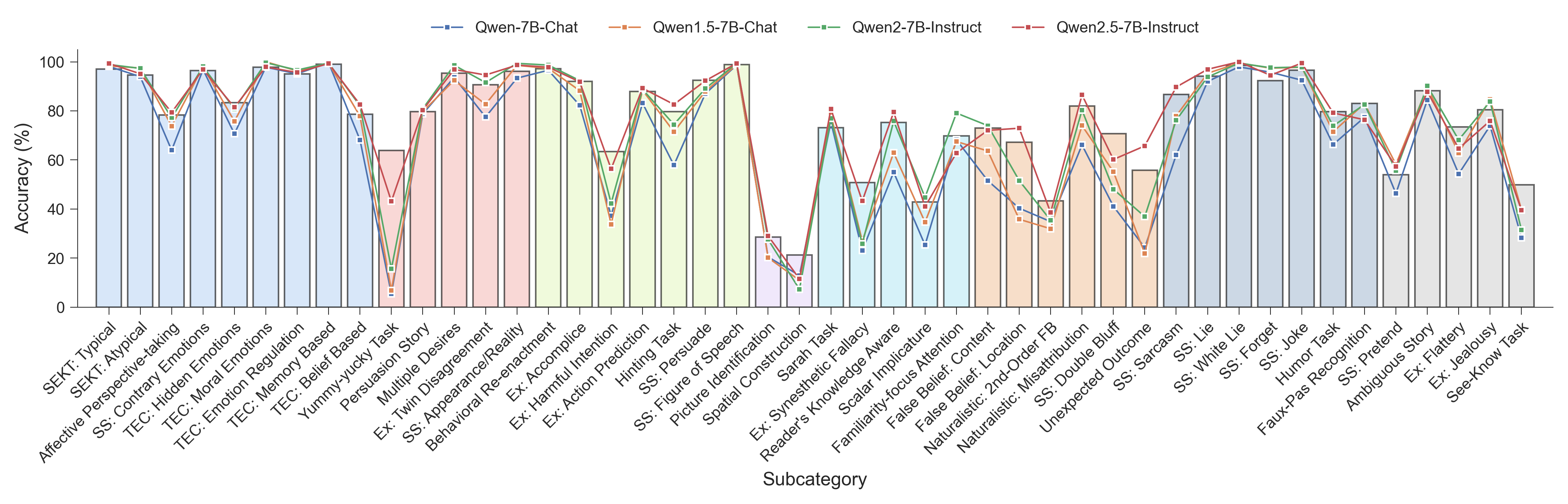}
    \includegraphics[width=0.9\linewidth]{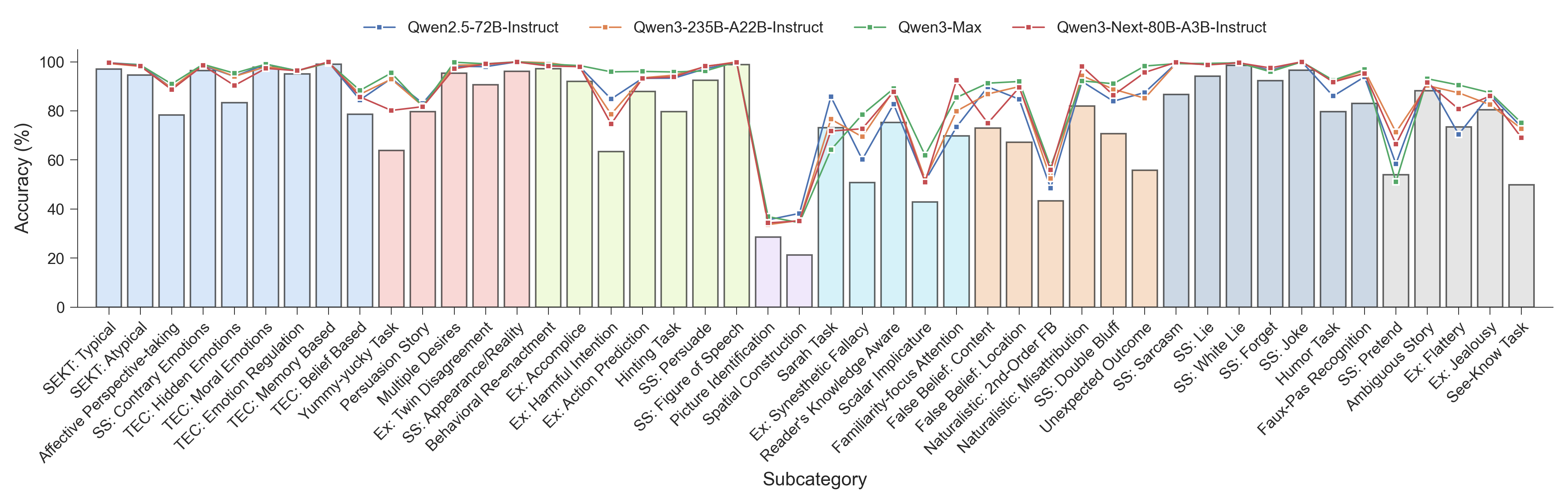}
    \caption{All evaluated LLMs' performance across 46 tasks.}
    \label{fig:full_bar_line_1}
\end{figure*}

While Section~\ref{sec:result_46_tasks} presents the performance trajectories for four representative models, we provide the complete results for all evaluated models here for a comprehensive overview, as shown in Figure~\ref{fig:full_bar_line_1}.

\section{Dataset Details}

\subsection{Details of Theory-of-Mind Abilities}
\label{app:36subability}

We adopt the ATOMS framework from psychology to identify 36 social cognitive abilities spanning seven ability dimensions in our dataset, which are used for Theory of Mind (ToM) evaluation. We compared our work with 11 datasets, and the comparison of capability assessment coverage is shown in Table \ref{tab:comparison2}. We categorized the dataset based on 36 ToM capabilities and compiled detailed statistics, as shown in Table \ref{tab:datastat}.

\begin{table*}
    \centering
    \small
    \setlength{\tabcolsep}{4.5pt}  
    \begin{tabular}{c*{18}{c}}
        \hline
        \multirow{2}{*}{\textbf{Ability}} & \multicolumn{9}{c}{\textbf{Emotion}} & \multicolumn{3}{c}{\textbf{Percept}} & \multicolumn{6}{c}{\textbf{Belief}} \\
        & E1 & E2 & E3 & E4 & E5 & E6 & E7 & E8 & E9 & P1 & P2 & P3 & B1 & B2 & B3* & B4 & B5 & B6 \\
        \hline
        ToMi    & & & & & & & & & & & & & $\checkmark$ & $\checkmark$ & $\checkmark$ & & & \\
        FANTOM    & & & & & & & & & & & & & $\checkmark$ & $\checkmark$ & $\checkmark$ & & & \\
        HI-TOM    & & & & & & & & & & & & & $\checkmark$ & $\checkmark$ & $\checkmark$* & & $\checkmark$ & \\
        MindGames    & & & & & & & & & & $\checkmark$ & & & $\checkmark$ & $\checkmark$ & $\checkmark$* & & $\checkmark$ & \\
        BigToM    & & & & & & & & & & & & $\checkmark$ & $\checkmark$ & $\checkmark$ & & & $\checkmark$ & \\
        SimpleToM    & & & & & & & & & & & & $\checkmark$ & $\checkmark$ & $\checkmark$ & & & $\checkmark$ & \\    
        OpenToM    & & & & & $\checkmark$ & $\checkmark$ & & & & & & $\checkmark$ & $\checkmark$ & $\checkmark$ & $\checkmark$ & & & \\
        NegotiationToM    & & & & & & & & & & & & & & & $\checkmark$* & & $\checkmark$ & \\
        EmoBench    & $\checkmark$ & $\checkmark$ & $\checkmark$ & $\checkmark$ & $\checkmark$ & $\checkmark$ & $\checkmark$ & $\checkmark$ & $\checkmark$ & & & & & & & & & \\
        EPITOME    & & & & $\checkmark$ & & & & & & & & & $\checkmark$ & $\checkmark$ & & & $\checkmark$ & $\checkmark$ \\
        ToMBench  & $\checkmark$ & $\checkmark$ & $\checkmark$ & $\checkmark$ & $\checkmark$ & $\checkmark$ & $\checkmark$ & & & & & & $\checkmark$ & $\checkmark$ & $\checkmark$ & $\checkmark$ & $\checkmark$ & $\checkmark$ \\    
        Ours      & $\checkmark$ & $\checkmark$ & $\checkmark$ & $\checkmark$ & $\checkmark$ & $\checkmark$ & $\checkmark$ & $\checkmark$ & $\checkmark$ & $\checkmark$ & $\checkmark$ & $\checkmark$ & $\checkmark$ & $\checkmark$ & $\checkmark$* & $\checkmark$ & $\checkmark$ & $\checkmark$ \\
        \hline
        \multirow{2}{*}{\textbf{Ability}} & \multicolumn{4}{c}{\textbf{Desire}} & \multicolumn{4}{c}{\textbf{Intention}} & \multicolumn{4}{c}{\textbf{Knowledge}} & \multicolumn{6}{c}{\textbf{Non-literal}} \\
        & D1 & D2 & D3 & D4 & I1 & I2 & I3 & I4 & K1 & K2 & K3 & K4 & N1 & N2 & N3 & N4 & N5 & N6 \\
        \hline
        ToMi    & & & & & & & & & & & & & & & & & & \\
        FANTOM    & & & & & & & & & & & $\checkmark$ & & & & & & & \\
        HI-TOM    & & & & & & & & & & & & & & $\checkmark$ & & & & \\
        MindGames    & & & & & & & & & & $\checkmark$ & & & & & & & & \\
        BigToM    & & & $\checkmark$ & & & & & & & $\checkmark$ & & & & & & & & \\
        SimpleToM    & & & & & & & & & & $\checkmark$ & $\checkmark$ & & & & & & & \\
        OpenToM    & & & & & & & & & & $\checkmark$ & & & & & & & & \\
        NegotiationToM    & $\checkmark$ & $\checkmark$ & $\checkmark$ & & & $\checkmark$ & $\checkmark$ & $\checkmark$ & & & & & & & & & & \\
        EmoBench    & & & & & & & & & & & & & & & & & & \\
        EPITOME    & & & & $\checkmark$ & & & & $\checkmark$ & & & $\checkmark$ & & $\checkmark$ & $\checkmark$ & $\checkmark$ & $\checkmark$ & $\checkmark$ & \\
        ToMBench  & $\checkmark$ & $\checkmark$ & $\checkmark$ & $\checkmark$ & $\checkmark$ & $\checkmark$ & $\checkmark$ & $\checkmark$ & $\checkmark$ & $\checkmark$ & $\checkmark$ & $\checkmark$ & $\checkmark$ & $\checkmark$ & $\checkmark$ & $\checkmark$ & $\checkmark$ & $\checkmark$ \\
        Ours      & $\checkmark$ & $\checkmark$ & $\checkmark$ & $\checkmark$ & $\checkmark$ & $\checkmark$ & $\checkmark$ & $\checkmark$ & $\checkmark$ & $\checkmark$ & $\checkmark$ & $\checkmark$ & $\checkmark$ & $\checkmark$ & $\checkmark$ & $\checkmark$ & $\checkmark$ & $\checkmark$ \\
        \hline
    \end{tabular}
    \caption{Comparison of ToM ability coverage across benchmarks. The checkmarks indicate covered abilities. Specifically in the B3* column, $\checkmark$ denotes the evaluation of second-order beliefs only, while $\checkmark$* indicates the evaluation of third-order or higher-order beliefs.}
    \label{tab:comparison2}
\end{table*} 

\bigskip

\noindent {\bf Emotion} entails understanding how contextual factors influence emotional states, recognizing the potential for experiencing complex emotions, and acknowledging the ability to regulate emotional expressions. This dimension comprises 9 capabilities. Notably, the ATOMS framework originally includes a ``Comprehensive measure involving emotion'' that is difficult to quantify in isolation; therefore, we decomposed this measure into capabilities (E8) and (E9) to capture aspects distinct from (E1) through (E7).

\noindent (E1) Typical emotional reactions~\cite{Ab-E1}: Inferring a person’s emotional reactions based on situations that typically elicit certain emotions/inferring a preceding event based on a person’s emotional reaction.

\noindent (E2) Atypical emotional reactions~\cite{Task-APT}: Inferring or explaining a person’s emotional reactions based on situations eliciting emotions that are atypical compared to what is usually expected.

\noindent (E3) Discrepant emotions~\cite{Ab-E3}: Understanding that people may have discrepant feelings about an event.

\noindent (E4) Mixed emotions~\cite{Ab-E4}: Understanding that people may feel mixed emotions or different emotions successively.

\noindent (E5) Hidden emotions~\cite{Ab-E5}: Understanding that other people may hide their emotions.

\noindent (E6) Moral emotions~\cite{Task-TEC}: Understanding that negative feelings might arise following a reprehensible action.

\noindent (E7) Emotion regulation~\cite{Task-TEC}: Understanding that others might use strategies to regulate their emotions.

\noindent (E8) Emotion based on memory~\cite{Task-TEC}: Understanding that others might be emotionally triggered by memories.

\noindent (E9) Emotion based on belief~\cite{Task-TEC}: Understanding that others' emotions are influenced by their beliefs.

\bigskip

\noindent {\bf Desire} entails understanding that individuals hold subjective desires, preferences, and needs, and recognizing how these factors influence their emotions and actions. This dimension comprises 4 capabilities.

\noindent (D1) Discrepant desires~\cite{Task-YT}: Understanding that different people may have discrepant desires.

\noindent (D2) Multiple desires~\cite{Task-MDT}: Understanding the co-existence of multiple desires simultaneously or successively in one person.

\noindent (D3) Desires influence on emotions and actions~\cite{Ab-D3a,Ab-D3b}: Understanding that people’s emotions and actions are influenced by their desires/preferences.

\noindent (D4) Desire-action contradiction~\cite{Ab-D4}: Producing plausible explanations when actions contradict stated desires/preferences.

\bigskip

\noindent {\bf Intention} entails understanding the ability of individuals to take actions to achieve goals and intentions. This dimension comprises 4 capabilities.

\noindent (I1) Completion of failed actions~\cite{Task-BRP}: Understanding another person’s intent, as demonstrated by completing their failed action.

\noindent (I2) Discrepant intentions~\cite{Ab-I2}: Understanding that identical actions/results can be achieved with different intentions.

\noindent (I3) Prediction of actions~\cite{Ab-I3}: Predicting people’s actions based on their intentions.

\noindent (I4) Intentions explanations~\cite{Ab-I4}: Producing plausible intention explanations for different types of observed social events.

\bigskip

\noindent {\bf Percept} entails understanding the subjectivity of perceptual experiences and distinguishing between the perceptual information available to oneself and others. This dimension comprises 3 capabilities.

\noindent (P1) Simple visual perspective taking~\cite{Task-PIT}: Acknowledging that others have different visual percepts and adopting the visual perspective of another person.

\noindent (P2) Complex visual perspective taking~\cite{Task-SCT}: Adopting another person’s visual perspective in tasks demanding complex mental rotation or visualization.

\noindent (P3) Percept-action link~\cite{Ab-P3}: Understanding that other’s actions are linked to their visual percepts.

\bigskip

\noindent {\bf Knowledge} entails understanding that individuals possess distinct knowledge based on their percepts, received information, or familiarity with objects. This dimension comprises 4 capabilities.

\noindent (K1) Knowledge-pretend play links~\cite{Task-ST}: Understanding that someone who does not know something exists cannot engage in ``pretend play'' that incorporates that knowledge.

\noindent (K2) Percepts-knowledge links~\cite{Task-SkT}: Understanding that someone who does not have access to perceptual information (i.e., by looking, hearing, etc.) may not have access to knowledge.

\noindent (K3) Information-knowledge links~\cite{Task-ARKT}: Understanding that someone who was not informed or is not familiar with something may not know.

\noindent (K4) Knowledge-attention links~\cite{Task-FAT}: Understanding that something new is more interesting to someone than something already known.

\bigskip

\noindent {\bf Belief} entails understanding that individuals may hold beliefs about the world that diverge from reality or differ from one's own. This dimension comprises 6 capabilities. Notably, regarding the second-order belief capability in (B3), we extend it to higher-order belief capabilities, denoted as (B3*).

\noindent (B1) Content false beliefs~\cite{Task-FBC}: Familiar container with an unexpected content: Understanding the false belief held by someone who never opened the container.

\noindent (B2) Location false beliefs~\cite{Task-FBL}: Unseen change: Understanding the false belief held by someone who did not witness or was not informed of a displacement or change of action.

\noindent (B3*) Second-order (High-order*) belief~\cite{Task-FB2nd}: Understanding the second-order (high-order*) belief or false belief held by someone who does not know somebody else was informed.

\noindent (B4) Identity false beliefs~\cite{Ab-B4}: Understanding that when something looks/sounds/smells like something else, a person may hold a false belief about its identity.

\noindent (B5) Beliefs based action~\cite{Ab-B5}: Predicting another person's actions based on their stated beliefs or inferring another person’s belief based on their stated action.

\noindent (B6) Sequence false beliefs~\cite{Task-UOT}: Understanding the false belief created when a predictable sequence of stimuli is broken with the intrusion of an unexpected stimulus.

\bigskip

\noindent {\bf Non-literal} entails understanding that communication can convey information beyond literal meaning. This dimension comprises 6 capabilities.

\noindent (N1) Irony/sarcasm~\cite{Task-SS}: Understanding that other people may lie in order to be ironic/sarcastic.

\noindent (N2) Egocentric lies~\cite{Task-SS}: Understanding that someone may consciously lie in order to avoid a problem or to get their way.

\noindent (N3) White lies~\cite{Task-SS}: Understanding that someone may lie in order to spare another’s feelings.

\noindent (N4) Involuntary lies~\cite{Task-SS}: Understanding that someone may tell a ``lie'' without knowing.

\noindent (N5) Humor~\cite{Task-SS}: Understanding that someone may tell a ``lie'' in order to make a joke.

\noindent (N6) Faux pas~\cite{Task-FRT}: Ability to recognize faux pas (social gaffe) situations.

\subsection{Details of Theory-of-Mind Tasks}
\label{app:46subability}

To objectively and comprehensively assess the aforementioned 36 capabilities, we selected social cognitive tasks from psychology that are suitable for evaluating ToM. This process resulted in 46 text-based ToM tasks adapted for LLM evaluation. Each task involves single or multiple capabilities; therefore, based on the specific capabilities addressed, we categorized the tasks into eight groups. Seven of these categories correspond to the seven capability dimensions, while the final category, Comprehensive tasks, involves the assessment of capabilities across different dimensions.

\bigskip

\noindent {\bf Emotion} comprises 9 tasks dedicated solely to assessing Emotion capabilities.

\noindent {\bf (e1) Situation-based Emotion Knowledge Tasks [Typical]}~\cite{Task-SEKT}: Presented with a story context where a protagonist exhibits a typical emotional reaction, participants are asked to analyze the cause of this emotion. This task involves the assessment of capability (E1). 1 group of example data is shown as Table \ref{tab:e1_example}. 

\noindent {\bf (e2) Situation-based Emotion Knowledge Tasks [Atypical]}~\cite{Task-SEKT}: Presented with a story context where a protagonist exhibits an atypical emotional reaction, participants are asked to analyze the cause of this emotion. This task involves the assessment of capability (E2). 1 group of example data is shown as Table \ref{tab:e2_example}. 

\noindent {\bf (e3) Affective Perspective-taking Test}~\cite{Task-APT}: Faced with a story where two protagonists are affected differently by the same event, participants are asked to accurately predict that the two protagonists will have different emotional reactions. This task involves the assessment of capability (E3). 1 group of example data is shown as Table \ref{tab:e3_example}.

\noindent {\bf (e4) Strange Stories [Contrary Emotions]}~\cite{Task-SS}: Faced with a story where a protagonist experiences two contradictory emotions, participants are asked to analyze the causes of this complex emotion. This task involves the assessment of capability (E4). 1 group of example data is shown as Table \ref{tab:e4_example}.

\noindent {\bf (e5) Tests of Emotion Comprehension [Hidden Emotions]}~\cite{Task-TEC}: Given a context where the protagonist hides their emotion, participants are asked to identify the hidden emotion and explain the reason for hiding it. This task involves the assessment of capability (E5). 1 group of example data is shown as Table \ref{tab:e5_example}.

\noindent {\bf (e6) Tests of Emotion Comprehension [Moral Emotions]}~\cite{Task-TEC}: Given a context where the protagonist commits an immoral act, participants are asked to analyze the emotion generated by the protagonist's internal moral values. This task involves the assessment of capability (E6). We rigorously revised the correct answers for each question to ensure that the testing criteria are grounded in a set of generally accepted moral standards, thereby avoiding unnecessary ethical risks. 1 group of example data is shown as Table \ref{tab:e6_example}.

\noindent {\bf (e7) Tests of Emotion Comprehension [Emotion Regulation]}~\cite{Task-TEC}: Given a context where the protagonist faces an event likely to cause negative emotions, participants are asked to predict the protagonist's method of emotional regulation. This task involves the assessment of capability (E7). 1 group of example data is shown as Table \ref{tab:e7_example}.

\noindent {\bf (e8) Tests of Emotion Comprehension [Memory-based Emotions]}~\cite{Task-TEC}: Given a context along with the protagonist's past experience, participants are asked to predict the protagonist's emotional change based on the memories they might recall. This task involves the assessment of capability (E8). 1 group of example data is shown as Table \ref{tab:e8_example}.

\noindent {\bf (e9) Tests of Emotion Comprehension [Belief-based Emotions]}~\cite{Task-TEC}: Given a context where two protagonists hold different beliefs about the same event, participants are asked to accurately predict their differing emotional reactions. This task involves the assessment of capability (E9). 1 group of example data is shown as Table \ref{tab:e9_example}.

\bigskip

\noindent {\bf Desire} comprises 5 tasks dedicated solely to assessing Desire capabilities.

\noindent {\bf (d1) Yummy-yucky Task}~\cite{Task-YT}: Presented with a two-person interaction context, participants are asked to accurately identify the item desired by the protagonist when their own preferences differ from those of the protagonist. This task involves the assessment of capability (D1). We selected scenario designs that are more closely aligned with the original psychological paradigms. 1 group of example data is shown as Table \ref{tab:d1_example}.

\noindent {\bf (d2) Persuasion Story Task}~\cite{Task-PST}: Presented with a negotiation context where they face an individual with desires different from their own, participants are required to demonstrate the ability to understand and select effective persuasion strategies. This task involves the assessment of capability (D1). 1 group of example data is shown as Table \ref{tab:d2_example}.

\noindent {\bf (d3) Multiple Desires Task}~\cite{Task-MDT}: Given a context where the protagonist's original plan is interrupted, participants are asked to understand the protagonist's ability to maintain their original desire. This task involves the assessment of capability (D2). 1 group of example data is shown as Table \ref{tab:d3_example}.

\noindent {\bf (d4) Expanding Tasks [Twin Disagreement]}: Given a context where two protagonists hold different desires, participants are asked to predict their actions or emotions based on these desires across different questions. This task involves the assessment of capability (D3). Building upon our understanding of~\cite{Ab-D3a,Ab-D3b}, we redesigned this series of tasks to ensure comprehensive coverage of the assessment of this capability. 1 group of example data is shown as Table \ref{tab:d4_example}.

\noindent {\bf (d5) Strange Stories [Appearance/Reality]} \cite{Task-SS}: Faced with a story where the protagonist denies their own desire, participants are asked to analyze the reasons behind the protagonist's refusal to acknowledge the desire. This task involves the assessment of capability (D4). 1 group of example data is shown as Table \ref{tab:d5_example}.

\bigskip

\noindent {\bf Intention} comprises 7 tasks dedicated solely to assessing Intention capabilities.

\noindent {\bf (i1) Behavioral Re-enactment Procedure}~\cite{Task-BRP}: Presented with a two-person interaction context involving a failed action, participants are asked to identify the original intention behind the action and successfully achieve this intention during re-enactment. This task involves the assessment of capability (I1). 1 group of example data is shown as Table \ref{tab:i1_example}.

\noindent {\bf (i2) Expanding Tasks [Accomplice]}: Faced with a story where two protagonists jointly contribute to a negative outcome, participants are asked to accurately analyze the distinct intentions of each protagonist. This task involves the assessment of capability (I2). Building upon our understanding of~\cite{Ab-I2}, we redesigned this series of tasks to ensure comprehensive coverage of the assessment of this capability. 1 group of example data is shown as Table \ref{tab:i2_example}.

\noindent {\bf (i3) Expanding Tasks [Harmful Intention]}: Given a base context, participants are asked to accurately analyze whether the harmfulness of the protagonist's intention aligns with the harmfulness of the outcome across different scenario branches. This task involves the assessment of capability (I2). The selection of our scenarios is based on~\cite{Task-ExHI}, but we have made necessary adjustments to the focus of the questions to ensure comprehensive coverage of the assessment of this capability. 1 group of example data is shown as Table \ref{tab:i3_example}.

\noindent {\bf (i4) Expanding Tasks [Action Prediction]}: Given a context, participants are asked to accurately analyze the protagonist's intention and predict their next action. This task involves the assessment of capability (I3). Building upon our understanding of~\cite{Ab-I3}, we redesigned this series of tasks to ensure comprehensive coverage of the assessment of this capability. 1 group of example data is shown as Table \ref{tab:i4_example}.

\noindent {\bf (i5) Hinting Task}~\cite{Task-HT}: Given a context, participants are asked to infer the speaker's true intention from indirect hints within a social interaction. This task involves the assessment of capability (I4). 1 group of example data is shown as Table \ref{tab:i5_example}.

\noindent {\bf (i6) Strange Stories [Persuade]}~\cite{Task-SS}: Faced with a story where the protagonist influences others' beliefs through exaggeration or feigned weakness to achieve their own goal, participants are asked to identify the intention behind the speaker's behavior. This task involves the assessment of capability (I4). 1 group of example data is shown as Table \ref{tab:i6_example}.

\noindent {\bf (i7) Strange Stories [Figure of Speech]}~\cite{Task-SS}: Faced with a story where the protagonist uses rhetoric or idioms to describe the current situation, participants are asked to identify the meaning the speaker truly intends to convey. This task involves the assessment of capability (I4). 1 group of example data is shown as Table \ref{tab:i7_example}.

\bigskip

\noindent {\bf Percept} comprises 2 tasks dedicated solely to assessing Percept capabilities.

\noindent {\bf (p1) Picture Identification Task}~\cite{Task-PIT}: Presented with a multi-faceted object, participants are asked to understand that the pattern they see differs from the pattern seen by a person in a different position. This task involves the assessment of capability (P1). 1 group of example data is shown as Table \ref{tab:p1_example}.

\noindent {\bf (p2) Spatial Construction Task}~\cite{Task-SCT}: Presented with a set of objects and the position of another participant, participants are asked to reconstruct the visual information seen by the other participant based on the visual information they see themselves. This task involves the assessment of capability (P2). 1 group of example data is shown as Table \ref{tab:p2_example}.

\bigskip

\noindent {\bf Knowledge} comprises 5 tasks dedicated solely to assessing Knowledge capabilities.

\noindent {\bf (k1) Sarah Task}~\cite{Task-ST}: Given a social context where the knowledge system differs from human social structure, participants are asked to identify the source of the protagonist's imitative actions based on the type of knowledge the protagonist holds. This task involves the assessment of capability (K1). 1 group of example data is shown as Table \ref{tab:k1_example}.

\noindent {\bf (k2) Expanding Tasks [Synesthetic Fallacy]}: Presented with a special object that requires the simultaneous use of two senses for successful recognition, participants are asked to predict the misperception of a person lacking one of the senses. This task involves the assessment of capability (K2). Building upon our understanding of~\cite{Task-SkT} and drawing inspiration from the classic parable of the blind men and the elephant, we redesigned this series of tasks to ensure comprehensive coverage of the assessment of this capability. 3 groups of example data is shown as Table \ref{tab:k2_example}. 

\noindent {\bf (k3) Awareness of a Reader’s Knowledge Task}~\cite{Task-ARKT}: Given a letter-writing context, participants are asked to describe the same concept differently based on the recipient's varying levels of knowledge. This task involves the assessment of capability (K3). 1 group of example data is shown as Table \ref{tab:k3_example}.

\noindent {\bf (k4) Scalar Implicature Task}~\cite{Task-SIT}: Given a context, participants are asked to predict the protagonist's estimate of a quantity based on their degree of access to information regarding that quantity. This task involves the assessment of capability (K3). We introduced appropriate complexity to the story scenarios to enhance the depth of the questions, while simultaneously rigorously revising the selection of numerical values for each option. 1 group of example data is shown as Table \ref{tab:k4_example}.

\noindent {\bf (k5) Familiarity-focus of Attention Task}~\cite{Task-FAT}: Given a context, participants are asked to identify that the protagonist's attention is focused on unknown objects rather than known ones. This task involves the assessment of capability (K4). 1 group of example data is shown as Table \ref{tab:k5_example}.

\bigskip

\noindent {\bf Belief} comprises 6 tasks dedicated solely to assessing Belief capabilities.

\noindent {\bf (b1) False Belief Tasks [Content]}~\cite{Task-FBC,Task-FB2nd}: Participants are asked to distinguish between their own true beliefs and others' false beliefs regarding content information, and further predict second-order beliefs. This task involves the assessment of capabilities (B1) and (B3). 1 group of example data is shown as Table \ref{tab:b1_example}.

\noindent {\bf (b2) False Belief Tasks [Location]}~\cite{Task-FBL,Task-FBLL,Task-FB2nd}: Participants are asked to distinguish between their own true beliefs and others' false beliefs regarding location information, and further predict second-order beliefs. This task involves the assessment of capabilities (B2) and (B3). 1 group of example data is shown as Table \ref{tab:b2_example}.

\noindent {\bf (b3) Naturalistic Stories [2nd-Order False Belief]}~\cite{Task-NS}: Participants are asked to distinguish between their own true second-order beliefs and others' false second-order beliefs regarding first-order belief information, and further predict third-order beliefs. This task involves the assessment of capability (B3*). 2 groups of example data is shown as Table \ref{tab:b3_example}. 

\noindent {\bf (b4) Naturalistic Stories [Misattribution]}~\cite{Task-NS}: Given a context where a misunderstanding occurs, participants are asked to understand that the misunderstanding was caused by the holding of a false belief. This task involves the assessment of capability (B5). 1 group of example data is shown as Table \ref{tab:b4_example}.

\noindent {\bf (b5) Strange Stories [Double Bluff]}~\cite{Task-SS}: Given an adversarial context, participants are asked to understand the strategic interaction regarding the narration of facts between two parties based on mutual distrust. This task involves the assessment of capability (B5). We introduced appropriate complexity to the story scenarios to enhance the depth of the questions. 1 group of example data is shown as Table \ref{tab:b5_example}.

\noindent {\bf (b6) Unexpected Outcome Task}~\cite{Task-UOT}: Building upon tasks (b1) and (b2), an unexpected outcome is introduced to form sequential false beliefs; participants are asked not to be distracted by this information. This task involves the assessment of capabilities (B3) and (B6). Our understanding of the scenario for this task differs significantly from that of ToMBench; therefore, we chose to refer to the literature cited in ATOMS regarding the assessment of ``Sequence False Belief''. 1 group of example data is shown as Table \ref{tab:b6_example}.

\bigskip

\noindent {\bf Non-literal} comprises 7 tasks dedicated solely to assessing Non-literal capabilities.

\noindent {\bf (n1) Strange Stories [Sarcasm]}~\cite{Task-SS}: Presented with a story where a protagonist is sarcastic towards other characters, participants are asked to understand this specific social context. This task involves the assessment of capability (N1). 1 group of example data is shown as Table \ref{tab:n1_example}.

\noindent {\bf (n2) Strange Stories [Lie]}~\cite{Task-SS}: Presented with a story where a protagonist lies for selfish purposes, participants are asked to understand this specific social context. This task involves the assessment of capability (N2). 1 group of example data is shown as Table \ref{tab:n2_example}.

\noindent {\bf (n3) Strange Stories [White Lie]}~\cite{Task-SS}: Presented with a story where a protagonist lies with good intentions, participants are asked to understand this specific social context. This task involves the assessment of capability (N3). 1 group of example data is shown as Table \ref{tab:n3_example}.

\noindent {\bf (n4) Strange Stories [Forget]}~\cite{Task-SS}: Presented with a story where a protagonist makes untrue statements because they have forgotten important facts, participants are asked to understand this specific social context. This task involves the assessment of capability (N4). 1 group of example data is shown as Table \ref{tab:n4_example}.

\noindent {\bf (n5) Strange Stories [Joke]}~\cite{Task-SS}: Presented with a story where a protagonist makes a joke, participants are asked to understand this specific social context. This task involves the assessment of capability (N5). 1 group of example data is shown as Table \ref{tab:n5_example}.

\noindent {\bf (n6) Humor Task}~\cite{Task-HumTT,Task-HumT}: Given the first half of a joke, participants are asked to complete the joke and explain its punchline. This task involves the assessment of capability (N5). 1 group of example data is shown as Table \ref{tab:n6_example}.

\noindent {\bf (n7) Faux-pas Recognition Test}~\cite{Task-FRT}: Given a context containing a typical faux pas, participants are asked to identify the inappropriate language involved. This task involves the assessment of capability (N6). 1 group of example data is shown as Table \ref{tab:n7_example}.

\bigskip

\noindent {\bf Comprehensive} comprises 5 tasks, where each task assesses capabilities that span more than one dimension.

\noindent {\bf (c1) Strange Stories [Pretend]}~\cite{Task-SS}: Presented with two items that are similar in appearance but essentially distinct, participants are asked to understand both pretend play and identity false beliefs simultaneously. This task involves the assessment of capabilities (I4), (K1), and (B4). We introduced appropriate complexity to the story scenarios to broaden the scope of the questions. 1 group of example data is shown as Table \ref{tab:c1_example}.

\noindent {\bf (c2) Ambiguous Story Task}~\cite{Task-AST}: Presented with ambiguous social stories, participants are asked to understand the intentions, beliefs, and emotions of others under conditions of uncertainty. This task involves the assessment of capabilities (E9) and (I4). 1 group of example data is shown as Table \ref{tab:c2_example}.

\noindent {\bf (c3) Expanding Tasks [Flattery]}: Given a story involving flattery, participants are asked to understand this specific social context and analyze the characters' intentions and beliefs. This task involves the assessment of capabilities (I4) and (B3). This task is an original design based on our understanding of ``flattery'' in complex social contexts. 1 group of example data is shown as Table \ref{tab:c3_example}.

\noindent {\bf (c4) Expanding Tasks [Jealousy]}: Given a story involving jealousy, participants are asked to understand this specific social context and analyze the characters' emotions and beliefs. This task involves the assessment of capabilities (E5), (B5), and (N1). This task is an original design based on our understanding of ``jealousy'' in complex social contexts. 1 group of example data is shown as Table \ref{tab:c4_example}.

\noindent {\bf (c5) See-know Task}~\cite{Task-SkT}: Given a context, participants are asked to understand that the characters' knowledge and behaviors are influenced by their perceptual information. This task involves the assessment of capabilities (P3) and (K2). 1 group of example data is shown as Table \ref{tab:c5_example}.

\begin{table*}[t]
\footnotesize
  \centering
  \setlength{\tabcolsep}{7.5pt}

  \begin{tabular}{lccccccc}
    \toprule
    \multirow{2}{*}{\textbf{Ability}} & \multirow{2}{*}{\textbf{Number}} & \multicolumn{2}{c}{\textbf{ASL}} & \multicolumn{2}{c}{\textbf{AQL}} & \multicolumn{2}{c}{\textbf{AOL}} \\
    \cmidrule(lr){3-4} \cmidrule(lr){5-6} \cmidrule(lr){7-8}
     & & \textbf{En.} & \textbf{Zh.} & \textbf{En.} & \textbf{Zh.} & \textbf{En.} & \textbf{Zh.} \\
    \midrule

    \multicolumn{8}{l}{\textbf{Emotion}} \\
    \midrule
    Typical emotional reactions          & 118 & 42.97 & 61.08 & 7.14 & 13.00 & 37.87 & 49.41 \\
    Atypical emotional reactions         & 119 & 69.49 & 96.37 & 7.36 & 13.00 & 57.26 & 73.92 \\
    Discrepant emotions                  & 230 & 34.04 & 45.19 & 4.60 & 10.27 & 8.58 & 16.00 \\
    Mixed emotions                       & 107 & 76.15 & 109.74 & 11.38 & 17.14 & 78.15 & 97.27 \\
    Hidden emotions                      & 328 & 65.06 & 93.57 & 12.02 & 20.15 & 27.67 & 40.97 \\
    Moral emotions                       & 105 & 107.29 & 152.48 & 24.78 & 37.15 & 75.08 & 122.31 \\
    Emotion regulation                   & 110 & 104.71 & 155.15 & 9.09 & 19.00 & 97.05 & 141.93 \\
    Emotion based on memory              & 85  & 58.93 & 82.36 & 6.34 & 12.49 & 45.21 & 59.72 \\
    Emotion based on belief              & 288 & 89.31 & 133.92 & 9.34 & 14.39 & 22.27 & 34.25 \\
    \midrule

    \multicolumn{8}{l}{\textbf{Desire}} \\
    \midrule
    Multiple desires                     & 228 & 83.38 & 118.27 & 10.25 & 13.86 & 77.82 & 111.53 \\
    Discrepant desires                   & 115 & 91.30 & 119.80 & 16.03 & 24.31 & 50.02 & 64.98 \\
    Desires influence on actions/emotions& 490 & 59.72 & 77.16 & 13.17 & 20.02 & 15.10 & 22.62 \\
    Desire-action contradiction          & 111 & 119.32 & 160.68 & 12.53 & 15.82 & 50.53 & 65.79 \\

    \midrule

    \multicolumn{8}{l}{\textbf{Intention}} \\
    \midrule
    Completion of failed actions         & 105 & 74.87 & 107.69 & 9.44 & 17.37 & 72.82 & 103.85 \\
    Discrepant intentions                & 368 & 101.39 & 141.72 & 13.32 & 24.88 & 111.01 & 157.23 \\
    Prediction of actions                & 107 & 85.16 & 123.64 & 10.02 & 15.16 & 54.33 & 73.19 \\
    Intentions explanations              & 723 & 72.21 & 106.24 & 13.02 & 20.06 & 53.80 & 74.25 \\
    \midrule

    \multicolumn{8}{l}{\textbf{Percept}} \\
    \midrule
    Simple visual perspective taking         & 200 & 73.96 & 96.25 & 9.60 & 15.73 & 9.12 & 17.11 \\
    Complex visual perspective taking        & 300 & 84.75 & 117.59 & 7.34 & 14.22 & 165.50 & 200.56 \\
    Percept-action link                      & 234 & 88.59 & 147.07 & 14.14 & 19.68 & 16.10 & 46.87 \\
    \midrule

    \multicolumn{8}{l}{\textbf{Knowledge}} \\
    \midrule
    Knowledge-pretend play links         & 242 & 68.26 & 115.98 & 13.95 & 22.60 & 20.28 & 31.23 \\
    Percepts-knowledge links             & 309 & 77.66 & 117.05 & 12.27 & 17.98 & 20.33 & 35.64 \\
    Information-knowledge links          & 430 & 42.98 & 65.24 & 20.36 & 33.45 & 65.10 & 90.88 \\
    Knowledge-attention links            & 112 & 152.71 & 236.75 & 7.11 & 10.00 & 53.23 & 67.05 \\
    \midrule

    \multicolumn{8}{l}{\textbf{Belief}} \\
    \midrule
    Content false beliefs                & 206 & 62.96 & 97.04 & 13.78 & 24.00 & 9.04 & 17.19 \\
    Location false beliefs               & 182 & 99.99 & 131.66 & 14.73 & 23.30 & 26.84 & 31.59 \\
    Second-order beliefs                 & 863 & 95.45 & 134.30 & 19.68 & 30.48 & 22.21 & 31.37 \\
    High-order beliefs*                  & 93  & 145.78 & 200.03 & 12.18 & 22.17 & 19.26 & 28.18 \\
    Identity false beliefs               & 130 & 36.78 & 57.56 & 19.74 & 27.53 & 24.61 & 32.53 \\
    Beliefs based action/emotions        & 401 & 72.60 & 104.43 & 17.43 & 26.30 & 57.77 & 74.76 \\
    Sequence false beliefs               & 196 & 87.73 & 123.24 & 12.84 & 21.53 & 11.86 & 19.43 \\
    \midrule

    \multicolumn{8}{l}{\textbf{Non-Literal Communication}} \\
    \midrule
    Irony/Sarcasm                        & 205 & 87.96 & 124.64 & 17.34 & 23.30 & 45.99 & 63.25 \\
    Egocentric lies                      & 111 & 97.10 & 135.13 & 5.18 & 9.02 & 82.29 & 112.50 \\
    White lies                           & 111 & 83.83 & 120.77 & 6.01 & 9.17 & 101.33 & 140.04 \\
    Involuntary lies                     & 104 & 86.34 & 115.38 & 5.79 & 9.02 & 74.78 & 101.50 \\
    Humor                                & 220 & 64.15 & 96.05 & 15.44 & 25.76 & 62.13 & 92.93 \\
    Faux pas                             & 127 & 139.42 & 218.61 & 6.00 & 15.00 & 54.20 & 90.83 \\
    \midrule

    \multicolumn{8}{l}{\textbf{Total}} \\
    \midrule
    SUM.                        & 8513 & - & - & - & - & - & - \\
    AVG.                        & -    & 79.51 & 114.48 & 13.22 & 21.08 & 47.47 & 66.65 \\
    \bottomrule
  \end{tabular}

  \caption{Data statistics. ASL: Average scene length (English/Chinese). AQL: Average question length (English/Chinese). AOL: Average option length (English/Chinese).}
  \label{tab:datastat}
\end{table*} 

\begin{table*}[htbp]
\footnotesize
    \centering
    \begin{tabular}{|p{0.95\linewidth}|}
        \hline
        \textbf{Scene:} Xiaoming saw their class's little dog ``Dou Dou'' on the playground, with eyes smiling like crescent moons, jumping and jumping over. \\
        
        \colorbox{Emotion}{Ability-E1: Emotion/Typical emotional reactions} \\
        
        \textbf{Question-1:} Why did Xiaoming have such a reaction? \\
        \textbf{Options-1:} (A) He remembered something sad. \textbf{(B) He saw his favorite little animal.} (C) He was pushed down by other children. (D) He suddenly felt unwell. \\
        \textbf{Explanation-1:} Xiaoming showed a happy mood, and only option B reflected the happy mood, which is consistent with the typical reasons for emotional reactions. \\
        
        \hline
    \end{tabular}
    
    \caption{1 group of example data for \colorbox{Emotion}{Situation-based Emotion Knowledge Tasks [Typical]}.}
    \label{tab:e1_example}
\end{table*} 

\begin{table*}[htbp]
\footnotesize
    \centering
    \begin{tabular}{|p{0.95\linewidth}|}
        \hline
        \textbf{Scene:} Xiaomei won first place in the singing competition, and everyone applauded for her, but she lowered her head and tightly grasped the hem of her clothes with her fingers, looking very uncomfortable. \\
        
        \colorbox{Emotion}{Ability-E2: Emotion/Atypical emotional reactions} \\
        
        \textbf{Question-1:} Why did Xiaomei have such a reaction? \\
        \textbf{Options-1:} (A) She likes singing. (B) She forgot how to laugh. (C) She thinks the prize is not good enough. \textbf{(D) She is worried that her good friend will be jealous or disappointed.} \\
        \textbf{Explanation-1:} Xiaomei showed tension and difficulty, and only option D reflected the tension and difficulty, which is consistent with the causes of atypical emotional reactions. \\
        
        \hline
    \end{tabular}
    
    \caption{1 group of example data for \colorbox{Emotion}{Situation-based Emotion Knowledge Tasks [Atypical]}.}
    \label{tab:e2_example}
\end{table*} 

\begin{table*}[htbp]
\footnotesize
    \centering
    \begin{tabular}{|p{0.95\linewidth}|}
        \hline
        \textbf{Scene:} Xiao Ming, who loves extreme sports, pulled Xiao Hong on a roller coaster. During the ride, Xiao Ming screamed loudly while Xiao Hong closed her eyes tightly. \\
        
        \colorbox{Emotion}{Ability-E3: Emotion/Discrepant emotions} \\
        
        \textbf{Question-1:} How would Xiaoming feel? \\
        \textbf{Options-1:} \textbf{(A) Excitement.} (B) Surprise. (C) Afraid. (D) Regret. \\
        \textbf{Explanation-1:} Xiaoming kept screaming loudly, and he likes extreme sports, indicating that this incident has aroused Xiaoming's excitement, which is consistent with option A. \\

        \colorbox{Emotion}{Ability-E3: Emotion/Discrepant emotions} \\
        
        \textbf{Question-2:} How would Xiaoming feel? \\
        \textbf{Options-2:} (A) Excitement. (B) Surprise. \textbf{(C) Afraid.} (D) Regret. \\
        \textbf{Explanation-2:} Xiaohong closed her eyes tightly, indicating that this incident had caused Xiaohong to develop a fear opposite to that of Xiaoming, consistent with option C. \\
        
        \hline
    \end{tabular}
    
    \caption{1 group of example data for \colorbox{Emotion}{Affective Perspective-taking Test}.}
    \label{tab:e3_example}
\end{table*} 

\begin{table*}[htbp]
\footnotesize
    \centering
    \begin{tabular}{|p{0.95\linewidth}|}
        \hline
        \textbf{Scene:} Li Xiao and Wang Qiang teamed up to participate in an academic competition, and their team won first place, while Wang Qiang won the Best Paper Award. Li Xiao said to Wang Qiang, ``I'm really happy that my friend won the Best Paper Award!'' When she got home in the evening, Li Xiao said to her father, ``I'm very sad.'' \\
        
        \colorbox{Emotion}{Ability-E4: Emotion/Mixed emotions} \\
        
        \textbf{Question-1:} Why did Li Xiao say she was both happy and sad? \\
        \textbf{Options-1:} \textbf{(A) She is happy that Wang Qiang won the Best Paper award, but sad that she did not win the individual award.} (B) She was happy that their team won first place, but sad that Wang Qiang did not comfort her. (C) She was happy that Wang Qiang won the best paper, but sad that Wang Qiang did not comfort her. (D) She was happy that their team won first place, but sad that she didn't win an individual award. \\
        \textbf{Explanation-1:} Li Xiao's happy emotions are due to ``my friend won the Best Paper Award'', and his sad emotions are due to not winning the award himself. Option A simultaneously explains the reasons for both happy and sad emotions and interprets them correctly. \\
        
        \hline
    \end{tabular}
    
    \caption{1 group of example data for \colorbox{Emotion}{Strange Stories [Contrary Emotions]}.}
    \label{tab:e4_example}
\end{table*} 

\begin{table*}[htbp]
\footnotesize
    \centering
    \begin{tabular}{|p{0.95\linewidth}|}
        \hline
        \textbf{Scene:} In Chinese class, the teacher read out several grammatical errors in Yifan's essay in front of the whole class, and the students chuckled. The teacher asked him to stand up and read his essay, and he squeezed out a smile on his face, laughing along with everyone and saying, ``My writing is just a mess.'' \\
        
        \colorbox{Emotion}{Ability-E5: Emotion/Hidden emotions} \\
        
        \textbf{Question-1:} What is Yifan's true feeling? \\
        \textbf{Options-1:} (A) Happy. (B) Nervous. \textbf{(C) Sadness.} (D) Excitement. \\
        \textbf{Explanation-1:} This is a typical scenario where Yifan hides her negative emotions and pretends to be okay. Therefore, C should be chosen. \\

        \colorbox{Emotion}{Ability-E5: Emotion/Hidden emotions} \\
        
        \textbf{Question-2:} Why does Yifan seem indifferent or even a little happy on the surface? \\
        \textbf{Options-2:} (A) Because he doesn't care about his Chinese grades at all. (B) Because he finds it fun to be laughed at by everyone. \textbf{(C) Because he wants to pretend that he also finds it funny to cover up his embarrassment.} (D) Because he already knew he wrote very well. \\
        \textbf{Explanation-2:} Yifan's happy behavior hides negative emotions, and the reason he hides his emotions is that he doesn't want others to think he cares too much about his grades, which would make the situation more awkward. Therefore, C should be chosen. \\
        
        \hline
    \end{tabular}
    
    \caption{1 group of example data for \colorbox{Emotion}{Tests of Emotion Comprehension [Hidden Emotions]}.}
    \label{tab:e5_example}
\end{table*} 

\begin{table*}[htbp]
\footnotesize
    \centering
    \begin{tabular}{|p{0.95\linewidth}|}
        \hline
        \textbf{Scene:} Li Hua is a student who loves plants. He planted some rare flowers in the school garden. Zhao Min loves photography very much and took pictures of these flowers while passing by the school garden. Zhang Wei is a gardener at school. He saw Zhao Min taking photos of flowers without permission, but because he knew that Zhao Min's family was in difficulty and photography was her only hobby, he did not stop her. When Li Hua discovered that someone had taken unauthorized photos of his flowers and posted them on social media, he began to ask people around him. Finally, Zhao Min found out that those flowers belonged to Li Hua and admitted her actions. \\
        
        \colorbox{Emotion}{Ability-E6: Emotion/Moral emotions} \\
        
        \textbf{Question-1:} What emotions does Zhang Wei feel when he sees Zhao Min shooting flowers but doesn't stop her? \\
        \textbf{Options-1:} \textbf{(A) Zhang Wei may feel conflicted and sympathetic because he understands Zhao Min's predicament, but at the same time knows that her behavior is inappropriate.} (B) Zhang Wei may be satisfied because he believes that Zhao Min's photography can showcase the beauty of the school garden. (C) Zhang Wei may feel indifferent because he believes it is not his responsibility. (D) Zhang Wei may feel confused because he is unsure whether he should intervene in students' activities. \\
        \textbf{Explanation-1:} Option A reflects the contradiction and sympathy that Zhang Wei felt because he knew Zhao Min's family difficulties and did not stop her from filming in a timely manner, reflecting the relationship between emotions and moral thinking. \\
        
        \hline
    \end{tabular}
    
    \caption{1 group of example data for \colorbox{Emotion}{Tests of Emotion Comprehension [Moral Emotions]}.}
    \label{tab:e6_example}
\end{table*} 

\begin{table*}[htbp]
\footnotesize
    \centering
    \begin{tabular}{|p{0.95\linewidth}|}
        \hline
        \textbf{Scene:} Xiaohong and Xiaofang are watching other children play on the playground. They talked about some interesting things that happened on the playground and discussed going to the park together after school. Suddenly, Xiaohong signaled to Xiaofang and looked at Xiaomei on the swing. Then, Xiaohong smiled at Xiaofang again. Xiaofang nodded and the two walked towards Xiaomei. Xiaomei saw these two unfamiliar girls walking towards her. She noticed that they had exchanged glances and smiles before. Although the three of them were in the same class, Xiaomei had never spoken to them before. \\
        
        \colorbox{Emotion}{Ability-E7: Emotion/Emotion regulation} \\
        
        \textbf{Question-1:} What will Xiaomei do next to maintain emotional stability? \\
        \textbf{Options-1:} (A) Go straight to Xiaohong and Xiaofang and greet them proactively, asking if they need help or want to play together. (B) Ignore Xiaohong and Xiaofang and instead go talk to her best friend about her feelings. \textbf{(C) Imagine that Xiaohong and Xiaofang may just want to meet new friends, without any malicious intent.} (D) Ignore the behavior of Xiaohong and Xiaofang and focus on playing on the swing. \\
        \textbf{Explanation-1:} Xiaomei may experience confusion and fear after seeing the behavior of Xiaohong and Xiaofang. In order to maintain emotional stability, she needs to regulate her confusion and fear emotions, consistent with the emotion regulation behavior described in option C. \\
        
        \hline
    \end{tabular}
    
    \caption{1 group of example data for \colorbox{Emotion}{Tests of Emotion Comprehension [Emotion Regulation]}.}
    \label{tab:e7_example}
\end{table*} 

\begin{table*}[htbp]
\footnotesize
    \centering
    \begin{tabular}{|p{0.95\linewidth}|}
        \hline
        \textbf{Scene:} Xiaohua's little dog got lost, and before that, she often played pet balls with her little dog. Today, one year later, Xiaohua found a pet ball on the shelf while shopping in the store, and she felt very sad.\\
        
        \colorbox{Emotion}{Ability-E8: Emotion/Emotion based on memory} \\
        
        \textbf{Question-1:} Why does Xiaohua feel sad? \\
        \textbf{Options-1:} (A) Because the ball is new. \textbf{(B) Because the ball reminded her of her puppy.} (C) Because she doesn't like playing ball. (D) Because she wants to eat. \\
        \textbf{Explanation-1:} Xiaohua's little dog got lost before, which made her feel very sad. Now that she sees the pet ball, it reminds her of this incident and makes her feel sad. Only option B reflects the sadness she felt because she remembered the puppy when she saw the ball. \\
        
        \hline
    \end{tabular}
    
    \caption{1 group of example data for \colorbox{Emotion}{Tests of Emotion Comprehension [Memory-based Emotions]}.}
    \label{tab:e8_example}
\end{table*} 

\begin{table*}[htbp]
\footnotesize
    \centering
    \begin{tabular}{|p{0.95\linewidth}|}
        \hline
        \textbf{Scene:} Luna and Lluvia are a pair of twins who have raised a kitten together. Today Luna found that the kitten was missing. She searched all day but couldn't find it, thinking that the kitten was lost. In fact, it was Lluvia who took the kitten to the pet hospital for a routine check-up without informing Luna.\\
        
        \colorbox{Emotion}{Ability-E9: Emotion/Emotion based on belief} \\
        
        \textbf{Question-1:} What is Luna's mood at this moment? \\
        \textbf{Options-1:} (A) Happy. \textbf{(B) Sad.} (C) Calm. (D) Afraid. \\
        \textbf{Explanation-1:} Luna's belief that the kitten had gone missing caused her to feel sad, which is consistent with the description of option B. \\

        \colorbox{Emotion}{Ability-E9: Emotion/Emotion based on belief} \\

        \textbf{Question-2:} What is Lluvia's mood at this moment? \\
        \textbf{Options-2:} (A) Happy. (B) Sad. \textbf{(C) Calm.} (D) Afraid. \\
        \textbf{Explanation-2:} Lluvia took the kitten for a routine check-up and held different beliefs from Luna, which led to a different level of calm emotions, consistent with option C. \\
        
        \hline
    \end{tabular}
    
    \caption{1 group of example data for \colorbox{Emotion}{Tests of Emotion Comprehension [Belief-based Emotions]}.}
    \label{tab:e9_example}
\end{table*} 

\begin{table*}[htbp]
\footnotesize
    \centering
    \begin{tabular}{|p{0.95\linewidth}|}
        \hline
        \textbf{Scene:} You and Ayaka sit in the room, and she tastes the food in front of you one by one. She ate the carrot with a look of disgust on her face and said, ``It's so bad!'' Then she ate the broccoli, showing a happy expression and saying, ``It's delicious!'' Finally, she tasted the cookie and showed a painful expression again, saying, ``It's not good!'' After a while, she reached out her hand to you and said, ``Please give me a little more.''\\
        
        \colorbox{Desire}{Ability-D1: Desire/Discrepant desires} \\
        
        \textbf{Question-1:} You really enjoy eating cookies, but dislike eating broccoli and carrots. What food would you hand her? \\
        \textbf{Options-1:} \textbf{(A) Broccoli.} (B) Cookies. (C) Carrot. (D) Orange. \\
        \textbf{Explanation-1:} Ayaka's reaction indicates that she thinks broccoli is tasty while carrots and biscuits taste bad. Therefore, when she asks for food again, you should hand her the broccoli she likes, which has nothing to do with what I like. So choose A. \\
        
        \hline
    \end{tabular}
    
    \caption{1 group of example data for \colorbox{Desire}{Yummy-yucky Task}.}
    \label{tab:d1_example}
\end{table*} 

\begin{table*}[htbp]
\footnotesize
    \centering
    \begin{tabular}{|p{0.95\linewidth}|}
        \hline
        \textbf{Scene:} Xiaohong is a 6-year-old child. Today is Saturday. Mom and dad are free today, she doesn't know what they can do together. Maybe go eat ice cream? Xiaohong really wants to go to the amusement park today. However, my father believes that there will be a lot of noise in amusement parks. He said, ``Xiaohong, this is not a good idea. I think there will be a lot of noise in the amusement park.''.\\
        
        \colorbox{Desire}{Ability-D1: Desire/Discrepant desires} \\
        
        \textbf{Question-1:} How should Xiaohong persuade her father? \\
        \textbf{Options-1:} \textbf{(A) Xiaohong can search for some information to prove that amusement parks have taken many measures to reduce noise, such as installing soundproof walls and using quieter equipment.} (B) Xiaohong can tell her father that she hasn't been to an amusement park for a long time, which is a very special wish for her. She really wants to go. (C) If Dad really doesn't want to go to the amusement park, Xiaohong can suggest going to other places, such as parks or zoos, so that everyone is happy. (D) Xiaohong can suggest, 'How about we go on weekdays?'? I heard that there were fewer people and it was quieter back then. We can avoid peak hours and have a quieter experience. \\
        \textbf{Explanation-1:} First, since today is Sunday, Xiaohong's parents have the time to go to the amusement park with her. However, her father does not want to go to a place with a lot of noise. Therefore, Xiaohong needs to persuade her father that the noise in the amusement park is not that serious. So choose A. \\
        
        \hline
    \end{tabular}
    
    \caption{1 group of example data for \colorbox{Desire}{Persuasion Story Task}.}
    \label{tab:d2_example}
\end{table*} 

\begin{table*}[htbp]
\footnotesize
    \centering
    \begin{tabular}{|p{0.95\linewidth}|}
        \hline
        \textbf{Scene:} Little boy Tom found some money on his way home from school. He was happy and began to think about how to use the money. As he walked, he met his aunt. Auntie said to him, ``Tom, if you help me pick apples, I will give you some money.'' So Tom decided to stay at Auntie's house and help pick apples.\\
        
        \colorbox{Desire}{Ability-D2: Desire/Multiple desires} \\
        
        \textbf{Question-1:} What will Tom do after helping Auntie pick apples? \\
        \textbf{Options-1:} (A) Tom will go to the park to play. \textbf{(B) Tom will go to the store to buy things.} (C) Tom will assist Auntie in making apple pie. (D) Tom will donate the money to charitable organizations. \\
        \textbf{Explanation-1:} Before meeting the aunt, Tom was considering how to spend the money he found. After helping the aunt pick apples, he did not give up his desire to spend. The option that best fits this desire is going to the store to buy things, so choose B. \\
        
        \hline
    \end{tabular}
    
    \caption{1 group of example data for \colorbox{Desire}{Multiple Desires Task}.}
    \label{tab:d3_example}
\end{table*} 

\begin{table*}[htbp]
\footnotesize
    \centering
    \begin{tabular}{|p{0.95\linewidth}|}
        \hline
        \textbf{Scene:} Luna and Lluvia are twins, and today is their birthday. Dad thinks we can go eat pizza, Mom wants to cook for them at home. Mom and Dad asked for Luna and Lluvia's opinions, Luna wanted to eat hotpot, while Lluvia wanted to eat steamed fish.\\
        
        \colorbox{Desire}{Ability-D3: Desire/Desires influence on emotions and actions} \\
        
        \textbf{Question-1:} What would Luna do if she were given the final choice? \\
        \textbf{Options-1:} (A) Take the whole family to the pizza shop. \textbf{(B) Take the whole family to a hotpot restaurant.} (C) Take the whole family to the steam fish shop. (D) Stay at home. \\
        \textbf{Explanation-1:} Luna wants to eat hotpot, so she will take her whole family to a hotpot restaurant. Choose B. \\

        \colorbox{Desire}{Ability-D3: Desire/Desires influence on emotions and actions} \\
        
        \textbf{Question-2:} What would Lluvia do if she were given the final choice? \\
        \textbf{Options-2:} (A) Take the whole family to the pizza shop. (B) Take the whole family to a hotpot restaurant. \textbf{(C) Take the whole family to the steam fish shop.} (D) Stay at home. \\
        \textbf{Explanation-2:} Lluvia wants to eat steamed fish, so she will take her whole family to a steam fish restaurant. Choose C. \\
        
        \colorbox{Desire}{Ability-D3: Desire/Desires influence on emotions and actions} \\
        
        \textbf{Question-3:} What would Luna's mood be if her parents took them to a hotpot restaurant in the end? \\
        \textbf{Options-3:} \textbf{(A) Happy.} (B) Disappointed. (C) Calm. (D) Doubt. \\
        \textbf{Explanation-3:} Luna wants to eat hotpot, and she can eat hotpot by going to a hotpot restaurant, so she will be happy. Choose A. \\

        \colorbox{Desire}{Ability-D3: Desire/Desires influence on emotions and actions} \\

        \textbf{Question-4:} What would Lluvia's mood be if her parents took them to a hotpot restaurant in the end? \\
        \textbf{Options-4:} (A) Happy. \textbf{(B) Disappointed.} (C) Calm. (D) Doubt. \\
        \textbf{Explanation-4:} Lluvia wants to eat steam fish, but she cannot eat steam fish at a hotpot restaurant, so she will be disappointed. Choose B. \\

        \colorbox{Desire}{Ability-D3: Desire/Desires influence on emotions and actions} \\

        \textbf{Question-5:} If in the end, Mom and Dad take them to the steamed fish shop, who will be happy? \\
        \textbf{Options-5:} (A) Luna. \textbf{(B) Lluvia.} (C) Father. (D) Mother. \\
        \textbf{Explanation-5:} Lluvia wants to eat steam fish, and she can eat steam fish by going to a steam fish restaurant, so she will be happy. Choose B. \\
        
        \hline
    \end{tabular}
    
    \caption{1 group of example data for \colorbox{Desire}{Expanding Tasks [Twin Disagreement]}.}
    \label{tab:d4_example}
\end{table*} 

\begin{table*}[htbp]
\footnotesize
    \centering
    \begin{tabular}{|p{0.95\linewidth}|}
        \hline
        \textbf{Scene:} Xiaoli's father is a chef who often cooks delicious dishes at home. Xiaoli likes Braised pork belly cooked by her father very much. The fried meat is very delicious. It takes a lot of time to make Braised pork belly. One night, when her father asked Xiaoli if she wanted to eat Braised pork belly, she said ``no''.\\
        
        \colorbox{Desire}{Ability-D4: Desire/Desire-action contradiction} \\
        
        \textbf{Question-1:} Why did Xiaoli say she didn't want to eat Braised pork belly? \\
        \textbf{Options-1:} (A) Because she wants to lose weight. (B) Because she thinks Braised pork belly is unhealthy. (C) Because she ate too many snacks and is no longer hungry. \textbf{(D) Because she doesn't want to trouble her father by spending too much time.} \\
        \textbf{Explanation-1:} Xiaoli was worried that making red-braised pork would take up too much of her father's time, so she suppressed her desire to eat red-braised pork. Choose D. \\
        
        \hline
    \end{tabular}
    
    \caption{1 group of example data for \colorbox{Desire}{Strange Stories [Appearance/Reality]}.}
    \label{tab:d5_example}
\end{table*} 

\begin{table*}[htbp]
\footnotesize
    \centering
    \begin{tabular}{|p{0.95\linewidth}|}
        \hline
        \textbf{Scene:} You and Kazuha were standing in the courtyard, with Kazuha holding a sword in front of the bamboo. He swung it forward and the sword made a piercing sound in the air, narrowly missing the bamboo and scratching the white cat passing by. Kazuha believes that he did not achieve his goal just now.\\
        
        \colorbox{Intention}{Ability-I1: Intention/Completion of failed actions} \\
        
        \textbf{Question-1:} What behavior can you use to achieve Kazuha's goal? \\
        \textbf{Options-1:} \textbf{(A) Cut off bamboo with an axe.} (B) Pulling out a whip with a loud cracking sound. (C) Kill the white cat with a kitchen knife. (D) Performing sword dance in the courtyard with a sword in hand. \\
        \textbf{Explanation-1:} Kazuha's initial goal was to cut the bamboo. Option A fits the purpose of cutting the bamboo, so choose A. \\
        
        \hline
    \end{tabular}
    
    \caption{1 group of example data for \colorbox{Intention}{Behavioral Re-enactment Procedure}.}
    \label{tab:i1_example}
\end{table*} 

\begin{table*}[htbp]
\footnotesize
    \centering
    \begin{tabular}{|p{0.95\linewidth}|}
        \hline
        \textbf{Scene:} James showcased his paintings at the school art exhibition. One day, he forgot to bring the painting home. Nick mistakenly thought that the painting was abandoned while cleaning the exhibition room, so he put it in the waste bin. Meanwhile, Ella knew that this was James' work and her own paintings would also be displayed in the art exhibition. She didn't want James' paintings to be more popular than hers, so she decided not to tell Nick. When James realized that the painting was missing, he felt very frustrated and began to ask his classmates. Later, Nick realized that he might have mishandled James' painting.\\
        
        \colorbox{Intention}{Ability-I2: Intention/Discrepant intentions} \\
        
        \textbf{Question-1:} What kind of intention may have driven Nick's behavior? \\
        \textbf{Options-1:} (A) Nick may not have noticed its value because he was not interested in painting. \textbf{(B) Nick may be unconscious or unaware because he mistakenly believed that James' work was abandoned.} (C) Nick may think that cleaning the exhibition room is more important than preserving the artwork. (D) Nick may have overlooked the importance of the work because he wanted to complete the task quickly. \\
        \textbf{Explanation-1:} Nick did not know that this painting was James's work and mistakenly thought it was an abandoned painting, so choose B. \\

        \colorbox{Intention}{Ability-I2: Intention/Discrepant intentions} \\

        \textbf{Question-2:} What kind of intention could Ella's behavior be? \\
        \textbf{Options-2:} \textbf{(A) Ella may have chosen to remain silent to see James in a difficult situation due to jealousy of his talent.} (B) Ella may not have had time to inform Nick that James' painting was mishandled due to her busy work schedule. (C) Ella believed that Nick would realize his mistake and correct it on his own, so she chose not to tell him. (D) Ella may not appreciate James' paintings and believes that they should not be displayed in art exhibitions. \\
        \textbf{Explanation-2:} Ella knew that this was James's work, but she did not want James's painting to be more popular than hers. She was actually jealous of James's talent, so she didn't tell Nick. Therefore, choose A. \\
        
        \hline
    \end{tabular}
    
    \caption{1 group of example data for \colorbox{Intention}{Expanding Tasks [Accomplice]}.}
    \label{tab:i2_example}
\end{table*} 

\begin{table*}[htbp]
\footnotesize
    \centering
    \begin{tabular}{|p{0.95\linewidth}|}
        \hline
        \textbf{Scene:} Xiao Wang is in the hospital duty room, and the doctor asked him to add a ``supplementary fluid'' to the infusion bottle.\\
        
        \colorbox{Intention}{Ability-I2: Intention/Discrepant intentions} \\
        
        \textbf{Question-1:} Which of the following four situations does the harm of Xiao Wang's intention and outcome differ? \\
        \textbf{Options-1:} (A) The supplementary solution is physiological saline. Xiao Wang saw the label that read 'regular supplement', added it to the infusion bottle, and the patient's condition stabilized after infusion. (B) Supplementing fluid is an excess of insulin. Xiao Wang saw a label that read 'Warning: fatal overdose' and added it to the infusion bottle. The patient experienced severe hypoglycemia after infusion. (C) Xiao Wang did not hear the doctor's request. \textbf{(D) The supplementary solution is physiological saline. Xiao Wang saw the label that read 'Warning: fatal overdose', added the infusion bottle, and the patient's condition stabilized after infusion.} \\
        \textbf{Explanation-1:} In option D, Xiao Wang wanted to do something bad to cause the patient's condition to deteriorate, but the patient's condition remained stable. Xiao Wang's intention was harmful, but the result was harmless; they were inconsistent. Therefore, choose D. \\

        \colorbox{Intention}{Ability-I2: Intention/Discrepant intentions} \\

        \textbf{Question-2:} In which of the following four situations is the intention of Xiao Wang consistent with the harmfulness of the outcome? \\
        \textbf{Options-2:} (A) Xiao Wang did not hear the doctor's request. \textbf{(B) Supplementing fluid is an excess of insulin. Xiao Wang saw a label that read 'Warning: fatal overdose' and added it to the infusion bottle. The patient experienced severe hypoglycemia after infusion.} (C) Supplementing fluid is an excess of insulin. Xiao Wang saw the label saying 'regular supplement' added to the infusion bottle, and the patient experienced severe hypoglycemia after infusion. (D) The supplementary solution is physiological saline. Xiao Wang saw the label that read 'Warning: fatal overdose', added the infusion bottle, and the patient's condition stabilized after infusion. \\
        \textbf{Explanation-2:} In option B, Xiao Wang wanted to do something bad to cause the patient's condition to deteriorate, and after the infusion, the patient's condition did indeed deteriorate. Therefore, both the intention and the result were harmful, so choose B. \\
        
        \hline
    \end{tabular}
    
    \caption{1 group of example data for \colorbox{Intention}{Expanding Tasks [Harmful Intention]}.}
    \label{tab:i3_example}
\end{table*} 

\begin{table*}[htbp]
\footnotesize
    \centering
    \begin{tabular}{|p{0.95\linewidth}|}
        \hline
        \textbf{Scene:} Zhiwei and Xiaolan are planning a short trip, sitting on a bench in the park flipping through a travel guide. Although Mingda was not invited to participate in this short trip, he still listened to their discussion behind them. Suddenly, Zhiwei noticed Mingda standing aside and glanced at Xiaolan. Then, Xiaolan saw Zhiwei's gaze, shook her head, and continued to look down at the travel guide.\\
        
        \colorbox{Intention}{Ability-I3: Intention/Prediction of actions} \\
        
        \textbf{Question-1:} What is Zhiwei's next action after noticing Mingda? \\
        \textbf{Options-1:} (A) Ignore Mingda and continue discussing travel plans with Xiaolan. (B) Invite Mingda to join their discussion. \textbf{(C) Propose to Xiaolan to find a different place to avoid discussing travel plans with Mingda.} (D) Tell Xiaolan about Mingda. \\
        \textbf{Explanation-1:} Zhiwei's behavior indicates that he does not want Mingda to hear their plan, and Option C can achieve this purpose. \\
        
        \hline
    \end{tabular}
    
    \caption{1 group of example data for \colorbox{Intention}{Expanding Tasks [Action Prediction]}.}
    \label{tab:i4_example}
\end{table*} 

\begin{table*}[htbp]
\footnotesize
    \centering
    \begin{tabular}{|p{0.95\linewidth}|}
        \hline
        \textbf{Scene:} Chen Wei is a college student from Guangdong. One year, there was a bird flu outbreak in Guangdong, and in the supermarket outside the university, a large New Orleans roasted chicken was sold at 50\(\%\) off for less than ten yuan. Chen Wei excitedly bought one. When checking out, the aunt behind him asked him, ``Young man, do you know any news recently.''\\
        
        \colorbox{Intention}{Ability-I4: Intention/Intentions explanations} \\
        
        \textbf{Question-1:} What did Auntie really mean when she said that? \\
        \textbf{Options-1:} (A) Auntie wants to ask Chen Wei for his opinion on political news. (B) Auntie was chatting casually and wanted to know about the latest news. \textbf{(C) Auntie hinted that buying roasted chicken during the season of avian influenza epidemic is not safe.} (D) Auntie wants to point out that the price of New Orleans roasted chicken is too cheap. \\
        \textbf{Explanation-1:} The background is that bird flu is spreading in Guangdong. Option C indicates that the auntie is hinting at this, making it the most reasonable choice. \\
        
        \hline
    \end{tabular}
    
    \caption{1 group of example data for \colorbox{Intention}{Hinting Task}.}
    \label{tab:i5_example}
\end{table*} 

\begin{table*}[htbp]
\footnotesize
    \centering
    \begin{tabular}{|p{0.95\linewidth}|}
        \hline
        \textbf{Scene:} Business tycoon Mr. Zhao recently encountered difficulties and failures, with huge losses, and is preparing to bid for a highly valuable piece of land in the city. On the eve of the bidding, he invited several competitors to a dinner together. At the dinner party, I talked extensively about my recent business failures and difficulties with these competitors. He showed such sincerity that several competitors sympathized with him, thinking that his bid was definitely hopeless.\\
        
        \colorbox{Intention}{Ability-I4: Intention/Intentions explanations} \\
        
        \textbf{Question-1:} Why did Mr. Zhao say that? \\
        \textbf{Options-1:} (A) Mr. Zhao said this because he has truly encountered business failures and difficulties and hopes that his competitors can provide support. \textbf{(B) Mr. Zhao said this in order to conceal his true intentions, let his competitors relax their vigilance towards him, and thus gain an advantage in the bidding.} (C) Mr. Zhao said this because he wants to establish closer cooperation with competitors and jointly tackle market challenges. (D) Mr. Zhao deliberately showed off his wealth and success at the dinner party by appearing generous and grand. \\
        \textbf{Explanation-1:} Boss Zhao's true intention is to buy the land. His complaining makes his competitors feel that he has no hope of winning the bid, which causes them to lower their guard. Therefore, choose B. \\
        
        \hline
    \end{tabular}
    
    \caption{1 group of example data for \colorbox{Intention}{Strange Stories [Persuade]}.}
    \label{tab:i6_example}
\end{table*} 

\begin{table*}[htbp]
\footnotesize
    \centering
    \begin{tabular}{|p{0.95\linewidth}|}
        \hline
        \textbf{Scene:} Zhang Haoyu is a very untidy boy. One day, his mother walked into his bedroom, even more chaotic than usual! Clothes, toys, and comics are everywhere. Zhang Haoyu's mother said to him, ``This room is like a pigsty.''\\
        
        \colorbox{Intention}{Ability-I4: Intention/Intentions explanations} \\
        
        \textbf{Question-1:} Why did Zhang Haoyu's mother say that? \\
        \textbf{Options-1:} \textbf{(A) Zhang Haoyu's mother said this was because she used a metaphor to compare the level of chaos in Zhang Haoyu's room to a pigsty, emphasizing the disorder and chaos in the room.} (B) Zhang Haoyu's mother said it was because she believed William had raised a pig. (C) Zhang Haoyu's mother said it's because she thinks there are many comic books and toys in the pigsty. (D) Zhang Haoyu's mother said it's because she thinks the pigsty is as clean as Zhang Haoyu's room. \\
        \textbf{Explanation-1:} Zhang Haoyu's room was too messy, and his mother's true intention was to use a pigsty as a metaphor for it. Therefore, choose A. \\
        
        \hline
    \end{tabular}
    
    \caption{1 group of example data for \colorbox{Intention}{Strange Stories [Figure of Speech]}.}
    \label{tab:i7_example}
\end{table*} 

\begin{table*}[htbp]
\footnotesize
    \centering
    \begin{tabular}{|p{0.95\linewidth}|}
        \hline
        \textbf{Scene:} In the self-study room of the library, you stand in front of a table with a six sided dice placed on it. The sum of the points on the opposite sides of the dice is 7. The six faces of the dice, from 1 to 6, are the sun, moon, stars, clouds, umbrellas, and snowflakes. Wang Wei and Li Na stood on either side of the table. Zhang Qiang is standing opposite you.\\
        
        \colorbox{Percept}{Ability-P1: Percept/Simple visual perspective taking} \\
        
        \textbf{Question-1:} You see the sun. What did Zhang Qiang see? \\
        \textbf{Options-1:} \textbf{(A) Snowflake.} (B) Umbrella. (C) Moon. (D) Star. \\
        \textbf{Explanation-1:} The opposite side of the sun pattern should be the snowflake pattern, so choose A. \\

        \colorbox{Percept}{Ability-P1: Percept/Simple visual perspective taking} \\
        
        \textbf{Question-2:} Wang Wei saw the moon. What did Li Na see? \\
        \textbf{Options-2:} (A) Cloud. \textbf{(B) Umbrella.} (C) Snowflake. (D) Sun. \\
        \textbf{Explanation-2:} The opposite side of the moon pattern should be the unbrella pattern, so choose B. \\
        
        \hline
    \end{tabular}
    
    \caption{1 group of example data for \colorbox{Percept}{Picture Identification Task}.}
    \label{tab:p1_example}
\end{table*} 

\begin{table*}[htbp]
\footnotesize
    \centering
    \begin{tabular}{|p{0.95\linewidth}|}
        \hline
        \textbf{Scene:} A regular quadrilateral table with different items placed in all four corners. You stand in front of the table, with Xiao Zhou, Xiao He, and Xiao Zhao standing on the left, right, and opposite sides of the table, respectively. You see on the table from near to far, with pencils and mice placed in the first row from left to right, and water bottles and apples placed in the second row from left to right.\\
        
        \colorbox{Percept}{Ability-P2: Percept/Complex visual perspective taking} \\
        
        \textbf{Question-1:} What would you see from Xiao Zhou's perspective? \\
        \textbf{Options-1:} (A) The first row is arranged from left to right with apples and water bottles, while the second row is arranged from left to right with mice and pencils. \textbf{(B) The first row is filled with water bottles and pencils from left to right, while the second row is filled with apples and a mouse from left to right.} (C) The first row is filled with mice and apples from left to right, while the second row is filled with pencils and water bottles from left to right. (D) The first row is arranged from left to right with pencils and mice, while the second row is arranged from left to right with water bottles and apples. \\
        \textbf{Explanation-1:} Xiao Zhou's first row left is your second row left, his first row right is your first row left, his second row left is your second row right, and his second row right is your first row right. So choose B. \\

        \colorbox{Percept}{Ability-P2: Percept/Complex visual perspective taking} \\
        
        \textbf{Question-2:} What would you see from Xiao He's perspective? \\
        \textbf{Options-2:} \textbf{(A) The first row is filled with mice and apples from left to right, while the second row is filled with pencils and water bottles from left to right.} (B) The first row is arranged from left to right with pencils and mice, while the second row is arranged from left to right with water bottles and apples. (C) The first row is filled with water bottles and pencils from left to right, while the second row is filled with apples and a mouse from left to right. (D) The first row is arranged from left to right with apples and water bottles, while the second row is arranged from left to right with mice and pencils. \\
        \textbf{Explanation-2:} Xiao He's first row left is your first row right, his first row right is your second row right, his second row left is your first row left, and his second row right is your second row left. So choose A. \\

        \colorbox{Percept}{Ability-P2: Percept/Complex visual perspective taking} \\
        
        \textbf{Question-3:} What would you see from Xiao Zhao's perspective? \\
        \textbf{Options-3:} \textbf{(A) The first row is arranged from left to right with apples and water bottles, while the second row is arranged from left to right with mice and pencils.} (B) The first row is filled with water bottles and pencils from left to right, while the second row is filled with apples and a mouse from left to right. (C) The first row is filled with mice and apples from left to right, while the second row is filled with pencils and water bottles from left to right. (D) The first row is arranged from left to right with pencils and mice, while the second row is arranged from left to right with water bottles and apples. \\
        \textbf{Explanation-3:} Xiao Zhao's first row left is your second row right, his first row right is your second row left, his second row left is your first row right, and his second row right is your first row left. So choose A. \\
        
        \hline
    \end{tabular}
    
    \caption{1 group of example data for \colorbox{Percept}{Spatial Construction Task}.}
    \label{tab:p2_example}
\end{table*} 

\begin{table*}[htbp]
\footnotesize
    \centering
    \begin{tabular}{|p{0.95\linewidth}|}
        \hline
        \textbf{Scene:} Tara is a curious robot living in an underwater city called Aquatica in the deep sea. In Aquatica, there are no birds or flying creatures, and Tara has never known about them. However, Aquatica is filled with a variety of marine creatures. Tara is engaged in imitative behavior: it elegantly swings its arms up and down, as if moving forward, very similar to the flapping of bird wings.\\
        
        \colorbox{Knowledge}{Ability-K1: Knowledge/Knowledge-pretend play links} \\
        
        \textbf{Question-1:} What might Tara be imitating? \\
        \textbf{Options-1:} \textbf{(A) Fish with sliding fins.} (B) Soaring Eagle. (C) A butterfly flapping its wings. (D) Bats in flight. \\
        \textbf{Explanation-1:} There are no birds or flying animals where Tara lives, but there is marine life. Therefore, the animal in Option A is the only one Tara could have seen and imitated. \\
        
        \hline
    \end{tabular}
    
    \caption{1 group of example data for \colorbox{Knowledge}{Sarah Task}.}
    \label{tab:k1_example}
\end{table*} 

\begin{table*}[htbp]
\footnotesize
    \centering
    \begin{tabular}{|p{0.95\linewidth}|}
        \hline
        \textbf{Scene:} Rain stick is a traditional musical instrument originating from South America, which uses a sun dried hollow bamboo stem. After hollowing out the middle, it is filled with fine particles such as seeds and sand and sealed. By repeatedly reversing it, it emits a sound similar to rain. Now you are using a rain stick in front of a blind person and a deaf person. \\
        
        \colorbox{Knowledge}{Ability-K2: Knowledge/Percepts-knowledge links} \\
        
        \textbf{Question-1:} What do blind people most likely think happened? \\
        \textbf{Options-1:} \textbf{(A) It's raining outside now.} (B) You are using a rain stick to simulate the sound of rain. (C) You are performing stick techniques in traditional martial arts. (D) You're playing the piano. \\
        \textbf{Explanation-1:} Blind people cannot see the form of musical instruments and can only hear the sound of rain, so they are most likely to recognize that it is raining outside. Choose A. \\

        \colorbox{Knowledge}{Ability-K2: Knowledge/Percepts-knowledge links} \\
        
        \textbf{Question-2:} What do deaf people most likely think happened? \\
        \textbf{Options-2:} (A) It's raining outside now. (B) You are using a rain stick to simulate the sound of rain. \textbf{(C) You are performing stick techniques in traditional martial arts.} (D) You're playing the piano. \\
        \textbf{Explanation-2:} Deaf people cannot hear the sound of musical instruments and can only see performers playing with a stick, so they are most likely to recognize it as a stick art performance. Choose C. \\
        
        \hline

        \textbf{Scene:} You stir fried a plate of sugar stir fried ginger shreds, which is not a common dish, but the way it is served is very similar to shredded potatoes. You take a photo of this dish and share it with Xiaohong in another city, then hold it in front of Xiaogang and cover his eyes, letting him smell the aroma of this dish. \\
        
        \colorbox{Knowledge}{Ability-K2: Knowledge/Percepts-knowledge links} \\
        
        \textbf{Question-1:} What would Xiaogang most likely think this dish is? \\
        \textbf{Options-1:} (A) Stir fried ginger shreds with sugar. \textbf{(B) Ginger Coke.} (C) Shredded Potatoes with Vinegar Sauce. (D) Sauteed Potato, Green Pepper and Eggplant. \\
        \textbf{Explanation-1:} Xiao Gang cannot see the shape of the dishes and can only smell the strong ginger flavor, so he is most likely to recognize them as more common ginger products such as ginger cola, and choose option B. \\

        \colorbox{Knowledge}{Ability-K2: Knowledge/Percepts-knowledge links} \\
        
        \textbf{Question-2:} What would Xiaohong most likely think this dish is? \\
        \textbf{Options-2:} (A) Stir fried ginger shreds with sugar. (B) Ginger Coke. \textbf{(C) Shredded Potatoes with Vinegar Sauce.} (D) Sauteed Potato, Green Pepper and Eggplant. \\
        \textbf{Explanation-2:} Xiao Gang couldn't smell the taste of the vegetables and could only see a dish resembling shredded potatoes in vinegar. Therefore, he is most likely to recognize it as the common shredded potatoes in vinegar and choose option C. \\
        
        \hline

        \textbf{Scene:} You are using a distiller to purify alcohol. During this process, two blindfolded people are next to you, one of whom is deaf and the other is a person with impaired smell. \\
        
        \colorbox{Knowledge}{Ability-K2: Knowledge/Percepts-knowledge links} \\
        
        \textbf{Question-1:} What do deaf people most likely think you're doing? \\
        \textbf{Options-1:} (A) Boil a kettle of water using a kettle. (B) Purification of alcohol using a distiller. (C) Configure precise solutions. \textbf{(D) Open a bottle of high proof liquor to drink.} \\
        \textbf{Explanation-1:} Deaf people cannot hear the sound of the distiller and can only smell the ethanol odor emitted during the distillation process, so they are more likely to recognize more common scenarios such as drinking alcohol and choosing D. \\

        \colorbox{Knowledge}{Ability-K2: Knowledge/Percepts-knowledge links} \\
        
        \textbf{Question-2:} What do people with olfactory dysfunction most likely think you are doing? \\
        \textbf{Options-2:} \textbf{(A) Boil a kettle of water using a kettle.} (B) Purification of alcohol using a distiller. (C) Configure precise solutions. (D) Open a bottle of high proof liquor to drink. \\
        \textbf{Explanation-2:} People with olfactory dysfunction cannot smell the smell of alcohol and can only hear the boiling sound produced by the distillation process, so they are more likely to recognize more common scenarios such as boiling water. Choose A. \\
        
        \hline
    \end{tabular}
    
    \caption{3 groups of example data for \colorbox{Knowledge}{Expanding Tasks [Synesthetic Fallacy]}.}
    \label{tab:k2_example}
\end{table*} 

\begin{table*}[htbp]
\footnotesize
    \centering
    \begin{tabular}{|p{0.95\linewidth}|}
        \hline
        \textbf{Scene:} You live in the North China region. Entering winter, it snows. You need to write a letter to your pen pal and share your experience of building a snowman.\\
        
        \colorbox{Knowledge}{Ability-K3: Knowledge/Information-knowledge links} \\
        
        \textbf{Question-1:} You need to write a letter to Susan who lives in high latitude areas. How should you mention 'snow' in the letter? \\
        \textbf{Options-1:} (A) Snow is a white, cold thing that floats down from the clouds. \textbf{(B) I made a super big snowman yesterday!} (C) You may not have seen snow before, it only falls in winter. (D) Let me explain to you first what snow is. \\
        \textbf{Explanation-1:} Friends in high-latitude regions must have seen snow, so there is no need to explain or introduce it. Therefore, choose B. \\

        \colorbox{Knowledge}{Ability-K3: Knowledge/Information-knowledge links} \\
        
        \textbf{Question-2:} You need to write a letter to Juanita who lives in the equatorial region. How should you mention 'snow' in the letter? \\
        \textbf{Options-2:} (A) Last week's heavy snow was so fun! (B) I made a super big snowman yesterday! \textbf{(C) Snow is small white ice crystals that fall from the sky, soft like cotton, both icy and cool.} (D) Do you like to have snowball fights? \\
        \textbf{Explanation-2:} Friends in equatorial regions have not seen snow, so you need to introduce it first to let them know what it looks like. Therefore, choose C. \\
        
        \hline
    \end{tabular}
    
    \caption{1 group of example data for \colorbox{Knowledge}{Awareness of a Reader’s Knowledge Task}.}
    \label{tab:k3_example}
\end{table*} 

\begin{table*}[htbp]
\footnotesize
    \centering
    \begin{tabular}{|p{0.95\linewidth}|}
        \hline
        \textbf{Scene:} In a box of 24 candies, Liu Qian knew that the salesperson said almost one-third were filled with gummies. Customer Chen Hao first took a look at 12 of them from the box, and then said to Liu Qian, ``Among these 12, 2 of them are stuffed gummies.''\\
        
        \colorbox{Knowledge}{Ability-K3: Knowledge/Information-knowledge links} \\
        
        \textbf{Question-1:} After Chen Hao's report, Liu Qian will speculate how many of these 24 candies are actually filled with gummies? \\
        \textbf{Options-1:} (A) It is speculated that 14 of them are stuffed gummies. \textbf{(B) It is speculated that 7 of them are sandwich gummies.} (C) It is speculated that 4 of them are sandwich gummies. (D) It is speculated that 2 of them are sandwich gummies. \\
        \textbf{Explanation-1:} Out of 12 candies, 2 are filled gummies, so out of 24 candies, there should be 2-14 filled gummies. Based on Liu Qian's existing knowledge, 7 sandwich gummies are close to one-third of 24 and within this range. Therefore, choose B. \\

        \colorbox{Knowledge}{Ability-K3: Knowledge/Information-knowledge links} \\
        
        \textbf{Question-2:} Chen Hao would speculate how many of these 24 candies are actually filled with gummies? \\
        \textbf{Options-2:} (A) It is speculated that 14 of them are stuffed gummies. (B) It is speculated that 7 of them are sandwich gummies. \textbf{(C) It is speculated that 4 of them are sandwich gummies.} (D) It is speculated that 2 of them are sandwich gummies. \\
        \textbf{Explanation-2:} Out of 12 candies, 2 are filled gummies, so out of 24 candies, there should be 2-14 filled gummies. Chen Hao has no other information, so based on the proportion, he is most likely to think that there are 4 sandwich gummies out of 24 candies. Therefore, choose C. \\
        
        \hline
    \end{tabular}
    
    \caption{1 group of example data for \colorbox{Knowledge}{Scalar Implicature Task}.}
    \label{tab:k4_example}
\end{table*} 

\begin{table*}[htbp]
\footnotesize
    \centering
    \begin{tabular}{|p{0.95\linewidth}|}
        \hline
        \textbf{Scene:} On a sunny weekend afternoon, Li Juan, Wang Qiang, and Xiao Fang were playing with toys at home together. They first played the first novel toy - a maze 3D puzzle, and then played the second novel toy together - an electric flying machine. At this moment, Li Juan said, ``I have something to do and need to leave the room,'' and then closed the door before leaving. Wang Qiang said, ``Oh, she's gone, it's okay, let's play together.'' Wang Qiang and Xiaofang played with the third novel toy - the smart rotating colorful light. After playing, they placed all three toys on a plate on the table. At this moment, Li Juan came back and looked at the three toys in the tray, excitedly saying, ``Oh, look! Look at that toy!'' and ``Xiaofang, can you pass it to me?''\\
        
        \colorbox{Knowledge}{Ability-K4: Knowledge/Knowledge-attention links} \\
        
        \textbf{Question-1:} What is Xiaofang most likely to do? \\
        \textbf{Options-1:} (A) Xiaofang picked out the electric flying machine and handed it over to Li Juan. \textbf{(B) Xiaofang picked out the smart rotating colorful light and handed it to Li Juan.} (C) Xiaofang randomly took one toy from three and handed it to Li Juan. (D) Xiaofang picked out the 3D puzzle of the maze and handed it to Li Juan. \\
        \textbf{Explanation-1:} For Li Juan, both the 3D maze puzzle and the electric flying machine are toys she has already played with. Therefore, her attention should be on the toys she hasn't played with yet, so choose B. \\
        
        \hline
    \end{tabular}
    
    \caption{1 group of example data for \colorbox{Knowledge}{Familiary-focus of Attention Task}.}
    \label{tab:k5_example}
\end{table*} 

\begin{table*}[htbp]
\footnotesize
    \centering
    \begin{tabular}{|p{0.95\linewidth}|}
        \hline
        \textbf{Scene:} Xiaogang found a handbag in the bedroom, with the label on it being cabbage. Xiaogang couldn't see what was inside the handbag, so he opened it and found the hat. There was no cabbage in the handbag, so Xiaogang closed it and put it back in its original place. Xiaoming entered the bedroom and saw the handbag.\\
        
        \colorbox{Belief}{Ability-B1: Belief/Content false beliefs} \\
        
        \textbf{Question-1:} After Xiaoming opened his handbag, what did Xiaogang expect to find inside? \\
        \textbf{Options-1:} (A) Plate ruler. (B) Cabbage. (C) Compass. \textbf{(D) Hat.} \\
        \textbf{Explanation-1:} Xiaogang knows that there is a hat in the handbag, so he expects to find a hat. This corresponds to Option D. \\

        \colorbox{Belief}{Ability-B1: Belief/Content false beliefs} \\
        
        \textbf{Question-2:} After Xiaoming opened his handbag, what did Xiaoming expect to find inside? \\
        \textbf{Options-2:} (A) Plate ruler. \textbf{(B) Cabbage.} (C) Compass. (D) Hat. \\
        \textbf{Explanation-2:} Xiaoming does not know that there is actually a hat in the handbag, so he expects to find a cabbage. This corresponds to Option B. \\

        \colorbox{Belief}{Ability-B3: Belief/Second-order belief} \\
        
        \textbf{Question-3:} After Xiaoming opened his handbag, Xiaogang wondered what Xiaoming expected to find inside? \\
        \textbf{Options-3:} (A) Plate ruler. \textbf{(B) Cabbage.} (C) Compass. (D) Hat. \\
        \textbf{Explanation-3:} Xiaogang is able to recognize that Xiaoming does not know there is actually a hat in the handbag, so he thinks that Xiaoming expects to find a cabbage. Therefore, the answer is B. \\

        \colorbox{Belief}{Ability-B3: Belief/Second-order belief} \\
        
        \textbf{Question-4:} After Xiaoming opened his handbag, Xiaoming wondered what Xiaogang expected to find inside? \\
        \textbf{Options-4:} (A) Plate ruler. \textbf{(B) Cabbage.} (C) Compass. (D) Hat. \\
        \textbf{Explanation-4:} Xiaoming does not know that there is a hat in the handbag, so he thinks that Xiaogang also expects to find a cabbage. Therefore, the answer is B. \\
        
        \hline
    \end{tabular}
    
    \caption{1 group of example data for \colorbox{Belief}{False Belief Tasks [Content]}.}
    \label{tab:b1_example}
\end{table*} 

\begin{table*}[htbp]
\footnotesize
    \centering
    \begin{tabular}{|p{0.95\linewidth}|}
        \hline
        \textbf{Scene:} Xiaogang and Xiaoming were strolling in the bedroom when they saw a handbag, briefcase, and backpack. They found cabbage in the handbag, and Xiaoming left the bedroom. Xiaogang moved the cabbage to the backpack.\\
        
        \colorbox{Belief}{Ability-B2: Belief/Location false beliefs} \\
        
        \textbf{Question-1:} Where will Xiaogang look for cabbage after Xiaoming returns to the bedroom? \\
        \textbf{Options-1:} \textbf{(A) Backpack.} (B) Briefcase. (C) Plastic bag. (D) Handbag. \\
        \textbf{Explanation-1:} Xiaogang knows the cabbage is in the backpack, so he will look for it in the backpack. Therefore, choose Option A. \\

        \colorbox{Belief}{Ability-B2: Belief/Location false beliefs} \\
        
        \textbf{Question-2:} Where will Xiaoming look for cabbage after Xiaoming returns to the bedroom? \\
        \textbf{Options-2:} (A) Backpack. (B) Briefcase. (C) Plastic bag. \textbf{(D) Handbag.} \\
        \textbf{Explanation-2:} Xiaoming does not know that Xiaogang moved the cabbage to the backpack, so he will look for the cabbage in the handbag. Therefore, choose Option D. \\

        \colorbox{Belief}{Ability-B3: Belief/Second-order belief} \\
        
        \textbf{Question-3:} After Xiaoming returned to the bedroom, where did Xiaoming think Xiaogang would look for cabbage? \\
        \textbf{Options-3:} (A) Backpack. (B) Briefcase. (C) Plastic bag. \textbf{(D) Handbag.} \\
        \textbf{Explanation-3:} Xiaoming does not know that Xiaogang moved the cabbage to the backpack, therefore he thinks that Xiaogang will also look for the cabbage in the handbag. So choose D. \\

        \colorbox{Belief}{Ability-B3: Belief/Second-order belief} \\
        
        \textbf{Question-4:} After Xiaoming returned to the bedroom, where did Xiaogang think Xiaoming would look for cabbage? \\
        \textbf{Options-4:} (A) Backpack. (B) Briefcase. (C) Plastic bag. \textbf{(D) Handbag.} \\
        \textbf{Explanation-4:} Xiaogang is able to recognize that Xiaoming does not know the cabbage is in the backpack, so Xiaogang thinks that Xiaoming will look for the cabbage in the handbag. Therefore, choose D. \\
        
        \hline
    \end{tabular}
    
    \caption{1 group of example data for \colorbox{Belief}{False Belief Tasks [Location]}.}
    \label{tab:b2_example}
\end{table*} 

\begin{table*}[htbp]
\footnotesize
    \centering
    \begin{tabular}{|p{0.95\linewidth}|}
        \hline
        \textbf{Scene:} Joe and Anna are setting the table for the holiday dinner in the restaurant. Anna accidentally spilled water on Joe's new shirt while pouring water for him. Anna placed the water cup on the table and left the restaurant to search for tissues. After Anna left, Joe took out a handkerchief and wiped his shirt and table dry. Anna secretly saw Joe's movements through the crack in the door, so she didn't bring back any tissues and returned to the restaurant.\\
        
        \colorbox{Belief}{Ability-B3: Belief/Second-order belief} \\
        
        \textbf{Question-1:} Before Anna returned to the restaurant, Joe thought Anna would think the shirt was in what condition? \\
        \textbf{Options-1:} \textbf{(A) Still wet.} (B) Already dry. (C) Replaced by Joe. (D) I don't know. \\
        \textbf{Explanation-1:} Joe does not know about Anna's act of peeking, so he does not know that Anna already knows he dried his shirt with a handkerchief. Therefore, Joe will think that Anna believes his shirt is still wet. Thus, choose Option A. \\

        \colorbox{Belief}{Ability-B3*: Belief/High-order belief} \\
        
        \textbf{Question-2:} Anna knows what Joe thinks she thinks about the state of the shirt? \\
        \textbf{Options-2:} \textbf{(A) Still wet.} (B) Already dry. (C) Replaced by Joe. (D) I don't know. \\
        \textbf{Explanation-2:} Anna knows that Joe does not know she already knows the shirt has been wiped dry. Therefore, Anna knows that Joe will assume she believes the shirt is still wet. So choose Option A. \\

        \colorbox{Belief}{Ability-B3: Belief/Second-order belief} \\
        
        \textbf{Question-3:} When Joe saw that Anna was not holding paper in her hand, what did he think Anna would think of the state of the shirt now? \\
        \textbf{Options-3:} (A) Still wet. \textbf{(B) Already dry.} (C) Replaced by Joe. (D) I don't know. \\
        \textbf{Explanation-3:} After Joe sees that Anna did not get a tissue, he realizes that Anna already knows his shirt is dry. Therefore, he now thinks that Anna believes the shirt is dry. So choose Option B. \\

        \hline
        \textbf{Scene:} Joe and Anna are setting the table for the holiday dinner in the restaurant. Anna accidentally spilled water on Joe's new shirt while pouring water for him. Anna placed the water cup on the table and left the restaurant to search for tissues, while Joe calmly said it was okay. After Anna left, Joe sighed at his clothes inside the room. Anna secretly saw Joe's behavior through the crack in the door and felt very guilty. After returning, she showed a troubled expression.\\
        
        \colorbox{Belief}{Ability-B3: Belief/Second-order belief} \\
        
        \textbf{Question-1:} How does Joe think Anna will understand his inner emotions at this moment before she returns to the restaurant? \\
        \textbf{Options-1:} (A) Sad. \textbf{(B) Calm.} (C) Happy. (D) Doubt. \\
        \textbf{Explanation-1:} Before Anna returned to the restaurant, Joe did not know about Anna's peeking, so he did not know that Anna already knew he was sad. Therefore, Joe thought that Anna would believe his emotion was calm. So choose B. \\

        \colorbox{Belief}{Ability-B3*: Belief/High-order belief} \\
        
        \textbf{Question-2:} Anna knows how Joe thinks she thinks about Joe's inner emotions? \\
        \textbf{Options-2:} (A) Sad. \textbf{(B) Calm.} (C) Happy. (D) Doubt. \\
        \textbf{Explanation-2:} Anna knows that Joe does not know she already knows he is sad. Therefore, Anna knows that Joe will assume she believes his mood is calm. So also choose B. \\

        \colorbox{Belief}{Ability-B3: Belief/Second-order belief} \\
        
        \textbf{Question-3:} When Joe saw Anna's embarrassed expression, how did he think Anna would understand his inner emotions during Anna's departure? \\
        \textbf{Options-3:} (A) Sad. (B) Calm. (C) Happy. \textbf{(D) Doubt.} \\
        \textbf{Explanation-3:} After Joe saw Anna showing a troubled expression, he realized that his true emotion had been secretly seen by Anna. Therefore, Joe would think that Anna would realize he was in a sad mood while she was away. So choose A. \\

        \hline
    \end{tabular}
    
    \caption{2 groups of example data for \colorbox{Belief}{Naturalistic Stories [2nd-Order False Belief]}.}
    \label{tab:b3_example}
\end{table*} 

\begin{table*}[htbp]
\footnotesize
    \centering
    \begin{tabular}{|p{0.95\linewidth}|}
        \hline
        \textbf{Scene:} David was drinking whiskey on the plane, and the bumpy takeoff caused the whiskey to spill on his forehead. When he wiped the wine with a handkerchief, the flight attendant saw this behavior and asked him if he wanted to turn on the air conditioning.\\
        
        \colorbox{Belief}{Ability-B5: Belief/Beliefs based action} \\
        
        \textbf{Question-1:} Why did the flight attendant say that to David? \\
        \textbf{Options-1:} (A) Because he knew David had spilled whiskey and wanted to help him clean it up. (B) Because he needs to check if David needs a drink. (C) Because he wants to remind David to fasten his seat belt. \textbf{(D) Because he saw David wiping his forehead and thought he felt very hot.} \\
        \textbf{Explanation-1:} Since the flight attendant only saw David wiping liquid off his forehead with a handkerchief, but did not see that David had spilled wine on his forehead earlier due to turbulence, the flight attendant mistakenly believed that David was wiping away sweat because of the heat. Therefore, she thought David felt very hot, so choose D. \\

        \hline
    \end{tabular}
    
    \caption{1 group of example data for \colorbox{Belief}{Naturalistic Stories [Misattribution]}.}
    \label{tab:b4_example}
\end{table*} 

\begin{table*}[htbp]
\footnotesize
    \centering
    \begin{tabular}{|p{0.95\linewidth}|}
        \hline
        \textbf{Scene:} During the war, the Blue Army captured a member of the Red Army. They wanted him to tell them where the Red Army tanks were, as they knew they were either by the sea or in the mountains. The prisoner replied, ``In the sea.'' So, the blue army went to the mountains to search for tanks.\\
        
        \colorbox{Belief}{Ability-B5: Belief/Beliefs based action} \\
        
        \textbf{Question-1:} Why did the Blue Army go to the mountains to search for tanks? \\
        \textbf{Options-1:} (A) The Blue Army has long known that tanks are in the mountains. \textbf{(B) The Blue Army believes that prisoners will definitely lie.} (C) The Blue Army wants to verify what the prisoners said. (D) The Blue Army has long known that tanks are not in the sea. \\
        \textbf{Explanation-1:} The Blue Army inferred the prisoner's motivation based on his hostile standpoint, determining that to protect the Red Army, he would inevitably mislead them by lying. Based on this belief, they chose the target opposite to what the prisoner stated. Therefore, choose B. \\

        \colorbox{Belief}{Ability-B5: Belief/Beliefs based action} \\
        
        \textbf{Question-2:} In fact, tanks are in the sea. Why did the prisoner answer truthfully? \\
        \textbf{Options-2:} (A) The prisoner felt that reporting truthfully would benefit him. (B) The prisoner believed that the blue army going to the sea would benefit him. \textbf{(C) The prisoner believed that the Blue Army would definitely not believe what he said.} (D) The prisoner is talking nonsense. \\
        \textbf{Explanation-2:} The Red Army prisoner anticipated the Blue Army's mistrust due to their hostile standpoint and concluded that the Blue Army would view his words as lies and make the opposite choice. Therefore, the prisoner utilized this reverse psychology and adopted a strategy of answering truthfully. This was intended to induce the illusion in the Blue Army that he was lying, causing them to search in the mountains instead, thereby achieving the ultimate goal of protecting the tank. Therefore, choose C. \\

        \hline
    \end{tabular}
    
    \caption{1 group of example data for \colorbox{Belief}{Strange Stories [Double Bluff]}.}
    \label{tab:b5_example}
\end{table*} 

\begin{table*}[htbp]
\footnotesize
    \centering
    \begin{tabular}{|p{0.95\linewidth}|}
        \hline
        \textbf{Scene:} Xiaogang and Xiaoming were strolling in the bedroom when they saw handbags, briefcases, and backpacks. They found cabbage inside the handbags. The two of them left the bedroom, and after a while, Xiaogang returned to the room first. He didn't find any cabbage in his handbag, and eventually found that the cabbage had been moved by Xiaohong into his briefcase. Xiaogang moved the cabbage to the backpack. Later, Xiaoming also returned to the bedroom.\\
        
        \colorbox{Belief}{Ability-B6: Belief/Sequence false beliefs} \\
        
        \textbf{Question-1:} Where will Xiaogang look for cabbage after Xiaoming returns to the bedroom? \\
        \textbf{Options-1:} \textbf{(A) Backpack.} (B) Briefcase. (C) Plastic bag. (D) Handbag. \\
        \textbf{Explanation-1:} Xiaogang knows the cabbage is in the backpack, so he will look for it in the backpack. Therefore, choose Option A. It is unrelated to the``Unexpected Outcome'' received by Xiao Gang, which is that the cabbage appeared in the handbag. \\

        \colorbox{Belief}{Ability-B6: Belief/Sequence false beliefs} \\
        
        \textbf{Question-2:} Where will Xiaoming look for cabbage after Xiaoming returns to the bedroom? \\
        \textbf{Options-2:} (A) Backpack. (B) Briefcase. (C) Plastic bag. \textbf{(D) Handbag.} \\
        \textbf{Explanation-2:} Xiaoming does not know that Xiaogang moved the cabbage to the backpack, so he will look for the cabbage in the handbag. Therefore, choose Option D. It is unrelated to the``Unexpected Outcome'' received by Xiao Gang, which is that the cabbage appeared in the handbag. \\

        \colorbox{Belief}{Ability-B3: Belief/Second-order belief} \\
        
        \textbf{Question-3:} After Xiaoming returned to the bedroom, where did Xiaoming think Xiaogang would look for cabbage? \\
        \textbf{Options-3:} (A) Backpack. (B) Briefcase. (C) Plastic bag. \textbf{(D) Handbag.} \\
        \textbf{Explanation-3:} Xiaoming does not know that Xiaogang moved the cabbage to the backpack, therefore he thinks that Xiaogang will also look for the cabbage in the handbag. So choose D. It is unrelated to the``Unexpected Outcome'' received by Xiao Gang, which is that the cabbage appeared in the handbag. \\

        \colorbox{Belief}{Ability-B3: Belief/Second-order belief} \\
        
        \textbf{Question-4:} After Xiaoming returned to the bedroom, where did Xiaogang think Xiaoming would look for cabbage? \\
        \textbf{Options-4:} (A) Backpack. (B) Briefcase. (C) Plastic bag. \textbf{(D) Handbag.} \\
        \textbf{Explanation-4:} Xiaogang is able to recognize that Xiaoming does not know the cabbage is in the backpack, so Xiaogang thinks that Xiaoming will look for the cabbage in the handbag. Therefore, choose D. It is unrelated to the``Unexpected Outcome'' received by Xiao Gang, which is that the cabbage appeared in the handbag. \\
        
        \hline
    \end{tabular}
    
    \caption{1 group of example data for \colorbox{Belief}{Unexpected Outcome Task}.}
    \label{tab:b6_example}
\end{table*} 

\begin{table*}[htbp]
\footnotesize
    \centering
    \begin{tabular}{|p{0.95\linewidth}|}
        \hline
        \textbf{Scene:} An Ming's mother spent a long time making his favorite fish and chips. But when she brought the food to Anming, who was watching TV, she didn't even look up or say thank you. Anming's mother said irritably, ``You're really polite!''\\
        
        \colorbox{NonLiteral}{Ability-N1: Non-literal/Irony} \\
        
        \textbf{Question-1:} Why did Mom say that? \\
        \textbf{Options-1:} (A) She thinks Anming is a very polite child. (B) She lied to encourage Anming to be a very polite child. (C) She wanted to mock her own food for not being delicious. \textbf{(D) She was trying to mock An Ming as an impolite child.} \\
        \textbf{Explanation-1:} An Ming's behavior was very impolite, so his mother's purpose was to mock his impoliteness. Therefore, choose D. \\
        
        \hline
    \end{tabular}
    
    \caption{1 group of example data for \colorbox{NonLiteral}{Strange Stories [Sarcasm]}.}
    \label{tab:n1_example}
\end{table*} 

\begin{table*}[htbp]
\footnotesize
    \centering
    \begin{tabular}{|p{0.95\linewidth}|}
        \hline
        \textbf{Scene:} Qianru borrowed a beautiful novel from her friend Jianguo, but when Jianguo returned the book, Qianru found it dirty. Jianguo said his little dog got the book dirty, but in reality, he accidentally splashed water on the book himself.\\
        
        \colorbox{NonLiteral}{Ability-N2: Non-literal/Egocentric lies} \\
        
        \textbf{Question-1:} Why does Jianguo say that? \\
        \textbf{Options-1:} (A) Jianguo did see the dog dirtying the book and wanted to make Qianru angry and blame the little dog. (B) Jianguo's joke is to blame Qianru. \textbf{(C) Jianguo lied not to make Qianru angry and blame herself.} (D) Jianguo lied not to make Qianru angry and blame the puppy. \\
        \textbf{Explanation-1:} Jianguo lied and shifted the blame onto the puppy. His purpose was to avoid being blamed by the other person, so choose C. \\
        
        \hline
    \end{tabular}
    
    \caption{1 group of example data for \colorbox{NonLiteral}{Strange Stories [Lie]}.}
    \label{tab:n2_example}
\end{table*} 

\begin{table*}[htbp]
\footnotesize
    \centering
    \begin{tabular}{|p{0.95\linewidth}|}
        \hline
        \textbf{Scene:} Uncle Zhang prepared a special dish for the community's Spring Festival gathering. At the party, Aunt Wang tried this dish and felt that the taste was a bit off. When Uncle Zhang asked her for her opinion on the dishes, Aunt Wang smiled and said it was one of the most delicious dishes she had ever eaten.\\
        
        \colorbox{NonLiteral}{Ability-N3: Non-literal/White lies} \\
        
        \textbf{Question-1:} Why does Aunt Wang say that? \\
        \textbf{Options-1:} \textbf{(A) Aunt Wang said that out of respect for Uncle Zhang and consideration for maintaining community relations, she didn't want Uncle Zhang to feel embarrassed or disappointed.} (B) Aunt Wang said this because she truly believed it was one of the most delicious dishes she had ever tasted. (C) Aunt Wang wants to establish her good character image in the community, so she chose to praise Uncle Zhang's dishes. (D) Aunt Wang said this because she was worried that if she revealed the truth, it would damage her reputation in the community. \\
        \textbf{Explanation-1:} Auntie Wang lied to spare Uncle Zhang's feelings, allowing him to feel happy about the dish he made and preventing awkwardness. Therefore, choose A. \\
        
        \hline
    \end{tabular}
    
    \caption{1 group of example data for \colorbox{NonLiteral}{Strange Stories [White Lie]}.}
    \label{tab:n3_example}
\end{table*} 

\begin{table*}[htbp]
\footnotesize
    \centering
    \begin{tabular}{|p{0.95\linewidth}|}
        \hline
        \textbf{Scene:} Xiao Li is doing math homework in the room. His mother told him to go to his aunt's house in the evening. Xiao Li was busy with his homework and only thought about finishing it quickly, so he casually responded. At dinner, Dad asked, ``Are you going to Auntie's house tonight?'' Xiao Li replied, ``No, we don't have any plans.''\\
        
        \colorbox{NonLiteral}{Ability-N4: Non-literal/Involuntary lies} \\
        
        \textbf{Question-1:} Why did Xiao Li say that? \\
        \textbf{Options-1:} (A) Xiaoli deliberately lied because he didn't want to go to his aunt's house. \textbf{(B) Xiao Li forgot the plan his mother told him because he was focused on his math homework.} (C) Xiaoli deliberately lied because he wanted to continue doing his homework. (D) Xiao Li was too busy thinking about dinner to listen to his mother's plans for tonight. \\
        \textbf{Explanation-1:} Xiao Li forgot what Mom said, which led him to say something contrary to the facts. Choose B. \\
        
        \hline
    \end{tabular}
    
    \caption{1 group of example data for \colorbox{NonLiteral}{Strange Stories [Forget]}.}
    \label{tab:n4_example}
\end{table*} 

\begin{table*}[htbp]
\footnotesize
    \centering
    \begin{tabular}{|p{0.95\linewidth}|}
        \hline
        \textbf{Scene:} Rongxuan and Jinyu saw Ms. Tang coming out of the barber shop one day. Due to the barber cutting her hair too short, she looks a bit comical. Rongxuan said to Jinyu, ``She must have had a fight with the lawn mower.''\\
        
        \colorbox{NonLiteral}{Ability-N5: Non-literal/Humor} \\
        
        \textbf{Question-1:} Why did Rong Xuan say that? \\
        \textbf{Options-1:} (A) Rongxuan did misunderstand that Ms. Tang had a fight with the lawn mower. (B) Rong Xuan lied to provoke Jin Yu's anger. (C) Rongxuan wants to attract Ms. Tang's attention. \textbf{(D) Rong Xuan joked to make Jin Yu laugh.} \\
        \textbf{Explanation-1:} Rongxuan was making a joke, so choose D. \\
        
        \hline
    \end{tabular}
    
    \caption{1 group of example data for \colorbox{NonLiteral}{Strange Stories [Joke]}.}
    \label{tab:n5_example}
\end{table*} 

\begin{table*}[htbp]
\footnotesize
    \centering
    \begin{tabular}{|p{0.95\linewidth}|}
        \hline
        \textbf{Scene:} In the psychological clinic, the patient nervously tells the doctor, ``Doctor, I think I have delusions.'' The doctor asks, ``Why?'' The patient says, ``I always feel like others like to talk about me in private.''\\
        
        \colorbox{NonLiteral}{Ability-N5: Non-literal/Humor} \\
        
        \textbf{Question-1:} The doctor looked at the patient and said seriously, ``...'' Please make up for what the doctor said and make it a joke. \\
        \textbf{Options-1:} (A) Don't be nervous, this is normal, we will help you with the treatment. (B) You need to bring your medical insurance card for your next visit. \textbf{(C) Relax, they will also say it in person.} (D) We have replaced the green curtains in the waiting room. \\
        \textbf{Explanation-1:} Doctor is aware of patients' social anxiety, and saying so can exacerbate their social anxiety, which make the story a funny joke. So choose C. \\

        \colorbox{NonLiteral}{Ability-N5: Non-literal/Humor} \\
        
        \textbf{Question-2:} The doctor looked at the patient and said seriously, ``Relax, they will also say it in person.'' Why is this joke funny? \\
        \textbf{Options-2:} (A) Because the doctor felt that the patient needed medication. \textbf{(B) Doctor is aware of patients' social anxiety, and saying so can exacerbate their social anxiety.} (C) Because the doctor's words are illogical, leaving patients and readers confused. (D) Because the doctor's words reflect the current common doctor-patient conflict. \\
        \textbf{Explanation-2:} Choose B. \\
        
        \hline
    \end{tabular}
    
    \caption{1 group of example data for \colorbox{NonLiteral}{Humor Task}.}
    \label{tab:n6_example}
\end{table*} 

\begin{table*}[htbp]
\footnotesize
    \centering
    \begin{tabular}{|p{0.95\linewidth}|}
        \hline
        \textbf{Scene:} On the first day of arriving at the new school, Xiao Wang said to his classmate Xiao Zhang, ``My mother works at our school and is one of the cafeteria aunties.'' Xiao Zhang nodded. At this moment, another classmate Xiaofang walked over and interjected, ``I really don't like the cafeteria aunties in our school. They all look so scary.'' Xiaowang didn't say anything after hearing this. Xiao Zhang looked at Xiao Wang and said, ``Xiao Fang, why don't we go to the game hall together after school?'' Xiao Fang said, ``Hmm, I'm not feeling well now. Let's go another day.''\\
        
        \colorbox{NonLiteral}{Ability-N6: Non-literal/Faux pas} \\
        
        \textbf{Question-1:} Which sentence is inappropriate? \\
        \textbf{Options-1:} (A) My mother works at our school and is one of the cafeteria aunties. \textbf{(B) I really don't like the cafeteria aunties in our school. They all look so scary.} (C) Xiao Fang, why don't we go to the game hall together after school? (D) There are no inappropriate sentences. \\
        \textbf{Explanation-1:} Xiaofang said that the cafeteria ladies are fierce. Although she didn't explicitly refer to a specific person, she still offended Xiao Wang's mother by offending the entire group. Therefore, choose B. \\
        
        \hline
    \end{tabular}
    
    \caption{1 group of example data for \colorbox{NonLiteral}{Faux-pas Recognition Test}.}
    \label{tab:n7_example}
\end{table*} 

\begin{table*}[htbp]
\footnotesize
    \centering
    \begin{tabular}{|p{0.95\linewidth}|}
        \hline
        \textbf{Scene:} Zhuoying, Yichun, and Yating are good friends. Yichun has never seen a phone call, Zhuoying has never seen a banana, and Yating has seen both. At this moment, mother brought a fruit platter, which contained apples, bananas, and pears.\\
        
        \colorbox{Belief}{Ability-B4: Belief/Identity false beliefs} \\
        
        \textbf{Question-1:} If Zhuoying picks up a banana from the fruit tray and puts it to her ear, what is the reason? \\
        \textbf{Options-1:} (A) She wants to eat bananas. \textbf{(B) She thought it was a phone call.} (C) She wants to tease Yichun. (D) She is pretending to make a phone call. \\
        \textbf{Explanation-1:} Zhuoying had never seen a banana before, but she knew a phone with a shape similar to a banana, which triggered identity false beliefs and mistook the banana for a phone, so choose B. \\

        \colorbox{Intention}{Ability-I4: Intention/Intentions explanations} \\
        
        \textbf{Question-2:} If Yating picks up a banana from the fruit tray and puts it to her ear, what is the reason? \\
        \textbf{Options-2:} (A) She wants to eat bananas. (B) She thought it was a phone call. (C) She wants to tease Yichun. \textbf{(D) She is pretending to make a phone call.} \\
        \textbf{Explanation-2:} Yating knows both bananas and phones, so she is doing pretend play instead of really thinking it's a phone call. Choose D. \\

        \colorbox{Knowledge}{Ability-K1: Knowledge/Knowledge-pretend play links} \\
        
        \textbf{Question-3:} If mother picks up a banana from the fruit tray and puts it to her ear, who can't understand her movements? \\
        \textbf{Options-3:} (A) Zhuoying. \textbf{(B) Yichun.} (C) Yating. (D) Mother. \\
        \textbf{Explanation-3:} Zhuoying had never seen a banana before, so she would immediately think that her mother was making a phone call, while Yating could understand her mother's pretend play. However, Yi Chun is unable to participate in the pretend game because he does not have knowledge of telephones. Therefore, choose B. \\
        
        \hline
    \end{tabular}
    
    \caption{1 group of example data for \colorbox{Comprehensive}{Strange Stories [Pretend]}.}
    \label{tab:c1_example}
\end{table*} 

\begin{table*}[htbp]
\footnotesize
    \centering
    \begin{tabular}{|p{0.95\linewidth}|}
        \hline
        \textbf{Scene:} Jianning and Mingkai are the joint captains of the football team. There is still one player missing from their team. They joked that the remaining players who were not selected were the ``best players''. After a while, Mingkai didn't say anything, blinked at Jianning, and then looked at Taotao, who was one of the remaining unselected players. Mingkai turned around and smiled at Jianning. Jianning nodded and prepared to choose Taotao to join their team. Taotao saw the blinking and smiling of Mingkai and Jianning.\\
        
        \colorbox{Intention}{Ability-I4: Intention/Intentions explanations} \\
        
        \textbf{Question-1:} Why did Mingkai look back at Jianning and smile? \\
        \textbf{Options-1:} (A) Mingkai and Jianning had a joke about selecting team members before, and Mingkai thought of that joke. (B) Mingkai smiled because he believed Tao Tao was the best player currently available. \textbf{(C) Mingkai smiled at Jianning because they both knew they could only choose Taotao.} (D) Mingkai and Jianning have a secret plan about Taotao, so he smiled. \\
        \textbf{Explanation-1:} First, the football team is currently missing one player. Before Mingkai turned to look at Jianning, he was joking that the remaining unpicked players were ``the best players'', which indicates that they were struggling to decide whom to choose as the last member. After Mingkai looked back at Jianning and smiled, he looked at Taotao. Furthermore, Jianning was preparing to choose Taotao to join their team. Therefore, the reason Mingkai smiled at Jianning is that they both knew they had no choice but to pick Taotao. Thus, choose C. \\

        \colorbox{Emotion}{Ability-E9: Emotion/Emotion based on belief} \\
        
        \textbf{Question-2:} What do you think Taotao's mood? \\
        \textbf{Options-2:} \textbf{(A) Taotao felt surprised and confused.} (B) Taotao felt like he was being teased and ridiculed. (C) Taotao felt recognized and proud. (D) Taotao is full of hope for showcasing his skills. \\
        \textbf{Explanation-2:} Because Taotao only saw Mingkai suddenly wink at Jianning and then look at himself (Taotao), Taotao would feel surprised and confused. Thus, choose A. \\

        \hline
    \end{tabular}
    
    \caption{1 group of example data for \colorbox{Comprehensive}{Ambiguous Story Task}.}
    \label{tab:c2_example}
\end{table*} 

\begin{table*}[htbp]
\footnotesize
    \centering
    \begin{tabular}{|p{0.95\linewidth}|}
        \hline
        \textbf{Scene:} At the school design exhibition, Xiaofang said to designer Xiaojie, ``Jie Jie, your creativity is truly unique. The entire exhibition stands out because of your work!''! Xiaofang knew that the design was not well received, and she privately complained that Xiaojie's design was too ordinary. Xiaoyu, who was present, knew that Xiaofang wanted to participate in the design competition and also knew Xiaofang's true opinion on the work.\\
        
        \colorbox{Intention}{Ability-I4: Intention/Intentions explanations} \\
        
        \textbf{Question-1:} What is the main purpose of Xiaofang praising Xiaojie? \\
        \textbf{Options-1:} (A) Sincerely express admiration for Xiaojie's design creativity. \textbf{(B) Obtaining competition opportunities by flattering Xiaojie.} (C) Show the audience one's appreciation for design. (D) Objectively describe Xiaojie's actual impact on the exhibition. \\
        \textbf{Explanation-1:} Xiao Jie's design quality this time was average, and Xiao Fang even complained about it in private. However, she still praised it highly at the exhibition. Considering Xiao Jie's social status as a designer, the probability that Xiao Fang is flattering her is relatively high. Therefore, choose B. \\

        \colorbox{Intention}{Ability-I4: Intention/Intentions explanations} \\
        
        \textbf{Question-2:} If Xiaojie believed Xiaofang's praise, what would she think of her? \\
        \textbf{Options-2:} (A) Think Xiaofang is deliberately teasing her. (B) Believe that Xiaofang fully understands the actual level of design creativity. \textbf{(C) Believe that Xiaofang genuinely appreciates her design creativity.} (D) Think Xiaofang wants to expose the extraordinary design. \\
        \textbf{Explanation-2:} Because Xiao Jie only saw Xiao Fang publicly praising her and did not know that Xiao Fang disliked her design in private, if she believes Xiao Fang's praise, she would likely assume that Xiao Fang genuinely admires her design skills. Therefore, choose C. \\

        \colorbox{Belief}{Ability-B3: Belief/Second-order belief} \\
        
        \textbf{Question-3:} Xiaoyu, who was present, knew Xiaofang's true intentions. Will Xiaoyu think that Xiaojie knows the true intention behind Xiaofang's praise? \\
        \textbf{Options-3:} (A) Xiaoyu will think that Xiaojie must know that Xiaofang is deliberately flattering her. \textbf{(B) Xiaoyu will think that Xiaojie has no idea about Xiaofang's true intentions.} (C) Xiaoyu may think that Xiaojie also understands the design shortcomings, but appreciates Xiaofang's attitude. (D) Xiaoyu will think that Xiaojie only focuses on the exhibition atmosphere and doesn't listen to Xiaofang's words at all. \\
        \textbf{Explanation-3:} Although Xiao Yu knows that Xiao Fang complained in private and was eyeing the competition opportunity, Xiao Zhang also knows that Xiao Jie is unaware of this matter and only knows that Xiao Fang publicly praised her. Consequently, [he/she] would think that Xiao Jie is completely unaware of Xiao Fang's true intentions. Therefore, choose B. \\
        
        \hline
    \end{tabular}
    
    \caption{1 group of example data for \colorbox{Comprehensive}{Expanding Tasks [Flattery]}.}
    \label{tab:c3_example}
\end{table*} 

\begin{table*}[htbp]
\footnotesize
    \centering
    \begin{tabular}{|p{0.95\linewidth}|}
        \hline
        \textbf{Scene:} Xiaohua's cousin was admitted to a prestigious university, and the whole family had a dinner to celebrate. Xiaohua smiled and said to his cousin, ``Congratulations! You'll be a busy person from now on, and you won't have time to play with us ordinary people anymore.'' Afterwards, Xiaohua spent the whole night playing with his phone. When asked about his studies, he said, ``I don't have that talent, just hang around''.\\
        
        \colorbox{Emotion}{Ability-E5: Emotion/Hidden emotions} \\
        
        \textbf{Question-1:} What was Xiaohua's true emotion when he spent the whole night playing on his phone and said ``just hang out''? \\
        \textbf{Options-1:} (A) Proud of his cousin. (B) Satisfied with own grades. \textbf{(C) Hide jealousy emotions.} (D) Bored by dining together. \\
        \textbf{Explanation-1:} When his older female cousin got into a prestigious school, Xiaohua was not only unhappy for her, but instead hid behind his phone all night and even mocked himself for ``just hang around'' This deliberate show of indifference is actually because he felt mentally unbalanced after seeing someone else's success and felt he had lost face. He acted this way to prevent others from seeing that he was jealous deep down. Therefore, choose C. \\

        \colorbox{NonLiteral}{Ability-N1: Non-literal/Irony} \\
        
        \textbf{Question-2:} What is the true intention behind Xiaohua's statement, ``Congratulations! You'll be a busy person from now on ...'' \\
        \textbf{Options-2:} (A) Sincere wishes for his cousin's future. (B) Expressing expectations for future interactions. \textbf{(C) Implied irony and estrangement.} (D) Seeking comfort from his cousin. \\
        \textbf{Explanation-2:} Xiao Hua deliberately elevated his cousin to a ``busy person'' (big shot) while lowering himself to an ``ordinary person.'' This is obviously him making sour, sarcastic remarks. By doing this, he drew a line between the two of them. The subtext is, ``You are amazing now, and you are no longer one of us.'' This is a stinging mockery intended to deliberately distance himself from her. Therefore, choose C. \\

        \colorbox{Belief}{Ability-B5: Belief/Beliefs based action} \\
        
        \textbf{Question-3:} What is Xiaohua's true belief in his learning ability when he says ``I don't have that talent''? \\
        \textbf{Options-3:} (A) His is convinced that he does not have any talent. \textbf{(B) Believing oneself to have talent but unwilling to admit it.} (C) Completely indifferent to learning. (D) Believe that talent determines everything. \\
        \textbf{Explanation-3:} When he said he has ``no talent'', he was actually making an excuse. He was afraid others would say he ``doesn't work hard'' or that ``even if he worked hard, he couldn't compare to his cousin.'' So, he simply denied his talent beforehand; this way, even if his grades are bad, it wouldn't be his fault. In his heart, he actually feels he isn't bad, but to save face, he refuses to admit verbally that he has potential he hasn't utilized. Therefore, choose B. \\
        
        \hline
    \end{tabular}
    
    \caption{1 group of example data for \colorbox{Comprehensive}{Expanding Tasks [Jealousy]}.}
    \label{tab:c4_example}
\end{table*} 

\begin{table*}[htbp]
\footnotesize
    \centering
    \begin{tabular}{|p{0.95\linewidth}|}
        \hline
        \textbf{Scene:} You and Kazuha are sitting in the room, where Ayaka, Shogun, and Yoimiya are also present. Ayaka, Shogun, and you are watching Kazuha while Yoimiya is sleeping. Kazuha showed you a picture with a maple tree and said there was a treasure hidden under it. At this moment, Shogun left the house and Kokomi walked in to face Kazuha with you. Kazuha showed you another picture with a cherry tree, saying that there is also treasure under this tree.\\
        
        \colorbox{Knowledge}{Ability-K2: Knowledge/Percepts-knowledge links} \\
        
        \textbf{Question-1:} Who has as much information as you? \\
        \textbf{Options-1:} (A) Shogun. (B) Yoimiya. \textbf{(C) Ayaka.} (D) Kokomi. \\
        \textbf{Explanation-1:} First, I received all the information provided by Kazuha throughout the entire process. During the scenario, only Ayaka and Yoimiya were present the whole time. However, since Yoimiya was sleeping the entire time and lacked the ability to perceive, she did not acquire the knowledge. Therefore, choose C. \\

        \colorbox{Percept}{Ability-P3: Percept/Percept-action link} \\
        
        \textbf{Question-2:} Who would dig under the maple tree? \\
        \textbf{Options-2:} \textbf{(A) Shogun and Ayaka.} (B) Yoimiya and Ayaka. (C) Ayaka and Kokomi. (D) Kokomi and Shogun. \\
        \textbf{Explanation-2:} Because when Kazuha showed a picture of a maple tree and said that treasure was hidden under it, only I, Shogun, Ayaka, and Yoimiya were in the room. Since Yoimiya was sleeping the entire time and did not hear the information, Shogun and Ayaka will go to dig under the maple tree. Therefore, choose A. \\

        \colorbox{Percept}{Ability-P3: Percept/Percept-action link} \\
        
        \textbf{Question-3:} Who would dig under the cherry tree? \\
        \textbf{Options-3:} (A) Shogun and Ayaka. (B) Yoimiya and Ayaka. \textbf{(C) Ayaka and Kokomi.} (D) Kokomi and Shogun. \\
        \textbf{Explanation-3:} Because when Kazuha showed a picture of a cherry tree and said that treasure was hidden under it, Shogun left the room and Kokomi entered. Thus, only I, Kokomi, Ayaka, and Yoimiya were in the room. Since Yoimiya was sleeping the entire time and did not hear the information, Ayaka and Kokomi will go to dig under the cherry tree. Therefore, choose C. \\
        
        \hline
    \end{tabular}
    
    \caption{1 group of example data for \colorbox{Comprehensive}{See-know Task}.}
    \label{tab:c5_example}
\end{table*} 

\end{document}